\documentclass[final,3p,times,authoryear]{elsarticle}

\usepackage{amssymb}
\usepackage{amsmath}
\usepackage{amsthm}
\newtheorem{proposition}{Proposition}
\usepackage{lineno}
\usepackage{booktabs}
\usepackage{multirow}
\usepackage{algorithm}
\usepackage[noend]{algpseudocode}
\usepackage{xcolor}
\usepackage{colortbl}
\definecolor{sotabg}{RGB}{206,234,226}
\newcommand{\overall}{\cellcolor{sotabg}}
\newcommand{\best}[1]{\textbf{#1}}
\newcommand{\second}[1]{\underline{#1}}
\DeclareRobustCommand{\sotaswatch}[1]{{\setlength{\fboxsep}{0.8pt}\colorbox{sotabg}{\strut #1}}}
\newcommand{\shadekey}{\sotaswatch{shaded}}
\newcommand{\rankkeytf}{\sotaswatch{Shaded} and \textbf{bold}: best per column; \underline{underlined}: second best}
\newcommand{\rankkey}{\sotaswatch{Shaded}: best of all ranked rows per column, trained or not; \textbf{bold} and \underline{underlined}: best and second best among training-free methods}
\usepackage[colorlinks,allcolors=blue!60!black]{hyperref}

\newcommand{\pp}[1]{$#1$}
\newcommand{\ci}[2]{\scriptsize[$#1$,\,$#2$]}
\newcommand{\dci}[3]{\begin{tabular}[c]{@{}c@{}}\pp{#1}\\[-3pt]\ci{#2}{#3}\end{tabular}}
\makeatletter
\newcommand{\appref}[1]{\@ifundefined{supplabel@#1}{\ref{#1}}{Supplementary Section~\ref{#1}}}
\makeatother
\expandafter\def\csname supplabel@supp:model\endcsname{S1}
\expandafter\def\csname supplabel@supp:horizon\endcsname{S2}
\expandafter\def\csname supplabel@supp:horizon:crops\endcsname{S2.1}
\expandafter\def\csname supplabel@supp:horizon:resolution\endcsname{S2.2}
\expandafter\def\csname supplabel@supp:horizon:prompting\endcsname{S2.3}
\expandafter\def\csname supplabel@supp:horizon:visdrone\endcsname{S2.4}
\expandafter\def\csname supplabel@supp:horizon:sources\endcsname{S2.5}
\expandafter\def\csname supplabel@supp:horizon:detectors\endcsname{S2.6}
\expandafter\def\csname supplabel@supp:method\endcsname{S3}
\expandafter\def\csname supplabel@supp:method:grammar\endcsname{S3.1}
\expandafter\def\csname supplabel@supp:method:prompts\endcsname{S3.2}
\expandafter\def\csname supplabel@supp:method:ladder\endcsname{S3.3}
\expandafter\def\csname supplabel@supp:method:correctness\endcsname{S3.4}
\expandafter\def\csname supplabel@supp:method:synth\endcsname{S3.5}
\expandafter\def\csname supplabel@supp:protocol\endcsname{S4}
\expandafter\def\csname supplabel@supp:protocol:splits\endcsname{S4.1}
\expandafter\def\csname supplabel@supp:protocol:leakage\endcsname{S4.2}
\expandafter\def\csname supplabel@supp:protocol:zeroshot\endcsname{S4.3}
\expandafter\def\csname supplabel@supp:protocol:constant\endcsname{S4.4}
\expandafter\def\csname supplabel@supp:repro\endcsname{S5}
\expandafter\def\csname supplabel@supp:repro:selection\endcsname{S5.1}
\expandafter\def\csname supplabel@supp:repro:settings\endcsname{S5.2}
\expandafter\def\csname supplabel@supp:repro:numbers\endcsname{S5.3}
\expandafter\def\csname supplabel@supp:repro:controls\endcsname{S5.4}
\expandafter\def\csname supplabel@supp:repro:tbv\endcsname{S5.5}
\expandafter\def\csname supplabel@supp:results\endcsname{S6}
\expandafter\def\csname supplabel@supp:results:xlrs\endcsname{S6.1}
\expandafter\def\csname supplabel@supp:results:mme\endcsname{S6.2}
\expandafter\def\csname supplabel@supp:results:mmeoptions\endcsname{S6.3}
\expandafter\def\csname supplabel@supp:results:source\endcsname{S6.4}
\expandafter\def\csname supplabel@supp:results:ablation\endcsname{S6.5}
\expandafter\def\csname supplabel@supp:results:window\endcsname{S6.6}
\expandafter\def\csname supplabel@supp:results:internvl\endcsname{S6.7}
\expandafter\def\csname supplabel@supp:results:budget\endcsname{S6.8}
\expandafter\def\csname supplabel@supp:umicro\endcsname{S6.9}
\expandafter\def\csname supplabel@supp:cost\endcsname{S7}
\expandafter\def\csname supplabel@supp:failures\endcsname{S8}

\hypersetup{pdftitle={GeoScope: Training-Free Program Execution beyond the Counting Horizon for Ultra-High-Resolution Remote Sensing Question Answering}}

\myfooter[L]{Preprint}

\usepackage[breakable]{tcolorbox}
\newtcolorbox{promptbox}[1]{breakable, title={#1}, fonttitle=\footnotesize\bfseries, fontupper=\scriptsize,
  colback=teal!3!white, colframe=teal!55!black, coltitle=white, colbacktitle=teal!55!black,
  boxrule=0.5pt, arc=1.5pt, left=5pt, right=5pt, top=2pt, bottom=2pt, toptitle=1pt, bottomtitle=1pt,
  before skip=8pt, after skip=8pt, grow to left by=6pt, grow to right by=6pt}
\usepackage{placeins}
\journal{ISPRS Journal of Photogrammetry and Remote Sensing}

\begin{document}
\addtocontents{toc}{\protect\setcounter{tocdepth}{-1}}

\hypersetup{pageanchor=false}
\begin{frontmatter}

\title{GeoScope: Training-Free Program Execution beyond the Counting Horizon for Ultra-High-Resolution Remote Sensing Question Answering}

\author[ahu,agro]{Guohui~Deng\fnref{eq}\corref{cor1}}
\makeatletter\@eadauthor={G. Deng}\makeatother
\ead{dlyj0207@whu.edu.cn}
\author[ahu]{Yuhang~Jiang\fnref{eq}}
\author[xtu]{Chengjun~Wang}
\author[szu]{Zhongwen~Hu}
\author[whurs]{Zhaocong~Wu}
\author[ahu]{Linsheng~Huang}
\author[ahu]{Jinling~Zhao}
\author[ahu]{Chao~Ruan}
\author[liesmars]{Miaozhong~Xu}
\author[liesmars]{Chi~Chen}
\cortext[cor1]{Corresponding author.}
\fntext[eq]{These authors contributed equally to this work.}
\affiliation[ahu]{organization={School of Internet, Anhui University},
  citysep={}, city={Hefei}, postcode={230039}, country={China}}
\affiliation[agro]{organization={National Engineering Research Center for Agro-Ecological Big Data Analysis \& Application, Anhui University},
  citysep={}, city={Hefei}, postcode={230601}, country={China}}
\affiliation[xtu]{organization={School of Computer Science and Cyberspace Security, Xiangtan University},
  citysep={}, city={Xiangtan}, postcode={411105}, country={China}}
\affiliation[szu]{organization={School of Architecture and Urban Planning, Shenzhen University},
  citysep={}, city={Shenzhen}, postcode={518060}, country={China}}
\affiliation[whurs]{organization={School of Remote Sensing and Information Engineering, Wuhan University},
  citysep={}, city={Wuhan}, postcode={430079}, country={China}}
\affiliation[liesmars]{organization={State Key Laboratory of Information Engineering in Surveying, Mapping and Remote Sensing, Wuhan University},
  citysep={}, city={Wuhan}, postcode={430079}, country={China}}

\begin{abstract}
In ultra-high-resolution (UHR) remote sensing images, the objects that questions count in a region or relate by
direction often span only a few pixels. Existing methods mainly
decide where a vision--language model (VLM) should look and trust its answer in the chosen view. We show that this
assumption fails for counting, and measure VLMs as one calibrates a sensor. All eight open VLMs tested fall
increasingly short once a view holds more than a certain number of objects, about five for most on DOTA aerial images.
This \emph{counting horizon} narrows as views grow and objects appear smaller. An empirical power-law response model predicts
the undercount a split recovers and the spurious counts it adds, and supports the rule of splitting only views beyond the
horizon. In GeoScope, a text-only language model parses the question into a typed spatial program that fixes the
region and its resolution. The VLM only grounds and counts within its operating range, and directions are computed
explicitly from the grounded boxes. To our knowledge, it is the first UHR method to split views by a measured limit of
the VLM. Without training, it improves Qwen3-VL-8B by 16.3 points on regional counting and 10.5 on spatial relations
in XLRS-Bench, and has the highest reported accuracy, trained methods included, on regional counting and on the whole
benchmark by question. On UHR-Micro it improves global and regional counting by 16.8 and 36.2 points, 15.6 of the
former from splitting. On MME-RealWorld counting it gains 11.4 points, and we show that an option that is
almost never correct yet often chosen distorts comparisons there.
Project page: \url{https://avalon-s.github.io/GeoScope/}.
\end{abstract}


\end{frontmatter}
\hypersetup{pageanchor=true}

\section{Introduction}
\label{sec:intro}
{\let\thefootnote\relax\footnotetext{\textit{Abbreviations:} MME-RS, the remote sensing part of MME-RealWorld; OSR,
object spatial relationship; RC, regional counting; UHR, ultra-high resolution; VLM, vision--language model; VQA, visual
question answering.}}

Ultra-high-resolution (UHR) remote sensing images, with scenes of several thousand to ten thousand pixels on a side, have become
a standard test of whether vision--language models (VLMs) can support Earth
observation~\citep{xlrsbench,mmerealworld,uhrmicro}. Many of their questions concern small objects in large places,
such as how many trucks are parked in the lot below the white building, or in which direction a pool lies from the
tennis court. The same questions arise in practice. Inventories of vehicles, ships or storage tanks, damage counts
after a disaster and the layout of facilities for planning all depend on how many objects a place holds or where one
lies relative to another. A VLM sees a few million pixels per image, so a scene of ten thousand pixels is shrunk at
least fivefold per side, and a car that spans 15 pixels at a ground sampling distance of 0.3\,m keeps two or three.
Before any object is recognised, a method must therefore decide which part of the scene the model sees and at what
resolution.

Existing methods address where to look. Trained UHR models extend the visual context, adapt to the ground sampling
distance or learn where to zoom through fine-tuning, reinforcement learning or
self-distillation~\citep{geollava8k,geoeyes,geolens,geovista,zoomearth,uhrbat,scaleearth,rsopsd}. Training-free methods keep the
model frozen and choose its input instead, by searching over zoom levels~\citep{zoomeye,zoomsearch}, retrieving
patches~\citep{rap}, following attention~\citep{radar}, selecting evidence~\citep{weaveearth} or
directing gaze~\citep{gazeearth}. Both lines assume that once the right view is chosen, the counts and directions the
VLM gives in it can be trusted. For counting this does not hold. Of position, colour and counting, the three
MME-RealWorld remote sensing categories, counting is the weakest for every method that reports all three
(Section~\ref{sec:exp:published}). WeaveEarth notes that it ``is still prone to errors on counting
tasks''~\citep{weaveearth}, and question-guided gaze raises MME-RealWorld counting accuracy by less than half a point,
from 21.35\% to 21.78\%~\citep{gazeearth}. Besides where to look, UHR question answering therefore has to ask when
the VLM's observations can be trusted.

\begin{figure}[!tp]
\centering
\includegraphics[width=\linewidth]{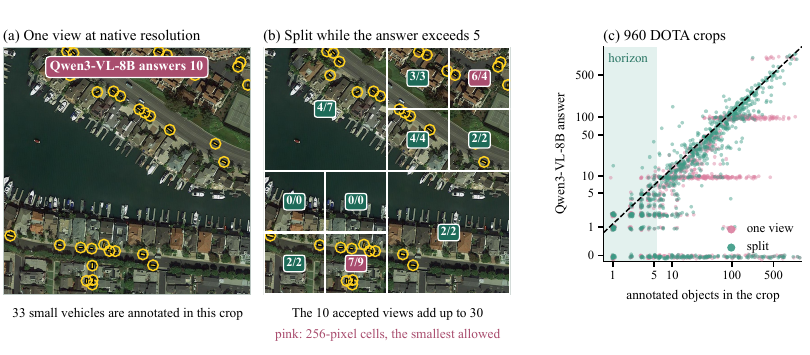}
\caption{The counting horizon on one DOTA crop. (a) In one view, Qwen3-VL-8B answers 10 for 33 annotated small
vehicles. (b) Views are split while the answer exceeds five, down to cells of 256 pixels; each shows the answer over
the annotated count, and the pink ones are such cells, which are not split further. (c) Answers for the 960 DOTA crops of Section~\ref{sec:horizon:measure}, in
one view and split.}
\label{fig:teaser}
\end{figure}

We answer this question as one calibrates a sensor. On DOTA aerial images with exact counts, Qwen3-VL-8B counts up to
about five objects in a single view with little bias (Fig.~\ref{fig:teaser}). Beyond that its answers fall
increasingly short and often settle on round numbers, like a sensor saturating beyond its dynamic range. We call the
range of reliable counts the \emph{counting horizon}. All eight open VLMs we test, from three families, saturate in the
same way, most of them beyond about five objects (Section~\ref{sec:horizon:measure}). The horizon narrows as the view grows and
its objects appear smaller. In small views shown magnified, Qwen3-VL-8B
counts up to 160 objects to within 21\%, but in large views at native resolution it falls short at every number of
objects. On cars in FAIR1M and ships in DIOR, two aerial datasets with smaller images, the horizon lies further out.

Beyond the horizon, the natural remedy is a smaller unit of observation, that is, a split view. Splitting recovers
much of the shortfall, yet most objects still end in parts that hold more than five, and an empty part is answered
with one or two spurious objects on average. An empirical power-law model of the VLM's counts, the
response curve of this sensor, predicts from unsplit views alone how much of the shortfall each split recovers
(Section~\ref{sec:horizon:theory}). The prediction holds on a second, independently drawn set of crops and, mostly within 20\%,
for seven other models. Its spurious-count term also explains the problem of fixed grids: a grid fine enough for
crowded regions overcounts sparse ones. The result is a single rule, to split a view only when its count
exceeds the counting horizon. The response model does not run inside the method. Like a calibration, it supports the
rule, and at run time the method needs only one threshold, the horizon.

Directions show a similar limit. In the direction questions of UHR-Micro the two objects are already marked with
rectangles, and their boxes alone determine the answer, yet Qwen3-VL-8B answers only about two thirds of them
correctly (Section~\ref{sec:exp:umicro}). Even when the objects are marked and visible, the VLM's judgement of their
relative direction cannot be relied on.

\begin{figure}[!tp]
\centering
\includegraphics[width=\linewidth]{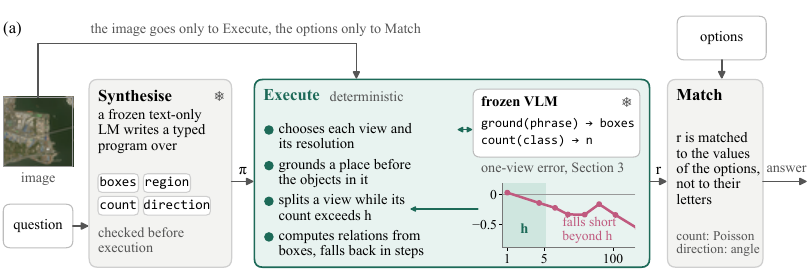}\\[4pt]
\includegraphics[width=\linewidth]{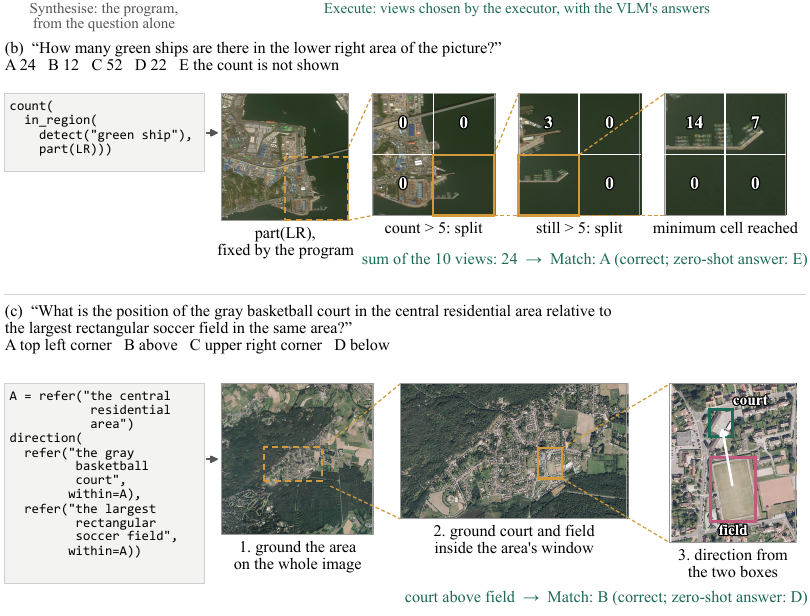}
\caption{GeoScope. (a) The three stages; the executor's split rule (teal) uses the horizon $h$ measured in
Section~\ref{sec:horizon} (inset: Qwen3-VL-8B in one view, as in Fig.~\ref{fig:horizon}a). (b, c) The same
stages on a counting and a relation test question. Dashed lines link a box to its enlarged view.}
\label{fig:pipeline}
\end{figure}

Building on these measurements, we present GeoScope (Fig.~\ref{fig:pipeline}), which treats the VLM as a sensor with a
calibrated operating range. The VLM only grounds objects and counts locally within that
range, while the spatial reasoning, from delimiting regions and summing the counts of split views to judging
directions, is done by deterministic geometric operations. To our knowledge, GeoScope is the first UHR question
answering method that decides how to split a view from a measured counting limit of the VLM. Each of its design
choices answers a measured limitation. Because the horizon narrows as views grow, the program first fixes the region
the question names, which the same pixel budget then shows at a higher resolution. A single view falls short beyond the
horizon, so only views counted beyond it are split. Small objects can vanish when a view is downsampled, so a zero from a downsampled view is split as well. A
small object is hard to see on the whole image while the place that contains it is not, so the place is grounded
first and then the object inside it. The VLM misjudges directions even between marked objects, so a direction is
computed from the two boxes with the direction-relation matrix~\citep{cdc}.

GeoScope differs from earlier work in three ways. First, UHR methods choose where to look from the VLM's own
confidence or attention and take the final answer from the VLM. In GeoScope a spatial program parsed from the
question fixes the region, and the VLM provides only local observations. Second, counting by splitting is not
new~\citep{ssdcnet,countgdpp,lvlmcount}, but earlier methods either rely on a trained counter or split once. GeoScope
splits on a measured property of a frozen VLM, and the response model estimates in advance what a split does. Third,
GeoScope keeps the typed spatial programs and executor of our earlier GeoSelect~\citep{geoselect}, which addresses
referring image segmentation. What it adds is perception only through a frozen VLM and all the perception design above. The
programs are written by a frozen, text-only language model that sees neither the image nor the options, so a program
depends on the question alone.

We evaluate GeoScope on three UHR benchmarks, making all design choices on development questions and reporting on
held-out ones, and reproduce four recent training-free methods with their official code. On the regional counting (RC)
and object spatial relationship (OSR) tasks of XLRS-Bench~\citep{xlrsbench}, GeoScope improves
Qwen3-VL-8B~\citep{qwen3vl} by 16.3 and 10.5 points. On the counting task of MME-RealWorld remote
sensing~\citep{mmerealworld} it gains 11.4 points. Each gain is larger than that of any reproduced method over its own
backbone. On
UHR-Micro~\citep{uhrmicro}, which scores a count by its error, it improves global and regional counting by 16.8 and
36.2 points and is above MAP, the benchmark's own training-free method. Its results on XLRS-Bench RC, on the whole of
XLRS-Bench by question over all 3{,}080 questions and, with InternVL3.5-8B, on UHR-Micro regional counting are the
highest reported, trained methods included.

Ablations show that the three benchmarks test different limits of the VLM (Section~\ref{sec:exp:ablation}). On
UHR-Micro global counting, where scenes are dense and a count is scored by its error, subdivision adds 15.6 of the 16.8
points. On regional counting in XLRS-Bench and UHR-Micro the gain comes mostly from the window that the program fixes, and on
spatial relations from the direction computed from the boxes. MME-RealWorld counting also exposes an option bias. The
option stating that the image does not show the number is correct for 5 of the 1{,}226 questions, yet the backbones
choose it for 18 to 57\% of the test questions. We therefore also compare with each backbone restricted to the numeric
options (Section~\ref{sec:discussion:subdivision}). This restricted backbone alone significantly beats every
reproduced method run on the whole test split, while GeoScope is level with it with Qwen3-VL-8B and significantly
above it with Qwen3-VL-4B.

\noindent Our contributions are:
\begin{itemize}
\item \textbf{(C1) The counting horizon and its response model.} Measuring frozen VLMs as one calibrates a sensor, we
  find that eight open VLMs from three families fall increasingly short once a view holds more objects than their
  counting horizon, which moves with the size of the view, the apparent size of the objects and the imagery. A
  power-law response model predicts how much of the undercount a split recovers and how many spurious counts it adds.
  It holds on a second set of crops and across models, and supports the rule of splitting only views counted beyond
  the horizon.
\item \textbf{(C2) GeoScope.} A training-free method for UHR question answering whose design follows from the measured
  limits of the VLM. A spatial program parsed from the question fixes the region, the VLM grounds and counts only
  within its operating range, views counted beyond the horizon are split, and directions are computed explicitly from
  the grounded boxes. On XLRS-Bench and MME-RealWorld it improves its backbone more than any reproduced training-free
  method, and on UHR-Micro it is above the benchmark's own training-free method with the same backbone.
\item \textbf{(C3) Sources of the gains and a benchmark bias.} Ablations attribute the gains to the region,
  subdivision and the computed directions. MME-RealWorld counting scores depend strongly on an option that is almost
  never correct and on where the correct number sits among the options, and we discuss what a benchmark needs in
  order to tell counting methods apart (Section~\ref{sec:discussion:subdivision}).
\end{itemize}

\section{Related work}
\label{sec:related}

\subsection{Remote sensing VQA}
\label{sec:related:uhr}
Trained task-specific networks answer remote sensing questions, including counting questions, on images of about a
thousand pixels on a side~\citep{hrvqa,earthvqanet} and on paired optical and SAR images~\citep{tgfnet}. Remote sensing
assistants instruction-tune a VLM on remote sensing data for captioning and VQA, most of them also for
grounding~\citep{rsgpt,geochat,earthgpt,lhrsbot,lhrsbotnova,skyeyegpt,skysensegpt,earthdial}. Remote sensing
foundation models pre-train image encoders on large image collections, by masked image modelling~\citep{ringmo} or by
aligning images with text, as \mbox{RemoteCLIP}~\citep{remoteclip} does, and are fine-tuned or probed for each task. \mbox{RemoteCLIP} also introduces a
counting benchmark. Further work trains grounding networks and VLMs for grounding and reasoning segmentation, often
with reinforcement learning~\citep{amvg,geopixel,migrant,dvgbench,think2seg,geor1,grasp}. On XLRS-Bench, GeoChat scores
below every general VLM evaluated alongside it, and EarthDial below most of them~\citep{geollava8k,uhrbat}.

XLRS-Bench~\citep{xlrsbench} asks perception and reasoning questions, including regional counting and object spatial
relationships, on images of up to $10{,}000\times10{,}000$ pixels. MME-RealWorld~\citep{mmerealworld} asks about
object position, colour and count in large aerial scenes. RSHR-Bench~\citep{rshrbench} adds further UHR scenes, and
UHR-Micro~\citep{uhrmicro} asks about objects that cover on average less than 0.01\% of the image. Trained approaches
extend the input with token pruning~\citep{geollava8k} or compress tokens under a budget~\citep{uhrbat}. Others learn
to zoom, crop or call tools through fine-tuning and reinforcement
learning~\citep{geoeyes,geolens,geovista,zoomearth,croptoolrl}, adapt to the ground sampling distance of the input with
gated low-rank adapters~\citep{scaleearth}, or self-distil from a teacher with privileged evidence~\citep{rsopsd}. All
of them answer from their own perception of the image or of the views they select.

\subsection{Training-free high-resolution perception}
\label{sec:related:tf}
ZoomEye searches a zoom tree~\citep{zoomeye}, RAP retrieves query-relevant crops and reassembles them~\citep{rap},
DC$^2$ divides and merges patch descriptions~\citep{dc2}, and ViCrop crops where attention points~\citep{vicrop}. In
remote sensing, ImageRAG retrieves patches for an answering model that it fine-tunes lightly~\citep{imagerag}, and
RADAR localises from attention~\citep{radar}. ZoomSearch searches a tree of 3$\times$3 splits scored by a visual
retriever and the VLM's confidence~\citep{zoomsearch}. WeaveEarth selects evidence tiles~\citep{weaveearth}. GazeEarth
gazes at no more than two cells of a 4$\times$4 grid and answers from a foveated warp~\citep{gazeearth}, and MAP builds
and refines a spatially grounded state of local evidence~\citep{uhrmicro}. These methods decide where the model looks;
none uses a measured limit of the VLM to decide when a view must be split.

\subsection{Programs, tools and agents}
\label{sec:related:programs}
VisProg and ViperGPT compose vision modules into programs written by a language model~\citep{visprog,vipergpt}.
DiSciPLE searches with a language model for interpretable programs that predict scientific variables, such as
population density, from satellite imagery~\citep{disciple}. Geode has a language model write code over geospatial
data and retrieval services~\citep{geode}, and Change-Agent lets a language model write Python over the outputs of
trained change models~\citep{changeagent}. RS-Agent and ThinkGeo plan and benchmark tool calls~\citep{rsagent,thinkgeo},
IC-EO generates auditable Earth-observation code~\citep{iceo}, and QVLM writes code over the masks of a segmenter
trained by its authors~\citep{qvlm}. GeoPilot and WeaveAgent train models to decide which tools a remote sensing
question needs~\citep{geopilot,weaveagent}, and \citet{selectivetooluse} fine-tune a VLM to decide when to call
deterministic change-analysis tools. GeoScout-Agent lets a proprietary model zoom, sharpen, run code and
segment~\citep{geoscoutagent}. GaiaAgent coordinates language-model agents around a trained change-detection
model~\citep{gaiaagent}, and BEDI benchmarks VLM-based agents on drones~\citep{bedi}. Changes between two images can
also be captioned without training, by having a text-only language model compare the VLM's description of
each~\citep{ricci2026changecap}. Concurrently with this work, TerraAgent wraps a frozen remote sensing model with agents
that split a question and route its parts to perception tools~\citep{terraagent}. Our previous work,
GeoSelect~\citep{geoselect}, compiles referring expressions into typed spatial programs that are checked and executed
deterministically. GeoScope keeps this scaffold, but perceives only through a frozen VLM and derives its counting
procedure from a measured property of that VLM.

\subsection{Counting limits of VLMs}
\label{sec:related:counting}
Humans enumerate small sets at a glance and estimate larger ones~\citep{subitizing}. VLMs fail beyond a few
objects~\citep{countclip,campbell2024binding,vlmblind}, also on a unified counting benchmark of real
images~\citep{unicbench}, and diverge beyond the subitising range on synthetic boards~\citep{pang2026bottleneck}. They
underestimate large numerosities~\citep{testolin2025enumeration,zhao2026halo} and favour round numbers such as 5 and
10~\citep{capture}. List-then-count and decomposition prompts, pointing and reasoning do not consistently improve
counting without training~\citep{alghisi2026pointing,vlmcountbench,visnumbench}. Diagnostic benchmarks list counting
beyond the subitising range among the cognitive biases of VLMs~\citep{sar2026realitycheck}, and show that models fall
back on object priors against the visual evidence~\citep{alzahrani2026countercount}.

Probing studies locate the failure inside the model. Numerosity is linearly decodable early in the vision encoder,
while performance differences follow the language model~\citep{numerosityvlm}. A counting circuit shared across visual
reasoning tasks yields precise counts for small numbers and noisy estimates for large ones~\citep{countingcircuits}.
As objects multiply, attention compression erodes the separation between the representations of neighbouring counts,
which a fine-tuned convolutional module alleviates~\citep{che2026convstack}. These works measure or explain the limit.
Dedicated open-world counters such as CountGD are trained on counting data and take a text or exemplar
prompt~\citep{countgd}. Remote sensing counters regress density maps of dense scenes and are trained to generalise
across domains~\citep{scoutcount}. Other methods work around the limit by splitting. SS-DCNet divides dense regions
until sub-region counts fall within the closed set its counter has seen~\citep{ssdcnet}. CountGD++ crops the image
adaptively once it has counted at least 800 objects~\citep{countgdpp}, and LVLM-Count cuts an image once into
object-aware sub-images~\citep{lvlmcount}. We instead decide when to split a view from a model of how a split changes a frozen VLM's answer, tested on aerial
imagery.

\section{Counting horizon of frozen VLMs}
\label{sec:horizon}

\subsection{Measurement on aerial crops}
\label{sec:horizon:measure}

\subsubsection{Setup}
\begin{figure}[!tp]
\centering
\includegraphics[width=\linewidth]{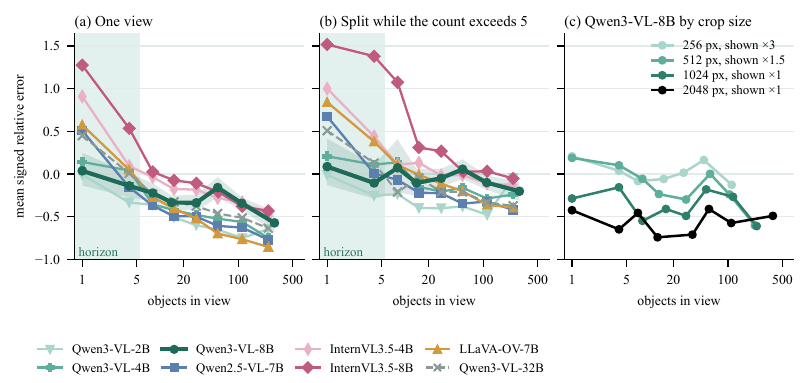}
\caption{Counting bias on DOTA-v1.5 crops: mean signed relative error by the number of objects (a) in one view for
eight VLMs, (b) with count-triggered subdivision at $h=5$ and (c) for Qwen3-VL-8B by crop side and enlargement. Bands:
95\% bootstrap intervals; shaded: at most five objects.}
\label{fig:horizon}
\end{figure}

We cut square crops 256 to 4{,}096 pixels on a side, clipped where the image is smaller, from a pool of 260 DOTA-v1.5
training images~\citep{dota,dotav2}. No image that appears in XLRS-Bench or MME-RealWorld is used, and neither these crops nor the FAIR1M and DIOR crops
used below come from an image of a UHR-Micro test question. Each crop is centred
on cars, trucks, ships or storage tanks, one class per crop, and its ground truth $N$ is the number of annotated
instances of that class in the crop. We draw 60 crops for each of eight bins of $N$, from one or two objects to more
than 160. Crops are shown at native resolution, with two exceptions: a crop whose long side is below 768 pixels is
enlarged to 768, and one with more than $2048^2$ pixels is reduced to that many. The question is the open-ended one the executor uses,
which names the class and asks for a single integer (\appref{supp:method:prompts}). We report the mean signed relative error
$(\hat n-N)/N$ per bin, where $\hat n$ is the model's answer. We also count every crop with count-triggered
subdivision, which splits a view into four parts, or two if it is elongated, while its answer exceeds a threshold $h$ and sums the answers of the
parts (Section~\ref{sec:method:count}). Besides four Qwen-VL models~\citep{qwen3vl,qwen25vl} we test InternVL3.5-4B
and 8B~\citep{internvl35} and LLaVA-OneVision-7B (LLaVA-OV-7B)~\citep{llavaov}. For Qwen3-VL-8B we also draw a second set
of crops with another seed, on which the model of Section~\ref{sec:horizon:theory} is tested with its exponent fitted on
the first. We also show its crops at a budget of $1024^2$ pixels, which changes only those larger than 1{,}024 pixels. Qwen3-VL-32B, queried through an API, is shown for reference. In Fig.~\ref{fig:horizon} the band around Qwen3-VL-8B is a
bootstrap interval over the 960 crops of both sets, and bins with fewer than eight crops are left out of panel (c).

\subsubsection{Findings}
Pooled over both sets of crops, the error of Qwen3-VL-8B stays within 15\% for up to five objects and turns
systematically negative beyond, down to $-57$\% above 160 objects (Fig.~\ref{fig:horizon}a). Beyond about five objects most answers are
undercounts, and many are round numbers. On sets of up to
20 objects, neither listing every instance before counting nor the reasoning variant Qwen3-VL-8B-Thinking is
significantly more exact, and asking for one box per object degenerates into long lists
(\appref{supp:horizon:prompting}). The shortfall also depends on the view (Fig.~\ref{fig:horizon}c). Crops of 256
pixels, shown enlarged threefold, stay within 21\% up to 160 objects, while crops of 2{,}048 pixels at native
resolution fall short at every number of objects. Our denser crops are also larger, so panels (a) and (b) confound the
number of objects with the size of the view and of the objects in it.

The other models show the same pattern with different onsets (Table~\ref{tab:fit}). Most fall short by more than 20\%
from the bin of 6 to 10 objects onward, Qwen3-VL-2B already from three to five, and InternVL3.5, which overcounts small
sets, only beyond 40. At the same number of objects, every model falls short more in views of 2{,}048 pixels than of
256 pixels, InternVL3.5-8B only slightly (\appref{supp:horizon:resolution}). We call the number of objects up
to which a model counts a view with little bias, before the onset in Table~\ref{tab:fit}, its counting horizon. Pooled over our DOTA crops it is about five.

\begin{table}[!tp]
\centering
\caption{Onset of the shortfall and fitted response of each model on the DOTA crops.}
\label{tab:fit}
\footnotesize
\setlength{\tabcolsep}{4pt}
\begin{tabular}{lccccc}
\toprule
Model & Shortfall from & $\gamma$ & $a$ & $s$ & Crops in fit \\
\midrule
Qwen3-VL-2B & 3--5 & 0.48 & 1.76 & 0.79 & 316 \\
Qwen3-VL-4B & 6--10 & 0.64 & 1.59 & 1.79 & 314 \\
Qwen3-VL-8B, first set & 6--10 & 0.76 & 1.36 & 1.56 & 297 \\
Qwen3-VL-8B, second set & 3--5 & 0.81 & 1.28 & 1.41 & 299 \\
Qwen3-VL-32B & 6--10 & 0.80 & 1.33 & 0.91 & 285 \\
Qwen2.5-VL-7B & 6--10 & 0.70 & 1.10 & 2.15 & 305 \\
InternVL3.5-4B & 41--80 & 0.67 & 1.78 & 1.66 & 352 \\
InternVL3.5-8B & 41--80 & 0.75 & 1.81 & 2.36 & 331 \\
LLaVA-OV-7B & 6--10 & 0.43 & 2.29 & 1.75 & 335 \\
\bottomrule
\end{tabular}
\par\smallskip
\begin{minipage}{\textwidth}\footnotesize
Shortfall from: first bin of the true count whose mean single-view error is below $-0.2$. $\gamma$, $a$: least
squares of $\log c$ on $\log N$ over the unsplit crops with more than five objects and a non-zero answer
(Eq.~\eqref{eq:response}); crops in fit: their number. $s$: mean answer to the parts of split crops without an instance of
the queried class.
\end{minipage}
\end{table}

\subsubsection{Other imagery}
\begin{figure}[!tp]
\centering
\includegraphics[width=\linewidth]{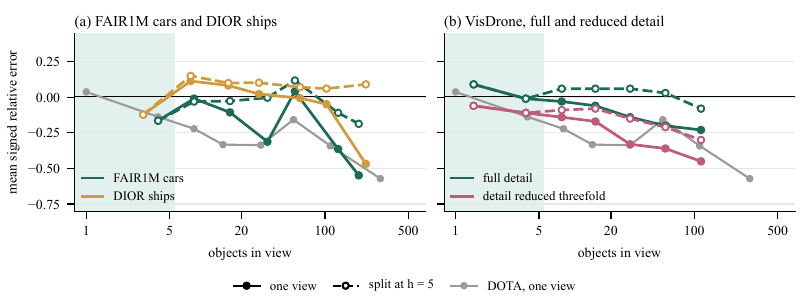}
\caption{Counting bias of Qwen3-VL-8B on other imagery: mean signed relative error by the number of objects in one
view and with subdivision at $h=5$, (a) on FAIR1M cars and DIOR ships and (b) on VisDrone at full and threefold
reduced detail. Grey: DOTA in one view; shaded: at most five objects; bins with fewer than eight crops are left out.}
\label{fig:imagery}
\end{figure}

We repeated the measurement with Qwen3-VL-8B on FAIR1M~\citep{fair1m} and DIOR~\citep{dior}, two aerial detection
datasets used by neither XLRS-Bench nor MME-RealWorld, and on VisDrone drone images~\citep{visdrone}, whose objects
are larger in pixels (Fig.~\ref{fig:imagery}). FAIR1M and DIOR give 480 and 445 crops of at most 1{,}500 pixels on a
side, against up to 4{,}096 on DOTA. The shortfall appears on all three but sets in later than on DOTA. FAIR1M cars
fall short by more than 20\% beyond 20 objects, though not monotonically, and DIOR ships only above 160. On VisDrone
the error stays within 15\% up to 40 objects. Once the VisDrone images are downsampled threefold, the error is
negative in every bin and passes $-20$\% beyond 20. Subdivision with the horizon measured on DOTA
removes most of the shortfall on FAIR1M cars and full-detail VisDrone, and part of it at reduced detail. On DIOR ships
it adds a surplus of up to 15\%, the spurious counts of views split while still within the horizon
(Section~\ref{sec:horizon:theory}). For FAIR1M trucks and DIOR vehicles the annotation does not match the question we ask
(\appref{supp:horizon:sources}). The horizon thus depends on the imagery, including the size of its views, as
well as on the model. GeoScope uses the horizon measured on DOTA, the source of 96\% of the MME-RealWorld counting images
(\appref{supp:results:source}).

\subsection{Response model of splitting}
\label{sec:horizon:theory}

The measurements suggest a simple model of a single view. Let a view contain $n$ objects of the queried class and let
$c(n)$ be the expected answer. We model
\begin{equation}
c(n) = a\,n^{\gamma}\ \ (n\geq 1),\qquad c(0)=s,
\label{eq:response}
\end{equation}
with $0<\gamma\leq 1$ and a spurious answer $s\geq 0$ for an empty view. Near the horizon $c(n)\approx n$, matching
the unbiased small counts, while for large $n$ the answer is only a fraction $a\,n^{\gamma-1}$ of the true count.
Fitted by least squares on logarithms, which recovers $\gamma$ under multiplicative noise, on unsplit crops with more
than five objects and a non-zero answer, the exponent is 0.76 for Qwen3-VL-8B on its first set and lies between 0.43
and 0.80 for the other models (Table~\ref{tab:fit}). Pooled over crop sizes, the
exponent also absorbs the loss of small objects in large views; fitted within each crop size up to 2{,}048 pixels, that of
Qwen3-VL-8B is closer to one (\appref{supp:model}). Eq.~\eqref{eq:response} is a descriptive
model of the answers on such crops, which we use to predict what a split does to them.

\begin{proposition}[Split amplification]\label{prop:split}
Split a view holding $n\geq 1$ objects into parts holding $n_1,\dots,n_K$ objects, $\sum_k n_k=n$, of which $m$
are empty. Under Eq.~\eqref{eq:response}, with the same $a$, $\gamma$ and $s$ for the parts as for the view, the
expected sum of the parts' answers is
\begin{equation}
\sum_{k=1}^{K} c(n_k) \;=\; c(n)\,\sum_{k:\,n_k>0}\Big(\frac{n_k}{n}\Big)^{\gamma} \;+\; m\,s ,
\label{eq:split}
\end{equation}
and the amplification $A=\sum_k (n_k/n)^{\gamma}$ satisfies $1\leq A\leq K^{1-\gamma}$. For $\gamma<1$, $A=1$ only
when all objects fall into one part.
\end{proposition}
\begin{proof}
Eq.~\eqref{eq:split} follows from Eq.~\eqref{eq:response} term by term. The bounds hold because $x\mapsto x^{\gamma}$
is concave and zero at zero, hence subadditive, and its sum over $K$ shares is largest when the shares are equal
(\appref{supp:theory}).
\end{proof}
A split therefore raises an undercounted answer by a factor fixed by how the objects divide, unless all objects fall into
one part, and adds $s$ for every empty part. Recursion shrinks the undercount of every part, but the spurious term
grows with the number of empty parts. A view already within the horizon gains nothing from a split, so splitting
should be triggered by the answer.

\begin{figure}[!tp]
\centering
\includegraphics[width=0.722\linewidth]{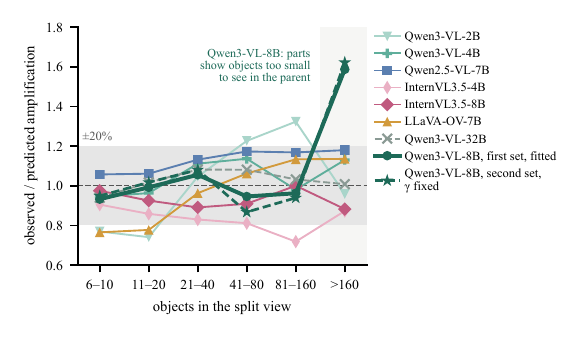}
\caption{Observed amplification of the count after one split, relative to the amplification $A$ that
Proposition~\ref{prop:split} predicts, by the number of objects in the split view. Ratio of the medians over the
split views of each bin; shaded band: within 20\% of the prediction.}
\label{fig:jensen}
\end{figure}

\subsubsection{Validation of the model}
Every part of the crops has a recorded answer, so the model can be checked at each split of the recursion of
Section~\ref{sec:method:count}. We compare the ratio of the parts' summed answers to the parent's answer with the amplification $A$, computed
from the true counts of the parts and the exponent of the same model in Table~\ref{tab:fit} (Fig.~\ref{fig:jensen}).
For Qwen3-VL-8B the two agree within 0.08 in every bin of parents from 6 to 160 objects. On its second set of crops,
with the exponent of the first set, they agree within 0.10 in four of five bins. For the seven other models
the observed amplification lies mostly within 20\% of the prediction (\appref{supp:model}). Empty parts are
answered with 0.8 to 2.4 spurious objects on average (column $s$ of Table~\ref{tab:fit}). Above 160 objects
the observed amplification is about 2.1, far above the predicted 1.3. Here the parts also reveal objects too small to
see in the parent, a limit of resolution that is separate from the undercount.

\begin{figure}[!tp]
\centering
\includegraphics[width=\linewidth]{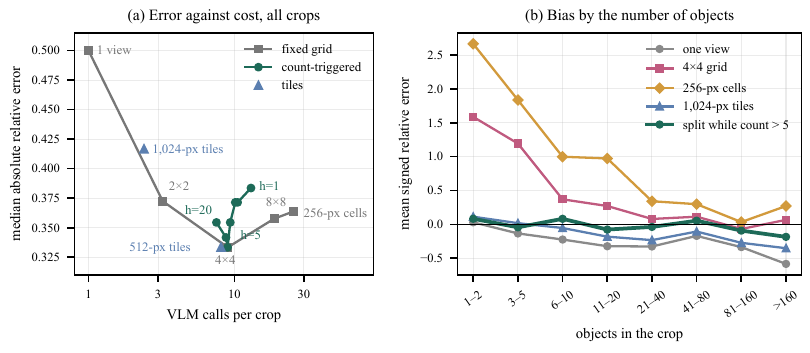}
\caption{Fixed grids and count-triggered subdivision on 937 DOTA crops (Qwen3-VL-8B). (a) Median absolute relative
error against VLM calls per crop: fixed grids (grey), tiles of bounded size (blue) and subdivision with $h$ from 20 to
1 (green). (b) Mean signed relative error by number of objects.}
\label{fig:grid}
\end{figure}

\subsubsection{Comparison with fixed grids}
Fixed grids, tiles of bounded size and subdivision with $h$ from 20 down to 1 are compared on the same answers of
Qwen3-VL-8B, given for every part of the 937 crops of both sets down to the 256-pixel cell (Fig.~\ref{fig:grid};
\appref{supp:model}). Over all crops a 4$\times$4 grid and subdivision at $h=5$ cost about the same, 9.0 and 9.4
calls per crop, and neither is significantly more accurate. They differ in where the errors and calls fall. On views
with more than 20 objects both about halve the median error of a single view, the grid slightly more and with fewer
calls. On views with
fewer objects, fixed grids overcount, the more so the finer they are. For one or two objects a 4$\times$4 grid answers
2.6 times the truth on average, while subdivision stays within 10\% of the truth at every number of objects up to 160
(Fig.~\ref{fig:grid}b). Up to 20 objects it spends 2.7 calls per crop,
against 7.3 for the grid. Tiles of at most 512 pixels are as accurate with 8.1 calls, but their number grows with the
area, to about 400 for a whole image of 10{,}000 pixels on a side. Table~\ref{tab:detectors} repeats the comparison on
the second set, with zeros from downsampled views also split, and Section~\ref{sec:exp:ablation} inside GeoScope.

\begin{table}[!tp]
\centering
\caption{Counting by detection against one view, a fixed grid and subdivision with Qwen3-VL-8B.}
\label{tab:detectors}
\footnotesize
\setlength{\tabcolsep}{4pt}
\begin{tabular}{lcccccc}
\toprule
 & & & & & \multicolumn{2}{c}{Signed} \\
\cmidrule(lr){6-7}
Method & Cost & Median & Within 20\% & WAPE & 1--5 & $>$160 \\
\midrule
\multicolumn{7}{l}{\emph{DOTA, second set, 458 crops}} \\
One view & 1.0 & 0.50 & 27 & 0.67 & $-$0.15 & $-$0.58 \\
Fixed 4$\times$4 & 9.0 & 0.30 & 37 & 0.33 & +1.36 & $-$0.12 \\
Subdivision as in GeoScope & 10.6 & 0.33 & 35 & 0.35 & $-$0.15 & $-$0.09 \\
MM-Grounding-DINO-L & 22.8 & 0.88 & 12 & 0.82 & +4.83 & $-$0.49 \\
OWLv2-L & 22.8 & 0.72 & 18 & 0.75 & +3.12 & +0.02 \\
LLMDet-L & 22.8 & 0.77 & 13 & 0.85 & +8.92 & $-$0.50 \\
\midrule
\multicolumn{7}{l}{\emph{FAIR1M cars, 279 crops}} \\
One view & 1.0 & 0.42 & 25 & 0.58 & +17.22 & $-$0.55 \\
Fixed 4$\times$4 & 3.2 & 0.24 & 44 & 0.36 & +2.52 & $-$0.19 \\
Subdivision as in GeoScope & 3.9 & 0.24 & 43 & 0.36 & +2.39 & $-$0.19 \\
MM-Grounding-DINO-L & 5.3 & 0.69 & 6 & 0.72 & +1.46 & $-$0.76 \\
OWLv2-L & 5.3 & 0.58 & 19 & 0.59 & +3.92 & +0.21 \\
LLMDet-L & 5.3 & 0.79 & 3 & 0.82 & +1.88 & $-$0.86 \\
\midrule
\multicolumn{7}{l}{\emph{DIOR ships, 262 crops}} \\
One view & 1.0 & 0.15 & 63 & 0.29 & $-$0.04 & $-$0.47 \\
Fixed 4$\times$4 & 3.0 & 0.11 & 70 & 0.16 & +0.10 & +0.09 \\
Subdivision as in GeoScope & 3.6 & 0.11 & 69 & 0.16 & $-$0.04 & +0.09 \\
MM-Grounding-DINO-L & 4.8 & 0.80 & 9 & 0.76 & +0.53 & $-$0.90 \\
OWLv2-L & 4.8 & 0.14 & 62 & 0.38 & +0.71 & $-$0.63 \\
LLMDet-L & 4.8 & 0.48 & 22 & 0.62 & +0.55 & $-$0.83 \\
\bottomrule
\end{tabular}
\par\smallskip
\begin{minipage}{\textwidth}\footnotesize
Cost: VLM calls per crop, or tiles per crop for the detectors, whose threshold is chosen on the first DOTA set. Median:
median absolute relative error. Within 20\%: share of crops counted within 20\% of the truth. WAPE: summed absolute
error over the summed true count. Signed: mean signed relative error at one to five and at more than 160 objects.
Subdivision as in GeoScope also splits zeros from downsampled
views, unlike the subdivision at $h=5$ of Fig.~\ref{fig:grid}.
\end{minipage}
\end{table}

\subsubsection{Comparison with open-vocabulary detectors}
Counting the boxes of three open-vocabulary detectors does worse than either the grid or subdivision
(Table~\ref{tab:detectors}). They count the same crops with the same class phrase, in tiles enlarged twofold, at a
threshold chosen on the first DOTA set~\citep{mmgdino,owlv2,llmdet}.
They overcount views with few objects. In views with more than 160 objects, two of them miss about half or more of
the objects on every set, and OWLv2 does so on DIOR ships. Only OWLv2 on DIOR ships, in small images, is level with
subdivision in median error, and its total error there is more than twice as large (\appref{supp:horizon:detectors}).

Splitting helps even when its leaves still hold more than $h$ objects. Nine in ten objects in our crops end in such
leaves, yet each split recovers the share of the undercount that Proposition~\ref{prop:split}
predicts. A fixed grid recovers as much on dense views but overcounts sparse ones, which most counting questions
concern (Section~\ref{sec:discussion:subdivision}). GeoScope therefore splits only views whose answer exceeds the
horizon or that answer zero at reduced resolution, and stops at a minimum cell (Section~\ref{sec:method:count}).

\section{GeoScope}
\label{sec:method}

\subsection{Overview}
\label{sec:method:overview}
Given a UHR image $I$, a question $q$ and answer options $\mathcal{O}$, GeoScope answers as
\begin{equation*}
\begin{aligned}
\pi &= \textsc{Synthesise}(q),\\
r &= \textsc{Execute}(\pi, I;\,V),\\
\hat o &= \textsc{Match}(r,\mathcal{O}).
\end{aligned}
\end{equation*}
A frozen, text-only language model writes a typed program $\pi$ from the question alone. A deterministic executor runs
it on the image, and the executed result $r$ is matched to the values of the options, not to their letters,
to give the answer $\hat o$.
Neither the synthesiser nor the VLM $V$ sees the options, and no parameter is updated. GeoScope takes from
GeoSelect~\citep{geoselect} the typed programs, their type checking and the stepwise fallbacks. What GeoScope adds is how the executor uses the
VLM, set by the measurements of Section~\ref{sec:horizon}. It asks the VLM only for local judgements: grounding a
phrase in a view where the target is visible, and counting a class in a view that is split while its count exceeds
the horizon. It chooses each view and its resolution, grounds a place before the objects inside it, and computes every
direction and distance from the grounded boxes. Fig.~\ref{fig:pipeline}a shows the three stages, and Fig.~\ref{fig:pipeline}b and c two test questions. For
the counting question the program
fixes the lower-right part of the image, and the executor splits it while the VLM's count exceeds five. For the relation
question it grounds the residential area, then both objects inside the area's window, and computes the direction
from their boxes. The backbone's own answer is wrong in both cases.

\subsection{Typed programs}
\label{sec:method:program}
A program is an expression tree over four types: sets of boxes, image regions, counts and directions. Its abstract
syntax is
\begin{center}
\footnotesize
\begin{tabular}{@{}>{\ttfamily}l@{\ }>{\ttfamily}c@{\ }>{\ttfamily}l@{}}
prog   & ::= & count(set) | direction(set, set) \\
set    & ::= & detect(class) | ref | in\_region(set, region) \\
       & |   & filter(set, dir, set) | near(set, set) | top1(set) \\
       & |   & argmax(set, key) | argmin(set, key) \\
ref    & ::= & refer(expr [, class] [, plural] [, within=(ref | part(dir))]) \\
region & ::= & image | red\_circle | part(dir) | around(set) \\
dir    & ::= & TOP | BOTTOM | LEFT | RIGHT | UL | UR | LL | LR | CENTER \\
key    & ::= & X | Y | AREA | LENGTH
\end{tabular}
\end{center}
A counting question becomes a \texttt{count} over a set. The set is described by a short class phrase that keeps the question's
colour and shape words, such as \texttt{red car}, and is restricted to a region of the image or to the surroundings of
an anchor object. A relation question becomes a \texttt{direction} between two referents. Each referent keeps the
object's own referring expression, its class, whether it denotes several objects and, when the question names one, the place or image part that contains
it. The synthesiser writes the program from this grammar and a few worked examples, and a checker rejects programs
that are not well typed. The full grammar, in the JSON form that the synthesiser writes, and the evaluation of each
operator are given in \appref{supp:method:grammar}. For MME-RealWorld counting, a program that the checker rejects or that drops
a colour, shape, place or position word of the question is written once more (\appref{supp:method:prompts}).

\subsection{Perception by a frozen VLM}
\label{sec:method:perception}
The VLM is called only to ground a phrase in a window, returning boxes, or to count a class in a window, returning an
integer. Every call renders its window at a fixed pixel budget $B$, so a large window is downsampled; we call the
rendered window a view. Grounding in a
large window is therefore coarse to fine, and candidate boxes found in the downsampled window are confirmed in
close-ups at near-native resolution. A referring expression is grounded as a whole, so that the VLM resolves its
attributes and local relations, and the box is then refined with the class phrase.

Small objects are often named through the place that contains them, as in ``the grey tennis court in the residential
area right of centre''. The place is visible on the whole image when the object is not, so GeoScope resolves such a
referent hierarchically. It grounds the place on the whole image and then the object inside a window around the place,
which the same budget renders at a much higher resolution.

\subsection{Count-triggered subdivision}
\label{sec:method:count}
A counting program fixes a region, and the VLM counts the class inside it. Enumerating boxes degenerates in dense
views, so GeoScope counts grounded boxes only for sets that the executor cannot turn into a class phrase and one
window, mostly a \texttt{filter} or \texttt{near} nested inside a region. This happens in 51 of the 962
MME-RealWorld counting programs executed on the test split (\appref{supp:method:ladder}).
Algorithm~\ref{alg:horizon} applies the rules of Section~\ref{sec:horizon:theory}. A view's count is accepted if it
lies within the horizon $h$. The one exception is a zero from a downsampled view, since objects too small to see at a
coarse scale may still be present. Any other view is split into four parts, or two if it is elongated. Each part is
counted in the same way down to a minimum cell, and the counts of the leaves, the parts not split further, are
summed. In
Algorithm~\ref{alg:horizon}, $|w|$ is the area of window $w$ in pixels, so the view is downsampled when $\zeta>1$;
$\mathrm{side}(w)$ is its longer side, $d$ its depth in the recursion, and \textsc{Split} returns the four or two
parts.

We use one threshold, $h=5$, with every backbone. It is the upper end of the last bin before the shortfall of most
models in Table~\ref{tab:fit}, read off the measurement rather than tuned. For Qwen3-VL-8B the exact value matters
little. On the DOTA crops of Section~\ref{sec:horizon}, $h=3$, 5 and 8 give median absolute relative errors of 0.37,
0.35 and 0.33 (\appref{supp:model}), and on development counting questions of MME-RealWorld, in an earlier
configuration that still accepted zeros from downsampled views, 40.4, 41.2 and 41.2\% accuracy
(\appref{supp:results:budget}). For InternVL3.5-8B, whose shortfall on DOTA begins beyond
40 objects, $h=40$ is not better on UHR-Micro (Section~\ref{sec:discussion:imagery}). Once zeros from downsampled
views are split, few views with more than ten objects are left unsplit (\appref{supp:model}). The regions of XLRS-Bench RC are small drawn circles, so their crop is counted in a single view.

\begin{algorithm}[t]
\caption{Count-triggered subdivision for counting class $\kappa$ in window $w$, with horizon $h$, depth limit $D$ and
minimum cell $\mu$.}
\label{alg:horizon}
\begin{algorithmic}[1]
\Function{HCount}{$w, \kappa, d$}
  \State $\hat n \gets \textsc{Count}(w, \kappa)$; \ $\zeta \gets \sqrt{|w|/B}$
  \If{$\big(\hat n \le h \wedge \neg(\hat n = 0 \wedge \zeta > 1)\big) \vee d = D \vee \mathrm{side}(w) < 2\mu$}
    \State \Return $\hat n$
  \EndIf
  \State \Return $\sum_{w' \in \textsc{Split}(w)} \textsc{HCount}(w', \kappa, d+1)$
\EndFunction
\end{algorithmic}
\end{algorithm}

\subsection{Geometry, option matching and fallbacks}
\label{sec:method:projection}
The direction of one box relative to another is computed from the direction-relation matrix~\citep{cdc}. It tiles
the plane around the reference box and weights each tile's direction by how much of the target falls in it. Unlike
the angle between box centres, it respects the extent of objects, so an object directly above a wide building is above
it even if its centre lies left of the building's centre. A count selects the option value with the highest Poisson
likelihood when the count is taken as the mean, and a direction selects the option closest in angle. On RC, a count
that equals no option value is taken twice more, from a wider and from an enlarged view, and Match uses the median of
the three. The VLM still never sees the options, and RC accuracy is the same without this step
(\appref{supp:method:ladder}). Execution degrades in steps
instead of failing. If an anchor or place of a counting
program is not found, it is searched for again without its colour and shape words and then also without its own
constraints. A referent of a relation question that yields no box is grounded by its class phrase. As a last
resort the question receives the VLM's own answer (\appref{supp:method:ladder}). With Qwen3-VL-8B, more than 95\% of
the RC, OSR and MME-RealWorld counting questions are answered by an executed program.

Some questions are answered without a synthesised program. A question that gives a pixel box is answered by a fixed
program. XLRS-Bench gives one for the category, colour and motion state of an object, and the VLM answers from the
whole image together with a close-up in which the box is outlined. UHR-Micro gives one for the region of a regional
counting question, which is then counted as in Section~\ref{sec:method:count}. It marks the two objects of its
direction and distance questions with coloured rectangles, from whose boxes the relation is computed
(Section~\ref{sec:exp:umicro}). Position and colour questions of MME-RealWorld are answered by the VLM directly.

\section{Experimental setup}
\label{sec:exp}

\subsection{Benchmarks and protocol}
\label{sec:exp:setup}
We evaluate on three benchmarks (Table~\ref{tab:benchmarks}). From the lite release of XLRS-Bench~\citep{xlrsbench} we
use regional counting (RC), about regions drawn as red circles, and object spatial relationship (OSR); the other
subtasks, among them overall counting of ships or cars in a whole scene, enter its overall score. From MME-RealWorld-RS (MME-RS), the remote sensing part of
MME-RealWorld~\citep{mmerealworld}, GeoScope answers the counting questions and its backbone the position and colour
questions. UHR-Micro~\citep{uhrmicro} has its own splits (Section~\ref{sec:exp:umicro}).

\begin{table}[!tp]
\centering
\caption{Benchmarks: development and test questions and the typical long side of the images (pixels).}
\label{tab:benchmarks}
\footnotesize
\setlength{\tabcolsep}{4pt}
\resizebox{\textwidth}{!}{%
\begin{tabular}{llrrrl}
\toprule
Benchmark & Task & Dev. & Test & Image side & Image sources \\
\midrule
XLRS-Bench (lite) & RC & 20 & 80 & 10{,}000 & DOTA, MiniFrance, HRSCD, ITCVD \\
 & OSR & 100 & 400 & 10{,}000 & \\
 & all 13 subtasks & \multicolumn{2}{c}{3{,}080} & 10{,}000 & \\
MME-RealWorld-RS & counting & 245 & 981 & 6{,}000 & DOTA, Potsdam, Toronto \\
 & position, colour & 502 & 2{,}010 & 6{,}000 & \\
UHR-Micro & five task types & 500 & 1{,}369 & 4{,}800 & DOTA-v2, FAIR1M, SODA-A, xView \\
\bottomrule
\end{tabular}}
\end{table}

All design choices are made on a seeded 20\% development split of each task; the configuration is frozen before any
test evaluation, and the remaining 80\% form the held-out test split. Gains are paired on the same questions, with 95\% bootstrap
intervals over questions~\citep{bootstrap}.

\subsection{Baselines}
\label{sec:exp:baselines}
The zero-shot baseline is the frozen VLM answering the multiple-choice question from the whole image at 4\,M pixels,
taking the option letter with the highest first-token score. With this protocol Qwen3-VL-8B reaches 74.46, 63.98
and 25.45 on all MME-RealWorld position, colour and counting questions, against 56.41, 48.07 and 21.35 reported
elsewhere~\citep{weaveearth,gazeearth,rsopsd}. Our zero-shot reference is stronger mainly because the image is given at a
larger size (\appref{supp:protocol:zeroshot}), and GazeEarth's own direct baseline, run on our test questions, comes
within 3.0 points of it on position (\appref{supp:repro:controls}). MME-RealWorld counting questions also offer the option stating that the image
does not show the number. It is correct for 5 of the 1{,}226 questions, yet the four backbones we ran zero-shot choose it for 18
to 57\% of the test questions. We therefore also report each backbone restricted to the numeric options, answering
with the numeric option of highest first-token score. This raises Qwen3-VL-8B from 25.45 to 35.64 on all counting
questions.

The constant-answer baseline always picks the option position or option-value rank, such as option B or the largest
number, that is most often correct on the development split, with ties broken in its favour on the test split
(\appref{supp:protocol:constant}).

\subsection{Reproduced methods}
\label{sec:exp:repro}
For every task we reproduce, with its official code, the training-free method whose reported result is the strongest:
WeaveEarth~\citep{weaveearth} for RC, GazeEarth~\citep{gazeearth} for OSR and XLRS-Bench as a whole, and
ZoomSearch~\citep{zoomsearch} and RAP~\citep{rap} for MME-RealWorld counting. Each method runs its own pipeline,
including its own answer extraction, on our test questions, and its gain is measured against the zero-shot answers of
its own backbone. The backbone is Qwen3-VL-8B for WeaveEarth, GazeEarth and RAP, and LLaVA-OV-7B for ZoomSearch and for a
second run of RAP on a fixed 20\% subset of the counting questions. Their settings are given in \appref{supp:repro}. Where other
papers have re-run these methods on our tasks, their numbers are mostly close to ours or lower, and lower than the
numbers the methods report for themselves (\appref{supp:repro:numbers}). We also re-run one trained method, Text-Before-Vision~\citep{textbeforevision},
with its released checkpoint and code on all XLRS-Bench questions (\appref{supp:repro:tbv}). It is the only published
method that reports a higher XLRS-Bench average than GeoScope.

\subsection{Implementation details}
\label{sec:exp:impl}
GeoScope uses Qwen3-4B~\citep{qwen3} as the synthesiser and Qwen3-VL-8B-Instruct~\citep{qwen3vl} as the perception
model, both decoding greedily. The pixel budget $B$ is 2.5\,M pixels, 1\,M pixels for RC crops and 4\,M pixels for
XLRS-Bench overall counting. Counting uses $h=5$, a depth limit $D=5$ and a minimum cell of $\mu=256$ pixels. A counted window
is enlarged by 15\% on each side, since grounded places are coarse and objects on their border would otherwise be cut; this
does not bias the counts upward (Section~\ref{sec:discussion:subdivision}). A window longer than 7{,}000 pixels is first cut into equal cells; the
images of XLRS-Bench overall counting are cut into cells of 2{,}000 pixels. Every experiment runs on one NVIDIA GeForce RTX 3090 with 24\,GiB of
memory, except the re-run of Text-Before-Vision, which its script serves on two GPUs, and Qwen3-VL-32B, which we query
through an API.

\section{Results}
\label{sec:results}

\subsection{Comparison with reproduced training-free methods}
\label{sec:exp:main}

\begin{table}[!tp]
\centering
\caption{Test accuracy (\%) of GeoScope and the reproduced training-free methods. $\Delta$: paired gain over each
method's own zero-shot backbone, with its 95\% bootstrap interval below.}
\label{tab:main}
\footnotesize
\setlength{\tabcolsep}{3pt}
\resizebox{\textwidth}{!}{%
\begin{tabular}{lllcccccc}
\toprule
 & & & \multicolumn{2}{c}{RC (80)} & \multicolumn{2}{c}{OSR (400)} & \multicolumn{2}{c}{MME-RS count (981)} \\
\cmidrule(lr){4-5}\cmidrule(lr){6-7}\cmidrule(lr){8-9}
Method & Venue & Backbone & Acc. & $\Delta$ & Acc. & $\Delta$ & Acc. & $\Delta$ \\
\midrule
Zero-shot & -- & Qwen3-VL-8B & 47.50 & -- & 33.50 & -- & 25.18 & -- \\
\ numeric options only$^{\mathrm{a}}$ & -- & Qwen3-VL-8B & -- & -- & -- & -- & 35.47 & -- \\
Zero-shot & -- & LLaVA-OV-7B & 35.00 & -- & 25.00 & -- & 32.31 & -- \\
\ numeric options only$^{\mathrm{a}}$ & -- & LLaVA-OV-7B & -- & -- & -- & -- & 37.82 & -- \\
Constant answer$^{\mathrm{b}}$ & -- & -- & 37.50 & -- & 30.25 & -- & 31.50 & -- \\
\midrule
WeaveEarth~\citep{weaveearth}$^{\mathrm{c}}$ & MM'26 & Qwen3-VL-8B & 45.00 & \dci{-2.5}{-10.0}{+5.0} & 32.75 & \dci{-0.8}{-4.8}{+3.2} & 30.48 & \dci{+5.3}{+2.5}{+8.2} \\
GazeEarth~\citep{gazeearth}$^{\mathrm{c}}$ & arXiv'26 & Qwen3-VL-8B & 47.50 & \dci{0.0}{-10.0}{+10.0} & 30.75 & \dci{-2.8}{-7.8}{+2.2} & 27.73 & \dci{+2.5}{-0.8}{+6.0} \\
ZoomSearch~\citep{zoomsearch}$^{\mathrm{c,d}}$ & arXiv'25 & LLaVA-OV-7B & 42.50 & \dci{+7.5}{-2.5}{+18.8} & 32.50 & \dci{+7.5}{+3.2}{+11.8} & 31.50 & \dci{-0.8}{-4.4}{+2.8} \\
RAP~\citep{rap}$^{\mathrm{c}}$ & ICML'25 & Qwen3-VL-8B & -- & -- & -- & -- & 30.38 & \dci{+5.2}{+2.3}{+8.1} \\
RAP~\citep{rap}$^{\mathrm{c,e}}$ & ICML'25 & LLaVA-OV-7B & -- & -- & -- & -- & 35.53 & \dci{+5.6}{-0.5}{+11.7} \\
\midrule
GeoScope & -- & Qwen3-VL-8B & \overall \best{63.75} & \dci{+16.3}{+3.8}{+28.7} & \overall \best{44.00} & \dci{+10.5}{+4.2}{+16.8} & \second{36.60} & \dci{+11.4}{+7.7}{+15.1} \\
GeoScope & -- & LLaVA-OV-7B & \overall \best{63.75} & \dci{+28.8}{+17.5}{+40.0} & 31.00 & \dci{+6.0}{+1.8}{+10.2} & 34.56 & \dci{+2.2}{-1.2}{+5.9} \\
\midrule
Zero-shot & -- & Qwen3-VL-4B & 40.00 & -- & 26.50 & -- & 14.27 & -- \\
\ numeric options only$^{\mathrm{a}}$ & -- & Qwen3-VL-4B & -- & -- & -- & -- & 27.52 & -- \\
GeoScope & -- & Qwen3-VL-4B & \second{60.00} & \dci{+20.0}{+7.5}{+32.5} & \second{40.75} & \dci{+14.3}{+8.8}{+19.8} & \overall \best{36.80} & \dci{+22.5}{+19.2}{+25.9} \\
\midrule
Zero-shot & -- & Qwen2.5-VL-7B & 41.25 & -- & 36.50 & -- & 18.65 & -- \\
\ numeric options only$^{\mathrm{a}}$ & -- & Qwen2.5-VL-7B & -- & -- & -- & -- & 32.52 & -- \\
GeoScope & -- & Qwen2.5-VL-7B & 52.50 & \dci{+11.3}{+1.2}{+21.2} & 40.25 & \dci{+3.8}{-1.5}{+9.2} & 34.45 & \dci{+15.8}{+12.4}{+19.3} \\
\bottomrule
\end{tabular}}
\par\smallskip
\begin{minipage}{\textwidth}\footnotesize
\rankkeytf{}; zero-shot rows, constant answers and the run on a subset are not ranked. $^{\mathrm{a}}$Backbone restricted to
the numeric options, which applies only to MME-RealWorld counting (Section~\ref{sec:exp:baselines}). $^{\mathrm{b}}$An upper
bound (Section~\ref{sec:exp:baselines}). $^{\mathrm{c}}$Run by us with the official code. $^{\mathrm{d}}$On RC, GeoScope with LLaVA-OV-7B is 21.3 points [8.8, 33.8] above ZoomSearch, paired by question. $^{\mathrm{e}}$On a fixed 20\% subset of the counting questions.
\end{minipage}
\end{table}

Against its own backbone on the same test questions (Table~\ref{tab:main}), GeoScope improves
Qwen3-VL-8B by 16.3 points on RC, 10.5 on OSR and 11.4 on MME-RealWorld counting, all significantly. No reproduced
method gains as much on any task. WeaveEarth and GazeEarth stay close to their backbone on RC and OSR, and WeaveEarth
gains 5.3 points on counting. ZoomSearch improves OSR but not counting, and RAP gains 5.2 points on counting. RAP with LLaVA-OV-7B was run only on a fixed
20\% subset of the counting questions, where GeoScope scores 29.95 with Qwen3-VL-8B and 30.96 with LLaVA-OV-7B, against 35.53 for RAP.
MME-RealWorld counting also rewards a fixed guess. Always choosing option B beats zero-shot Qwen3-VL-8B, no reproduced
method run on the whole split exceeds it, and GeoScope exceeds it by 5.1 points
(Section~\ref{sec:discussion:subdivision}).

GeoScope also gains on the other three backbones, though not significantly on OSR with Qwen2.5-VL-7B or on counting
with LLaVA-OV-7B. The gains grow with the weaker Qwen3-VL-4B, which matches the 8B system on counting and comes within
four points of it on RC and OSR. On LLaVA-OV-7B, the
backbone of ZoomSearch and RAP, GeoScope is significantly above ZoomSearch on RC, by 21.3 points, and level with it
on OSR and counting.

\begin{table}[!tp]
\centering
\caption{MME-RealWorld counting test results by the position of the correct number among the options.}
\label{tab:mmeoptions}
\footnotesize
\setlength{\tabcolsep}{4pt}
\begin{tabular}{llccccc}
\toprule
Method & Backbone & Absent & Largest & Other & All & $\Delta$ to numeric \\
\midrule
Constant answer B & -- & 0 & 19.4 & 35.0 & 31.5 & -- \\
Zero-shot & Qwen3-VL-8B & 35 & 5.4 & 31.0 & 25.2 & -- \\
\ numeric options only & Qwen3-VL-8B & 0 & 7.2 & \overall \best{43.7} & 35.5 & -- \\
Zero-shot & Qwen3-VL-4B & 57 & 5.0 & 17.0 & 14.3 & -- \\
\ numeric options only & Qwen3-VL-4B & 0 & 7.7 & 33.3 & 27.5 & -- \\
Zero-shot & LLaVA-OV-7B & 18 & 29.3 & 33.2 & 32.3 & -- \\
\ numeric options only & LLaVA-OV-7B & 0 & 33.8 & \second{39.0} & \overall \best{37.8} & -- \\
Zero-shot & Qwen2.5-VL-7B & 45 & 10.4 & 21.1 & 18.7 & -- \\
\ numeric options only & Qwen2.5-VL-7B & 0 & 17.1 & 37.0 & 32.5 & -- \\
\midrule
WeaveEarth & Qwen3-VL-8B & 12 & 3.6 & 38.3 & 30.5 & \dci{-5.0}{-7.8}{-2.1} \\
GazeEarth & Qwen3-VL-8B & 18 & 23.4 & 29.0 & 27.7 & \dci{-7.7}{-11.5}{-3.9} \\
RAP & Qwen3-VL-8B & 17 & 10.4 & 36.2 & 30.4 & \dci{-5.1}{-8.2}{-2.0} \\
ZoomSearch & LLaVA-OV-7B & 9 & 19.4 & 35.0 & 31.5 & \dci{-6.3}{-9.9}{-2.9} \\
\midrule
GeoScope & Qwen3-VL-8B & 1 & \overall \best{58.1} & 30.3 & 36.6 & \dci{+1.1}{-2.8}{+5.0} \\
GeoScope & Qwen3-VL-4B & 3 & \second{49.1} & 33.2 & \second{36.8} & \dci{+9.3}{+5.8}{+12.7} \\
GeoScope & LLaVA-OV-7B & 2 & 42.3 & 32.3 & 34.6 & \dci{-3.3}{-6.8}{+0.3} \\
GeoScope & Qwen2.5-VL-7B & 2 & 43.7 & 31.8 & 34.5 & \dci{+1.9}{-1.6}{+5.6} \\
\bottomrule
\end{tabular}
\par\smallskip
\begin{minipage}{\textwidth}\footnotesize
Absent: share (\%) of answers stating that the image does not show the number. Largest, Other: accuracy (\%) on the
222 questions whose answer is the largest number offered and on the remaining 759; All: on all 981. $\Delta$ to
numeric: paired difference to the backbone restricted to the numeric options, with its 95\% interval. \rankkeytf{}
(Largest, Other, All); here every row is ranked.
\end{minipage}
\end{table}

Table~\ref{tab:mmeoptions} splits the counting questions by where the correct number sits among the options, leaving
out the RAP run on part of the split. For 222 questions the answer is the largest number offered. A count that falls
short picks a smaller number there. GeoScope with Qwen3-VL-8B answers 58.1\% of these questions correctly, against
5.4\% for its zero-shot backbone and at most 23.4\% for a reproduced method.
The other 759 questions mostly ask about at most five objects. On them the reproduced methods and GeoScope all score
between 29 and 39, and Qwen3-VL-8B restricted to the numeric options does best. Against its backbone restricted in this way,
GeoScope therefore gains only on the questions whose answer is the largest number offered, with every backbone; part
of that gain is owed to the layout (Section~\ref{sec:discussion:subdivision}).

\subsection{Comparison with published results}
\label{sec:exp:published}

\begin{table}[!tp]
\centering
\caption{Published results (\%) on all questions of XLRS-Bench and MME-RealWorld-RS, with our re-run of
Text-Before-Vision.}
\label{tab:published}
\footnotesize
\setlength{\tabcolsep}{3pt}
\resizebox{\textwidth}{!}{%
\begin{tabular}{lllcccccccc}
\toprule
 & & & \multicolumn{4}{c}{XLRS-Bench} & \multicolumn{4}{c}{MME-RealWorld-RS} \\
\cmidrule(lr){4-7}\cmidrule(lr){8-11}
Method & Venue & Backbone & RC & OSR & Avg. & By q. & Position$^{\mathrm{j}}$ & Colour$^{\mathrm{j}}$ & Count & Avg. \\
\midrule
\multicolumn{11}{l}{\emph{Trained}} \\
GeoChat~\citep{geochat}$^{\mathrm{a}}$ & CVPR'24 & Vicuna-7B & 29.0 & 24.2 & 22.9 & -- & -- & -- & -- & -- \\
EarthDial~\citep{earthdial}$^{\mathrm{b}}$ & CVPR'25 & Phi-3-mini & 42.0 & 24.8 & 40.3 & 33.8 & -- & -- & -- & -- \\
GeoLLaVA-8K~\citep{geollava8k} & NeurIPS'25 & -- & 38.0 & 35.0 & 51.5 & -- & 34.90$^{\mathrm{c}}$ & 27.92$^{\mathrm{c}}$ & 22.27$^{\mathrm{c}}$ & 28.36$^{\mathrm{c}}$ \\
ZoomEarth~\citep{zoomearth} & CVPR'26 & Qwen2.5-VL-3B & 35.0$^{\mathrm{d}}$ & 20.8$^{\mathrm{d}}$ & 39.2$^{\mathrm{d}}$ & -- & 46.6$^{\mathrm{d}}$ & 41.9$^{\mathrm{d}}$ & 27.6$^{\mathrm{d}}$ & 38.7$^{\mathrm{d}}$ \\
GeoEyes~\citep{geoeyes} & arXiv'26 & Qwen2.5-VL-7B & 40.0 & 32.2 & 54.2 & -- & -- & -- & -- & -- \\
DualComp~\citep{dualcomp} & ECCV'26 & -- & 45.0 & 37.4 & 53.1 & -- & -- & -- & -- & -- \\
GeoLens~\citep{geolens} & arXiv'26 & Qwen2.5-VL-7B & 40.0 & 36.8 & 54.2 & -- & 63.64$^{\mathrm{e}}$ & 53.23$^{\mathrm{e}}$ & 19.74$^{\mathrm{e}}$ & 45.54$^{\mathrm{e}}$ \\
GeoVista~\citep{geovista} & arXiv'26 & Qwen2.5-VL-7B & -- & 42.40 & -- & 50.65 & -- & -- & -- & -- \\
Text-Before-Vision~\citep{textbeforevision}$^{\mathrm{f}}$ & ICML'26 & Qwen2.5-VL-7B & -- & -- & \overall 60.40 & -- & -- & -- & -- & -- \\
\ our re-run$^{\mathrm{g}}$ & -- & Qwen2.5-VL-7B & 40.00 & 28.40 & 50.54 & 43.15 & -- & -- & -- & -- \\
UHR-BAT~\citep{uhrbat} & ICML'26 & LongVA-7B & 33.0 & 44.8 & 48.6 & 44.0 & -- & -- & -- & -- \\
MLRS~\citep{mlrs} & ACCV'26 & InternVL3.5-8B & 39.0 & 31.8 & 51.3 & -- & -- & -- & -- & -- \\
RS-OPSD~\citep{rsopsd} & arXiv'26 & Qwen3-VL-8B & 55.0 & 29.2 & 53.1 & -- & \overall 76.8 & \overall 70.4 & 37.4 & \overall 61.5 \\
ScaleEarth~\citep{scaleearth}$^{\mathrm{h}}$ & arXiv'26 & Qwen3-VL-8B & 54.3 & 40.9 & 59.4 & -- & -- & -- & -- & -- \\
\midrule
\multicolumn{11}{l}{\emph{Training-free}} \\
ZoomEye~\citep{zoomeye}$^{\mathrm{c}}$ & EMNLP'25 & LLaVA-OV-7B & -- & -- & -- & -- & 43.52 & 60.88 & 30.10 & 44.83 \\
RAP~\citep{rap}$^{\mathrm{c}}$ & ICML'25 & LLaVA-OV-7B & -- & -- & -- & -- & 57.62 & \second{64.53} & \overall \best{40.25} & 54.13 \\
ZoomSearch~\citep{zoomsearch} & arXiv'25 & LLaVA-OV-7B & 42.0$^{\mathrm{d}}$ & 32.2$^{\mathrm{d}}$ & 48.0$^{\mathrm{d}}$ & -- & 67.62 & \best{66.14} & \second{39.15} & \second{57.64} \\
WeaveEarth~\citep{weaveearth} & MM'26 & Qwen3-VL-8B & 47.0$^{\mathrm{d}}$ & 36.40$^{\mathrm{i}}$ & 49.5$^{\mathrm{d}}$ & 47.14 & 62.60 & 51.39 & 27.65 & 47.21 \\
GazeEarth~\citep{gazeearth} & arXiv'26 & Qwen3-VL-8B & -- & 37.80$^{\mathrm{i}}$ & -- & 53.51 & \second{71.12} & 63.67 & 21.78 & 52.19 \\
\midrule
\multicolumn{11}{l}{\emph{Ours}} \\
GeoScope, all questions & -- & Qwen3-VL-8B & \overall \best{63.00} & \overall \best{46.40} & \best{59.98} & \overall \best{71.82} & \best{74.46} & 63.98 & 37.93 & \best{58.79} \\
GeoScope, held-out 80\% & -- & Qwen3-VL-8B & 63.75 & 44.00 & 59.53 & 71.51 & 74.65 & 63.55 & 36.60 & 58.26 \\
GeoScope, all questions & -- & Qwen2.5-VL-7B & 51.00 & \second{40.00} & \second{56.39} & \second{65.26} & 67.94 & 49.96 & 34.18 & 50.69 \\
GeoScope, all questions & -- & LLaVA-OV-7B & \second{57.00} & 31.40 & 52.11 & 62.53 & 62.45 & 58.80 & 34.67 & 51.97 \\
\bottomrule
\end{tabular}}
\par\smallskip
\begin{minipage}{\textwidth}\footnotesize
Avg.: mean over the 13 subtasks of XLRS-Bench and the 3 of MME-RealWorld-RS; By q.: accuracy over all 3{,}080
XLRS-Bench questions. \rankkey{}; the held-out row and our re-run are not ranked. Values from other papers:
$^{\mathrm{a}}$GeoLLaVA-8K~\citep{geollava8k}, mean over 12 subtasks; $^{\mathrm{b}}$UHR-BAT~\citep{uhrbat}; $^{\mathrm{c}}$ZoomSearch~\citep{zoomsearch}; $^{\mathrm{d}}$RS-OPSD~\citep{rsopsd};
$^{\mathrm{e}}$GazeEarth~\citep{gazeearth}. $^{\mathrm{f}}$Mean of 32 sampled answers under its own scoring code. $^{\mathrm{g}}$Run by us with the official
code; one greedy answer per question. $^{\mathrm{h}}$Split not stated. $^{\mathrm{i}}$From a question-weighted table. $^{\mathrm{j}}$For GeoScope, the zero-shot answers of its backbone.
\end{minipage}
\end{table}

Published numbers cover all questions of each task, so Table~\ref{tab:published} gives GeoScope on all questions as
well as on the held-out 80\%. GeoChat~\citep{geochat}, EarthDial~\citep{earthdial}, GeoLLaVA-8K~\citep{geollava8k},
UHR-BAT~\citep{uhrbat} and ScaleEarth~\citep{scaleearth} are fine-tuned. GeoEyes~\citep{geoeyes},
GeoLens~\citep{geolens}, GeoVista~\citep{geovista}, ZoomEarth~\citep{zoomearth} and
Text-Before-Vision~\citep{textbeforevision} add reinforcement learning, and MLRS~\citep{mlrs} uses reinforcement
learning alone. RS-OPSD~\citep{rsopsd} is self-distilled. DualComp~\citep{dualcomp} trains a router that sets how
GeoLLaVA-8K's tokens are compressed, so we list it as trained. The training-free methods are
ZoomEye~\citep{zoomeye}, RAP~\citep{rap}, ZoomSearch, WeaveEarth and GazeEarth. Where a paper averages otherwise, we
average its reported subtasks. GeoScope answers the position and colour questions of MME-RealWorld
with its backbone's zero-shot answers.

GeoScope has the highest reported accuracy on RC among trained and training-free methods, 63.00 against 55.0 for
RS-OPSD. On OSR it reaches 46.40 on all questions and 44.00 on the held-out ones, against 44.8 for UHR-BAT, the best
trained model, and 37.80 for GazeEarth, the best training-free method. GeoEyes, GeoLens and GeoVista
fine-tune Qwen2.5-VL-7B. Without training, GeoScope with this backbone is above GeoEyes and GeoLens on both tasks and
below GeoVista on OSR.

On XLRS-Bench as a whole, programs answer RC, OSR and overall counting. Questions that give a pixel box are answered
as in Section~\ref{sec:method:projection}, and the backbone answers the other seven subtasks. GeoScope then reaches
59.98 averaged over the 13 subtasks, above every published training-free result, and 71.82 over all 3{,}080
questions, the highest reported by question for any method, trained or not.
The close-ups of given pixel boxes add 7.1 of its 11.2 points over the backbone on the average over subtasks and 22.8
of the 26.2 points by question, and the programs for RC, OSR and overall counting add the rest
(\appref{supp:results:xlrs}).
GazeEarth reports 53.51 by question. Run with its official code on our held-out questions, it answers 39.61\% of them
correctly, against 46.71\% for zero-shot Qwen3-VL-8B and 71.51\% for GeoScope. It loses most on questions about
the object in a given box (\appref{supp:repro:numbers}). GeoScope's average is also above that of every trained
method that reports its subtasks. ScaleEarth reports 59.4~\citep{scaleearth}, and GeoLens
and GeoEyes report 54.2~\citep{geolens,geoeyes}.

Only the trained Text-Before-Vision~\citep{textbeforevision} reports a higher average, 60.40, the mean of 32
sampled answers per question, without results per subtask. Table~\ref{tab:published} therefore also lists our re-run
of its released checkpoint and code, scored like every other method with one answer per question, with greedy and
with sampled answers (\appref{supp:repro:tbv}). On the held-out questions the two re-runs reach 50.54 and 48.15, and
GeoScope, at 59.53, is 9.0 and 11.4 points above them. With Qwen2.5-VL-7B, the model Text-Before-Vision is trained
from, GeoScope reaches 56.93 on these questions, 6.4 and 8.8 points above, and all four differences are significant.

On MME-RealWorld, GeoScope's average is the highest among published training-free methods, mostly through the
backbone's position accuracy, and below the trained RS-OPSD. On counting, its 37.93 on all questions is above the 37.4
of RS-OPSD. It is below the 39.15 and 40.25 that ZoomSearch reports for itself and for RAP, both with
LLaVA-OV-7B. With that backbone GeoScope reaches 34.67 on all questions and 34.56 on our test split, where
ZoomSearch, re-run with its official code, reaches 31.50. Two later papers obtain 20.1 and
17.9~\citep{weaveearth,gazeearth}, and on RC and OSR our reproduction agrees with the values measured by
RS-OPSD~\citep{rsopsd} (\appref{supp:repro:numbers}). With
LLaVA-OV-7B, GeoScope is above ZoomSearch on RC and on the average over subtasks, and level on OSR. Its
MME-RealWorld average with that backbone is lower than those of ZoomSearch and RAP, because its position and colour
answers are the backbone's own.

\subsection{Results on UHR-Micro}
\label{sec:exp:umicro}

\begin{table}[!tp]
\centering
\caption{UHR-Micro test scores (\%, official metric) on the five task types that GeoScope covers.}
\label{tab:umicro}
\footnotesize
\setlength{\tabcolsep}{4pt}
\resizebox{\textwidth}{!}{%
\begin{tabular}{lcccccc}
\toprule
Method & Backbone & Global (304) & Regional (544) & Conditional (202) & Direction (160) & Distance (159) \\
\midrule
\multicolumn{7}{l}{\emph{Zero-shot}} \\
Zero-shot & Qwen2-VL-7B & 23.06 & 11.67 & 27.83 & 52.50 & 37.74 \\
Zero-shot & Qwen2.5-VL-7B & 26.44 & 18.41 & 39.30 & 63.12 & 50.31 \\
Zero-shot & Qwen3-VL-8B & 20.37 & 12.08 & 15.96 & 68.12 & 70.44 \\
\ our re-run & Qwen3-VL-8B & 23.71 & 13.13 & 23.45 & 65.63 & 71.70 \\
Zero-shot & Qwen3-VL-32B & 20.70 & 12.75 & 13.67 & 76.88 & 76.10 \\
Zero-shot & InternVL3.5-8B & 35.33 & 25.42 & 23.88 & 45.00 & 37.74 \\
\ our re-run & InternVL3.5-8B & 33.99 & 21.03 & 26.31 & 43.75 & 37.11 \\
Zero-shot & Qwen3-VL-235B-A22B & 26.64 & 14.32 & 16.12 & 87.50 & 75.47 \\
Zero-shot & InternVL3.5-38B & 42.05 & 30.34 & 22.56 & 53.12 & 41.51 \\
Zero-shot & GPT-5.5 & 37.12 & 10.40 & 26.09 & 81.88 & 81.76 \\
Zero-shot & Gemini-3.1-Pro & 36.67 & 20.96 & 31.91 & 96.25 & 84.28 \\
Zero-shot & Claude Sonnet 5 & 14.13 & 12.35 & 15.20 & 85.00 & 40.25 \\
\midrule
\multicolumn{7}{l}{\emph{Trained}} \\
DeepEyes~\citep{deepeyes} & Qwen2.5-VL-7B & 30.67 & 25.72 & 40.03 & 62.50 & 43.40 \\
GeoChat~\citep{geochat} & Vicuna-7B & 25.87 & 18.23 & \overall 46.88 & 27.50 & 22.01 \\
GeoLLaVA-8K~\citep{geollava8k} & -- & 0.00 & 0.00 & 0.00 & 30.00 & 22.64 \\
GeoEyes~\citep{geoeyes} & Qwen2.5-VL-7B & 28.22 & 22.44 & 25.36 & 57.50 & 40.88 \\
\midrule
\multicolumn{7}{l}{\emph{Training-free}} \\
HiDe~\citep{hide} & Qwen2.5-VL-7B & 40.95 & 20.20 & \second{40.84} & \second{85.62} & 41.51 \\
ZoomEye~\citep{zoomeye} & Qwen2.5-VL-3B & 22.92 & 17.49 & \best{44.87} & 23.75 & 32.70 \\
ZoomSearch~\citep{zoomsearch} & Qwen3-VL-8B & 25.68 & 11.42 & 15.39 & 58.13 & 47.80 \\
MAP~\citep{uhrmicro} & InternVL3.5-8B & \overall \best{43.19} & \second{57.81} & 30.86 & 44.38 & 38.36 \\
MAP~\citep{uhrmicro} & Qwen3-VL-8B & 36.87 & 45.09 & 24.21 & 79.38 & \second{67.92} \\
\midrule
\multicolumn{7}{l}{\emph{Ours}} \\
GeoScope & Qwen3-VL-8B & 40.49 & 49.34 & 17.48 & \overall \best{100.00} & \overall \best{100.00} \\
\ $\Delta$ & & \dci{+16.8}{+13.0}{+20.7} & \dci{+36.2}{+32.8}{+39.6} & \dci{-6.0}{-12.1}{+0.2} & \dci{+34.4}{+26.9}{+41.9} & \dci{+28.3}{+21.4}{+35.9} \\
GeoScope & InternVL3.5-8B & \second{42.23} & \overall \best{58.77} & 26.19 & \overall \best{100.00} & \overall \best{100.00} \\
\ $\Delta$ & & \dci{+8.2}{+4.3}{+12.3} & \dci{+37.7}{+34.1}{+41.5} & \dci{-0.1}{-5.8}{+5.6} & \dci{+56.3}{+48.8}{+63.8} & \dci{+62.9}{+55.4}{+70.4} \\
\bottomrule
\end{tabular}}
\par\smallskip
\begin{minipage}{\textwidth}\footnotesize
Global, Regional, Conditional: counting; Direction, Distance: relations between two objects marked in the image. Rows other than
ours are reported by UHR-Micro, including the general-domain high-resolution methods DeepEyes, HiDe
and ZoomEye. $\Delta$: paired difference to our zero-shot re-run with the same backbone, with its 95\%
interval. \rankkey{}; zero-shot rows take part in the shading only, and our re-runs are not ranked. GeoScope uses
$h=5$ with both backbones; with the threshold of InternVL3.5-8B, $h=40$, it scores 44.01, 56.88 and 26.40 on the three
counting tasks (\appref{supp:umicro}).
\end{minipage}
\end{table}

UHR-Micro~\citep{uhrmicro} uses images of about
$4{,}800\times3{,}200$ pixels from DOTA, FAIR1M, SODA-A and xView. A count $\hat c$ of $c$ objects scores
$\max(0, 1-|\hat c-c|/c)$, so a close count earns partial credit. We evaluate the five
task types that GeoScope covers, 1{,}369 test questions (Table~\ref{tab:umicro}). Regional counting
questions give the region either as a pixel box, which becomes the program's region, or in words, as conditional
counting questions do. Direction and distance questions mark the two objects with rectangles of fixed colours. The
executor finds them by colour, as it finds the red circle of XLRS-Bench RC, and computes the relation from their
boxes. We use the configuration of MME-RealWorld counting unchanged and add only the pixel-box region to the program
language; the validation split confirmed the 15\% margin used on the other benchmarks
(\appref{supp:umicro}).

GeoScope improves Qwen3-VL-8B by 16.8 points on global and 36.2 on regional counting, and is above MAP, the
benchmark's own training-free method, on both. Most of the global gain comes from subdivision. Counting each region in
a single unsplit view scores 24.87, close to zero-shot answering, and subdivision adds 15.6 points. On regional
counting the unsplit view of the given box already scores 42.43, and subdivision adds 6.9; both gains are significant
(\appref{supp:umicro}). InternVL3.5-8B counts better zero-shot, and MAP is also reported with it.
With this backbone GeoScope gains 8.2 and 37.7 points, has the highest reported score on regional counting, 58.77 against 57.81 for MAP,
and is one point below MAP on global counting.

Conditional counting does not improve significantly with either backbone. Half of these questions ask for a fine-grained aircraft type such as A220, on which GeoScope with
Qwen3-VL-8B scores 8.3 and zero-shot answering 3.3. In the others the place is described in words, and when the
program grounds the wrong place, the count belongs to another region. Every direction and distance question is
answered correctly, because once the marked boxes are found the relation is computed from them.

\subsection{Ablation study}
\label{sec:exp:ablation}

\begin{table}[!tp]
\centering
\caption{Ablations of GeoScope with Qwen3-VL-8B, one component changed at a time (accuracy, \%).}
\label{tab:ablation}
\footnotesize
\setlength{\tabcolsep}{4pt}
\begin{tabular}{lcccccc}
\toprule
 & \multicolumn{2}{c}{OSR} & \multicolumn{4}{c}{MME-RS count} \\
\cmidrule(lr){2-3}\cmidrule(lr){4-7}
 & dev & test & dev & test & $N\leq5$ & $N=6$--20 \\
\midrule
GeoScope & 56.0 & 44.00 & 43.3 & 36.60 & 37.29 & 35.25 \\
\ centre angle for directions & 46.0 & 42.00 & -- & -- & -- & -- \\
\ no hierarchical grounding & 49.0 & -- & -- & -- & -- & -- \\
\ zeros from downsampled views accepted & -- & -- & 41.2 & 36.70 & 38.14 & 33.33 \\
\ no subdivision & -- & -- & 40.0 & 35.98 & 37.71 & 32.57 \\
\ fixed 4$\times$4 grid instead of subdivision & -- & -- & 33.1 & 33.74 & 29.29 & 46.74 \\
\ no program (whole image) & -- & -- & 37.6 & 32.62 & 30.43 & 39.08 \\
\ \ and no subdivision & -- & -- & -- & 32.72 & 32.57 & 34.48 \\
\ programs by the VLM, text & 49.0 & -- & 37.6 & -- & -- & -- \\
\ programs by the VLM, image & 52.0 & -- & 38.0 & -- & -- & -- \\
\midrule
Backbone zero-shot & 23.0 & 33.50 & 26.5 & 25.18 & 28.43 & 15.33 \\
\bottomrule
\end{tabular}
\par\smallskip
\begin{minipage}{\textwidth}\footnotesize
$N\leq5$, $N=6$--20: test accuracy on the 700 and 261 counting questions with these true numbers of objects
(\appref{supp:results:ablation}).
\end{minipage}
\end{table}

A program fixes where to look, grounds objects, splits views and computes geometry, and the benchmarks reward these
steps differently. To separate the window from what follows it, we asked the backbone the original question on the
window each test program chose (\appref{supp:results:window}). On XLRS-Bench RC it answers as well as the count does,
so there the gain comes from the window. The circled regions are small, and a crop shows one at a far higher
resolution than the whole image does, the kind of view in which the horizon lies further out
(Fig.~\ref{fig:horizon}c). Such a crop is counted in a single view. On OSR the
computed direction is significantly more accurate, by 6.8 points, than the backbone's answer on the union of the two
referents. Grounding a place before the objects inside it adds 7 development points, and the direction-relation
matrix 10 development and 2.0 test points over the centre angle (Table~\ref{tab:ablation}). On MME-RealWorld counting
the count beats the direct answer on the window by 12.5 points, a significant difference, and the direct answer
restricted to the numeric options by 3.2 points, which is not significant. Most of the gain there comes
from the window and from answering with a number. Subdivision supplies most of the gain where scenes are dense and a
count is scored by its error, as on UHR-Micro global counting (Section~\ref{sec:exp:umicro}).

Without a program, the whole image is counted with the same procedure, and test accuracy on counting falls from 36.60
to 32.62. GeoScope's gain over zero-shot answering is 13.5 points on the 859 questions whose program restricts the
region, and slightly negative on the 103 about the whole image, whose answers are small.

Both effects that Proposition~\ref{prop:split} predicts, the recovered undercount beyond the horizon and the spurious
counts below it, appear on MME-RealWorld in the last two columns of Table~\ref{tab:ablation}. When the whole image is
counted, as in the rows without a program, removing subdivision costs a significant 4.6 points on questions with 6 to 20 objects and gains 2.1 on those
with at most five. Accepting zeros from downsampled views has a similar effect. Within the regions that programs name,
which already divide the scene, the loss on 6 to 20 objects shrinks to 2.7 points, which is not significant. Most
counting questions ask for at most five objects, so the two effects nearly cancel on the whole task. With
Qwen3-VL-4B, which undercounts more strongly, removing subdivision costs a significant 2.2 points overall. Since the
answer to a dense question is often the largest number offered, these differences also reward erring upwards;
UHR-Micro, which scores a count by its relative error, tests the proposition without this bias.

A fixed 4$\times$4 grid in place of count-triggered subdivision costs about as many VLM calls and lowers test accuracy
by 2.9 points, not significantly. On questions about at most five objects its median count is 2.7 times the truth,
while that of subdivision is exact, so it picks the largest number offered about twice as often. It therefore gains
significantly on the questions whose answer is the largest number offered and loses on the others. On XLRS-Bench
overall counting it needs three times as many calls and answers 10.4 points fewer test questions correctly, again
not significantly (\appref{supp:results:ablation}). Programs written by the VLM, with or without the image, lose four to seven
development points (\appref{supp:method:synth}).

\subsection{Computational cost}
\label{sec:exp:cost}

\begin{figure}[!tp]
\centering
\includegraphics[width=\linewidth]{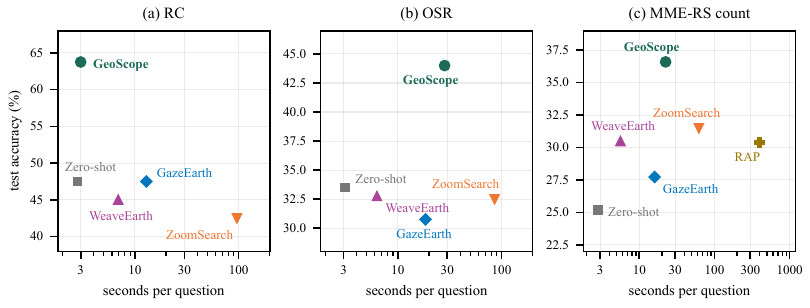}
\caption{Test accuracy against mean time per question on one NVIDIA GeForce RTX 3090, for every method run on the
whole test split of a task; GeoScope and zero-shot answering are timed on development questions.}
\label{fig:cost}
\end{figure}

GeoScope and zero-shot answering are timed on development questions. GeoScope starts from
scratch on each question, and its times include program synthesis. The reproduced methods are timed on
the test questions of each task (Fig.~\ref{fig:cost}), and RAP with LLaVA-OV-7B on the fixed subset
(\appref{supp:cost}). ZoomSearch and RAP swap their two models to fit in 24\,GiB, and their
times include the swap. GeoScope's cost depends on how many VLM calls its program makes. An RC question needs about two
VLM calls on small crops and costs about as much as zero-shot answering. OSR and counting questions need five and
seven calls at the median, with a long tail for dense regions. Synthesis takes one to four seconds per question, and
executing a program less than ten milliseconds; almost all the rest goes to VLM calls. On the same GPU, ZoomSearch is
2.8 to 32 times slower than GeoScope and RAP 18 to 37 times. WeaveEarth and GazeEarth are faster on OSR and counting.

\begin{figure}[!tp]
\centering
\includegraphics[width=0.63\linewidth]{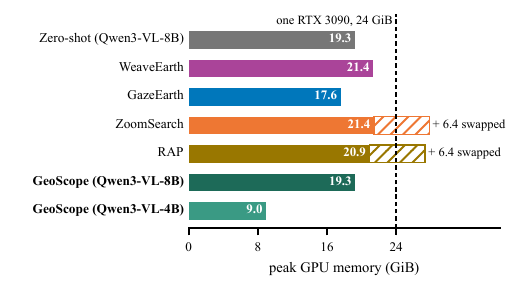}
\caption{Peak GPU memory on one RTX 3090, the largest over RC, OSR and MME-RS counting. GeoScope's bars show program
execution; its synthesiser runs before it, at 14.5\,GiB. Hatched: the retriever that ZoomSearch and RAP keep in host
memory.}
\label{fig:memory}
\end{figure}

Peak GPU memory differs little between methods, because the weights of an 8B backbone take about 16\,GiB of it
(Fig.~\ref{fig:memory}). GeoScope peaks at 19.3\,GiB while it executes programs, about as much as zero-shot answering,
and its synthesiser runs as a separate stage at 14.5\,GiB. With Qwen3-VL-4B, which matches the 8B system on counting
(Section~\ref{sec:exp:main}), program execution needs 9.0\,GiB, and the synthesiser stage sets the peak. ZoomSearch and RAP fit on the card only by keeping one of their two
models in host memory while the other runs; together with the retriever's 6.4\,GiB they would exceed 24\,GiB.

\subsection{Qualitative comparison}
\label{sec:exp:examples}

\begin{figure}[!tp]
\centering
\includegraphics[width=\linewidth]{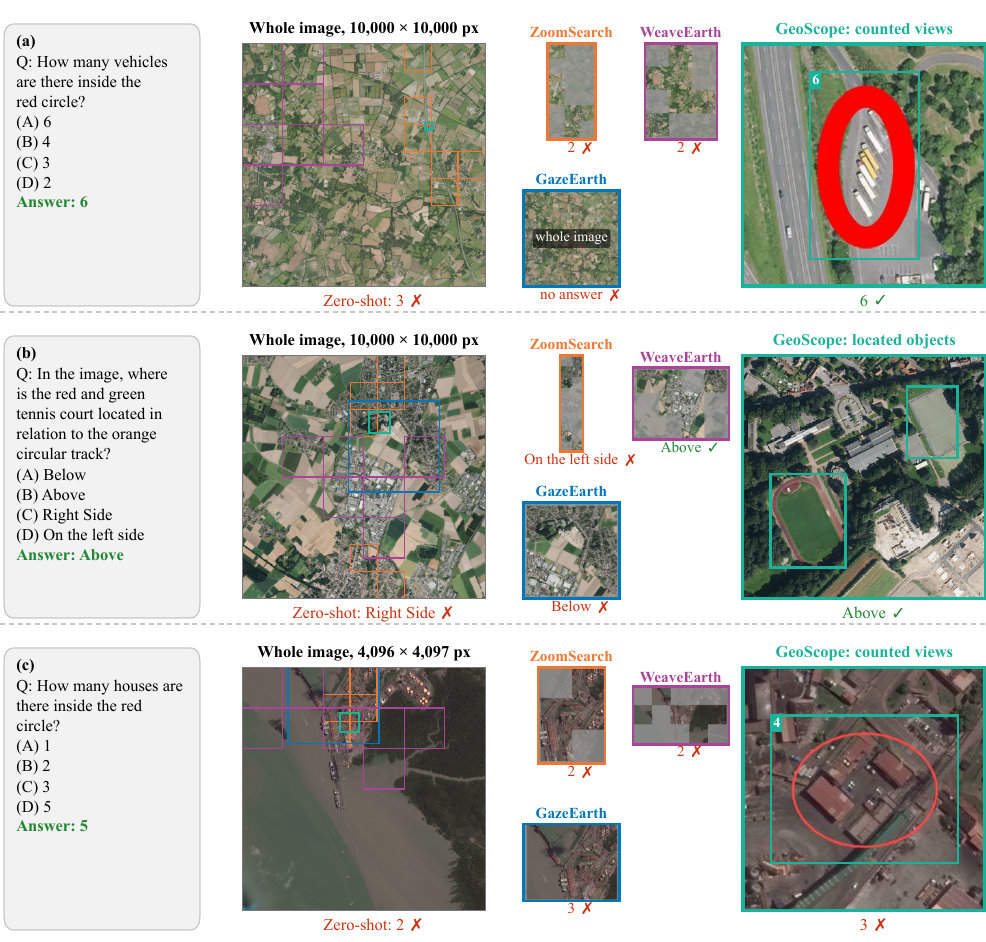}
\caption{Three XLRS-Bench test questions (ZoomSearch with LLaVA-OV-7B, the others with Qwen3-VL-8B). Second column: the regions
each method kept; third: each search method's final view, discarded regions in grey; right: GeoScope's counted views
or located objects, places and anchors dashed. WeaveEarth keeps cells of a 6$\times$6 grid. Green: correct.}
\label{fig:exxlrs}
\end{figure}

\begin{figure}[!tp]
\centering
\includegraphics[width=\linewidth]{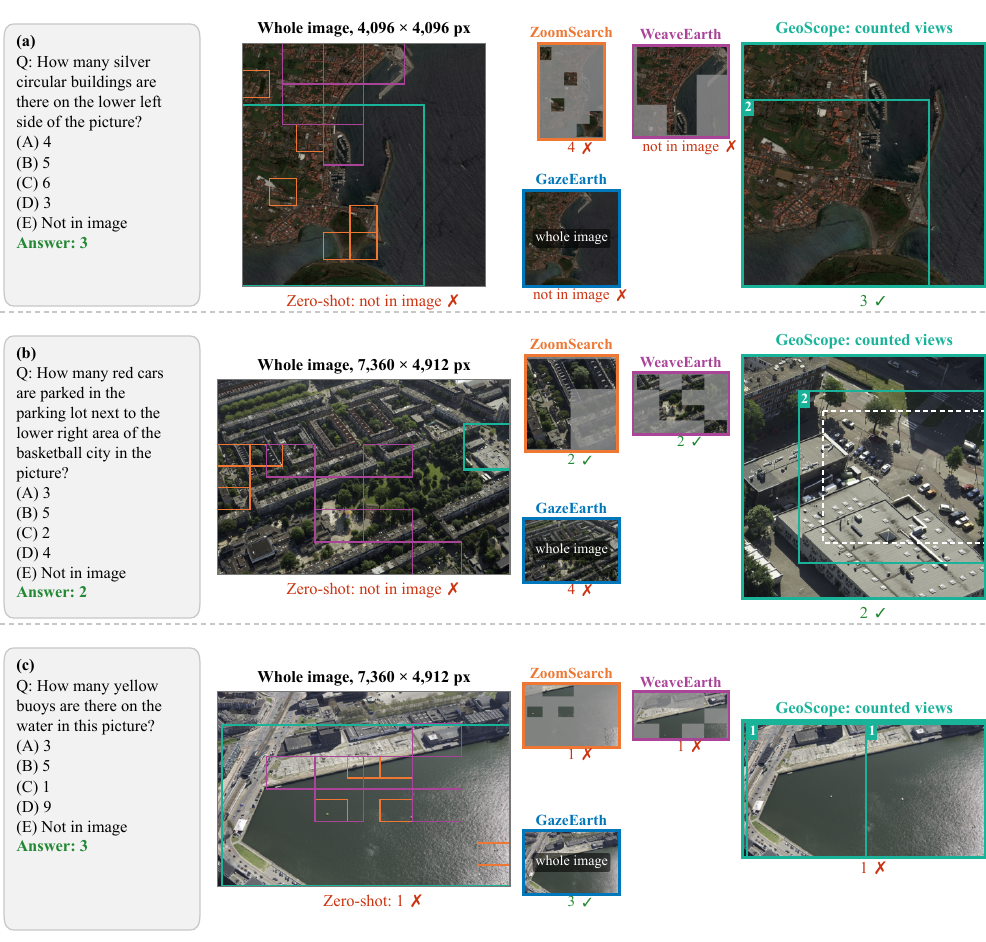}
\caption{Three MME-RealWorld counting test questions, shown as in Fig.~\ref{fig:exxlrs}.}
\label{fig:exmme}
\end{figure}

\begin{figure}[!tp]
\centering
\includegraphics[width=\linewidth]{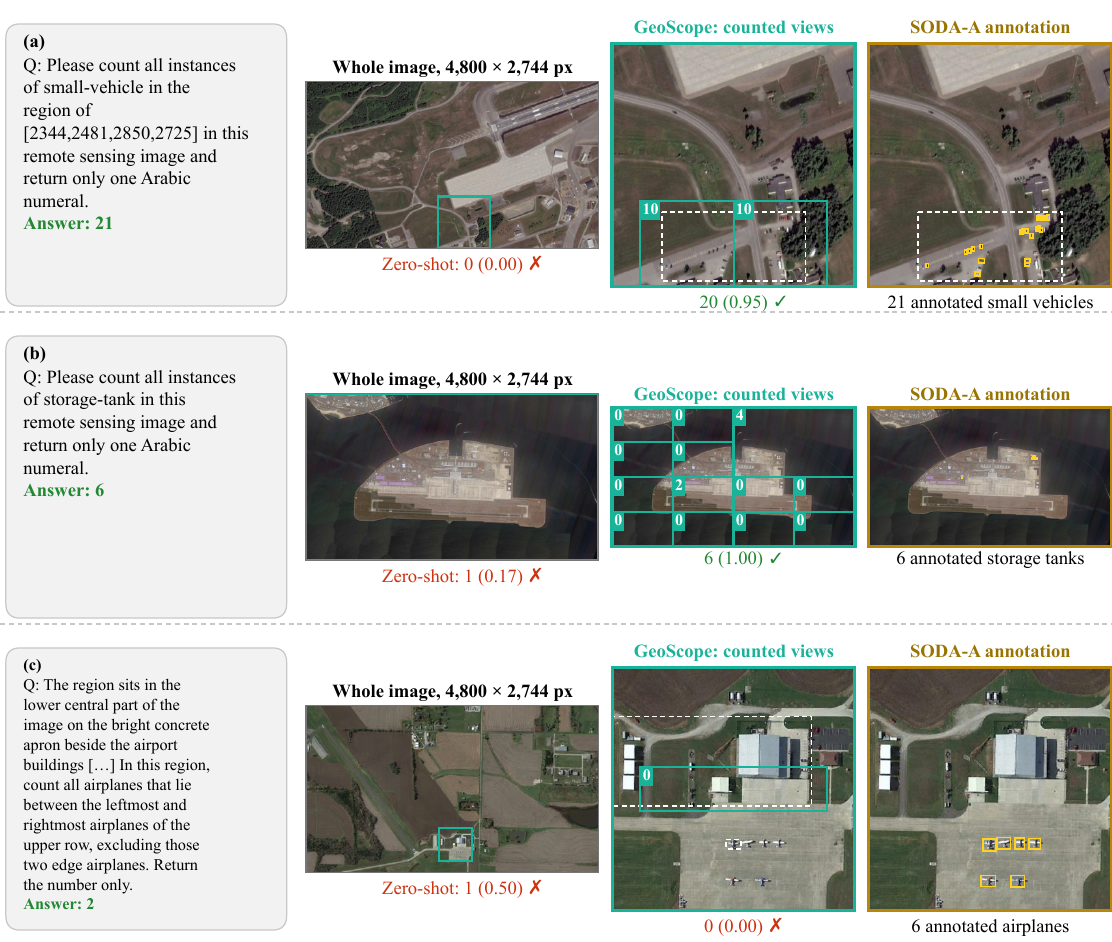}
\caption{Three UHR-Micro test questions (Qwen3-VL-8B), shown as in Fig.~\ref{fig:exxlrs}, with official scores in
brackets. Right: SODA-A boxes of the asked class, the question's region dashed. [\ldots] marks an omitted part of a
long question.}
\label{fig:exumicro}
\end{figure}

Figs.~\ref{fig:exxlrs}, \ref{fig:exmme} and~\ref{fig:exumicro} show test questions with the regions each method
inspected, from the runs in Tables~\ref{tab:main} and~\ref{tab:umicro}. The examples are drawn at random: on each
benchmark, two questions that GeoScope answers correctly and zero-shot answering does not, and one that it answers wrongly.
On UHR-Micro a score of at least 0.8 counts as correct, and the wrong example scores at most 0.3. The search methods pick regions by
the VLM's confidence, and in these examples their regions often miss the small one the question names. GeoScope's views follow from the
program instead: the crop of the red circle, the two referents, or the cells of a split region. Counts are matched
to the options as in Section~\ref{sec:method:projection}, so in Fig.~\ref{fig:exmme}(a) a count of 2 selects the
option 3. One correct answer,
in Fig.~\ref{fig:exmme}(b), is right for the wrong reason. GeoScope finds no parking lot next to the basketball court,
drops the anchor, and counts the red cars of a parking lot at the other end of the image. Among the wrong answers, in
Fig.~\ref{fig:exxlrs}(c) the VLM counts four of the five houses and option matching picks 3. In
Fig.~\ref{fig:exmme}(c) the water is counted in two halves at reduced resolution. Each half reports one buoy, a count
within the horizon, so neither is split; the sum of 2 falls one short of 3, and option matching picks 1. In Fig.~\ref{fig:exumicro}(c) the program places the
region beside the hangars, above the parked airplanes it should count.

\subsection{Failure analysis}
\label{sec:exp:failures}
XLRS-Bench overall counting asks how many ships or cars a whole scene holds, typically several dozen. GeoScope lifts
zero-shot answering from 10.42 to 35.42, but like every method we ran it does not beat always picking the
second-largest of the four numbers, which scores 37.50. Adjacent options lie about 17\% apart, so a correct choice
needs a count within about 8\%, and GeoScope's counts there are unbiased but scatter by 18\% around the truth.
Subdivision removes the undercount of Section~\ref{sec:horizon}, but not the spurious counts and unresolved objects of
many small leaves.

Across tasks, the main remaining error is grounding the wrong object. On OSR, 86\% of the
wrong answers point more than 45 degrees away from the correct option, which correctly grounded referents would not
produce, and counting errors are as often too high as too low (\appref{supp:failures}). In a manual check of 150 programs, those that simplify their question are answered correctly about as often as the
rest, and hand-written programs change accuracy by at most two of thirty questions per task
(\appref{supp:method:correctness}). Perception limits accuracy more than program synthesis does.

\section{Discussion}
\label{sec:discussion}

\subsection{Option bias in counting benchmarks}
\label{sec:discussion:subdivision}
Of the 981 MME-RealWorld counting test questions, 700 concern at most five objects, within the horizon, and only 81
more than ten. Counting dense regions well therefore moves the overall score little, and a procedure that adds
spurious counts to sparse regions, as a fixed grid does (Sections~\ref{sec:horizon:theory}
and~\ref{sec:exp:ablation}), can lose more than it gains.

The reproduced methods also choose the option stating that the image does not show the number
(Section~\ref{sec:exp:baselines}), for 9 to 18\% of the test questions (Table~\ref{tab:mmeoptions}; other
models in \appref{supp:results:mmeoptions}). The restricted score of Qwen3-VL-8B, 35.64 on all questions, is above what WeaveEarth and GazeEarth report with this model, and close to the trained RS-OPSD
(Table~\ref{tab:published}). The same restriction brings LLaVA-OV-7B within two points of what ZoomSearch reports for
itself and for RAP with this model, while ZoomSearch's table lists the model itself at 27.57. On our test split, every reproduced method run on the whole split falls
significantly below the restricted answers of its own backbone. ZoomSearch, which runs on LLaVA-OV-7B, is also
significantly below those of Qwen3-VL-8B, by 4.0 points.
GeoScope is level with them with three
backbones and above them with Qwen3-VL-4B. Counting gains on this benchmark therefore depend strongly on how often
the backbone picks the option stating that the image does not show the number. Comparing with the backbone
restricted to the numeric options controls for this.

Where the correct number sits among the options also matters. It is the largest number offered for 52\% of the test
questions about 6 to 10 objects and 41\% of those about 11 to 20, but never for one or two objects. A method that
undercounts dense scenes can rarely answer these questions, and one that overcounts gains on them whether or not its
count is close. The 15\% margin around a counted window does not make GeoScope overcount. In windows with the margin its counts are too high for 274 test questions and too low for 226, about as
in whole-image windows without it, 120 and 91. Without the margin, development accuracy falls from 43.3 to 37.6; as
many counts are too high as before, and more are too low (\appref{supp:results:budget}). On UHR-Micro,
whose score penalises overcounts, the margin raised the development score from 45.34 to 48.30 (\appref{supp:umicro}).

XLRS-Bench overall counting has a different problem. Its scenes hold dozens of objects, but its options lie closer together
than any method we ran can count (Section~\ref{sec:exp:failures}). To tell counting procedures apart, a benchmark would need questions spread over the
whole range of counts, and options spaced in proportion to the count and placed at random around it. It would also
need to drop any option that is almost never correct.

\subsection{Scope of the program language}
\label{sec:discussion:scope}
For the position and colour questions of
MME-RealWorld we built programs that locate the object and read its position from geometry or name its colour on a
close-up. They were to replace the backbone only if they beat it by at least five points on the development split,
a rule set before they were run. They reached 74.9 against 73.7 for the backbone on position, not a significant
difference, and 60.6 against 65.7 on colour, so these questions go to the backbone. The program language has types for sets of boxes, image
regions, counts and directions, but none for land cover, time or paths. Land use, anomaly and environmental reasoning,
route planning and change questions therefore lie outside it. Even within its scope, 43 of the 245 development
counting questions use constructions the grammar lacks, such as ordinals, \emph{between} or the state of an object.
They are answered by simplified programs, with 51.2\% accuracy against 41.6\% for the rest
(\appref{supp:method:correctness}). Questions about change or about measurement in physical units would need types for time and for units.

\subsection{Ground sampling distance and other imagery}
\label{sec:discussion:imagery}
Whether a VLM can count a view depends on how many pixels each object covers as shown, and also on the extent of the
view (Section~\ref{sec:horizon:measure}). The pixel size of an object follows from the ground sampling distance (GSD)
and the scale at which the view is rendered. A car 4.5\,m long spans 15 pixels at a GSD of
0.3\,m. Rendered whole at 2.5\,M pixels, a scene of 10{,}000 pixels on a side is reduced 6.3-fold, and the car spans
about two pixels. A 256-pixel cell of the same scene, enlarged to 768 pixels, shows it at about 45. Larger apparent objects account
for part of the later onset of the horizon on FAIR1M, DIOR and VisDrone (Fig.~\ref{fig:imagery}); on VisDrone, object
size alone does not explain it (\appref{supp:horizon:visdrone}). Measuring the horizon takes a few hundred crops with annotated counts, binned by
the number of objects; we used 373 to 480 per dataset. In our experiments GeoScope uses $h=5$ for every backbone and
dataset. A threshold for a given model and imagery could instead be estimated once in
this way, on the imagery it is used for. For InternVL3.5-8B, the threshold of 40 measured on DOTA raises UHR-Micro
global counting by 1.8 points and lowers regional counting by 1.9, because its single views of UHR-Micro already fall
short below 40 objects (\appref{supp:umicro}). GeoScope's programs, prompts and executor do not depend on the imagery.

\subsection{Generality beyond remote sensing}
\label{sec:discussion:beyond}
Every model we tested falls short beyond some number of objects, with an onset that depends on the model and, for
Qwen3-VL-8B, on the imagery (Section~\ref{sec:horizon:measure}), and recent probing studies locate the limit inside the
VLM~\citep{numerosityvlm,che2026convstack}. Our measurements, however, are confined to aerial and drone images
of a few object classes, and Proposition~\ref{prop:split} is tested only there. Whether the same response
curve and the same predictable gain from splitting hold for crowds, cells or everyday scenes, and whether the horizon
follows the architecture or the distribution of numbers in the training data, remain open. GeoScope's
counting procedure takes the horizon as a single threshold, so it can be applied in such a study by setting $h$ from a
measurement on the new imagery.

\section{Conclusion}
\label{sec:conclusion}

Frozen VLMs fall increasingly short when an aerial view holds more than a few objects of a class, about five on DOTA
for most of the models we tested. Where this horizon lies depends on the view, the objects and the imagery. A simple model of the answers predicts what a split recovers and what it adds, and motivates
splitting only views counted beyond the horizon. In GeoScope, a deterministic executor runs typed programs, chooses the
views and asks the VLM only to ground and count in them. It improves Qwen3-VL-8B on XLRS-Bench
regional counting, spatial relations and MME-RealWorld counting more than any reproduced training-free method improves
its own backbone, and it has the highest reported accuracy, trained or not, on XLRS-Bench regional counting and on the
whole benchmark by question. On UHR-Micro, which scores counts by their error, subdivision alone adds 15.6 points to
global counting, while on regional counting most of the gain comes from the window that the program fixes. On
MME-RealWorld counting, an option that is almost never correct distorts comparisons. Barred from it, the backbone
alone significantly beats every reproduced method run on all test questions, and GeoScope is level with it or above
it.

GeoScope leaves to its backbone the position and colour questions without a pixel box, and questions on land use and
reasoning. Whole-scene counts with closely spaced options need a more precise counting primitive than a single VLM view
(Section~\ref{sec:exp:failures}). A wrongly grounded place gives a count
for the wrong region, which limits GeoScope on the conditional counting of UHR-Micro (Section~\ref{sec:exp:umicro}).

More broadly, measuring where a frozen model can be trusted and computing everything else explicitly from what it
grounds improves answers without task-specific training.

\section*{CRediT authorship contribution statement}
\textbf{Guohui Deng:} Writing -- original draft, Writing -- review \& editing, Methodology, Software, Visualization,
Supervision, Funding acquisition. \textbf{Yuhang Jiang:} Writing -- original draft, Writing -- review \& editing,
Conceptualization, Methodology, Software, Investigation, Formal analysis, Data curation, Validation, Visualization.
\textbf{Chengjun Wang:} Writing -- review \& editing. \textbf{Zhongwen Hu:} Writing -- review \& editing, Supervision,
Funding acquisition. \textbf{Zhaocong Wu:} Writing -- review \& editing. \textbf{Linsheng Huang:} Writing -- review \&
editing. \textbf{Jinling Zhao:} Writing -- review \& editing, Supervision, Funding acquisition. \textbf{Chao Ruan:}
Writing -- review \& editing, Funding acquisition. \textbf{Miaozhong Xu:}
Writing -- review \& editing. \textbf{Chi Chen:} Writing -- review \& editing.

\section*{Declaration of competing interest}
The authors declare that they have no known competing financial interests or personal relationships that could have
appeared to influence the work reported in this paper.

\section*{Declaration of generative AI and AI-assisted technologies in the manuscript preparation process}
During the preparation of this work the authors used Claude (Anthropic) and ChatGPT (OpenAI) in order to assist with the
experiment code and to improve the readability and language of the manuscript. After using these tools, the authors
reviewed and edited the content as needed and take full responsibility for the content of the published article.

\section*{Acknowledgements}
This research was supported in part by the National Natural Science Foundation of China (Grant Nos.~42601629 and
42471351), the Natural Science Foundation of Anhui Province (Grant Nos.~2508085QD124 and 2508085MD080), the Scientific
Research Project of the Department of Education of Anhui Province (Natural Science) (Grant No.~2025AHGXZK40258), and
the Open Research Fund of the National Engineering Research Center for Agro-Ecological Big Data Analysis and
Application, Anhui University (Grant No.~AE202501).

\section*{Data availability}
XLRS-Bench, MME-RealWorld, UHR-Micro, SODA-A, DOTA, FAIR1M, DIOR and VisDrone are publicly available. The code will be made publicly available
upon acceptance.

\appendix

\section{Proof of Proposition 1}
\label{supp:theory}
The proof uses Eq.~\eqref{eq:response} through one assumption only. The expected answer to a view depends only on the
number of objects of the queried class in it. A part holding $n_k$ objects is therefore expected to be answered with
$c(n_k)$, whatever its size and position. This holds only approximately, since a smaller part shows its objects
larger (Section~\ref{sec:horizon:measure}). Eq.~\eqref{eq:response} describes views within the range of sizes it
was fitted on, with the effect of resolution absorbed into $\gamma$, and the prediction falls short where the parts
reveal objects too small to see in the parent, as above 160 objects (Section~\ref{sec:horizon:theory}).

\begin{proof}[Proof of Proposition~\ref{prop:split}]
Let $p_k=n_k/n$ be the share of the objects in part $k$, so that $0\leq p_k\leq 1$ and $\sum_k p_k=1$. Let
$J=\{k: n_k>0\}$ be the set of non-empty parts and $L=K-m$ its size. Since $n\geq 1$, $L\geq 1$.

\emph{The sum.} By linearity of expectation, the expected sum of the parts' answers is $\sum_{k=1}^{K}c(n_k)$. A part
in $J$ contributes $a\,n_k^{\gamma}=a\,n^{\gamma}p_k^{\gamma}=c(n)\,p_k^{\gamma}$, and each of the $m$ empty parts
contributes $c(0)=s$. Hence
\begin{equation*}
\sum_{k=1}^{K}c(n_k)\;=\;c(n)\sum_{k\in J}p_k^{\gamma}\;+\;m\,s .
\end{equation*}
An empty part has $p_k^{\gamma}=0$ because $\gamma>0$, so the sum over $J$ equals $A=\sum_{k=1}^{K}p_k^{\gamma}$, and
the display is Eq.~\eqref{eq:split}.

\emph{The lower bound.} For $0<p\leq 1$ we have $p^{\gamma}=p\cdot p^{\gamma-1}\geq p$, because $p\leq 1$ and
$\gamma-1\leq 0$ give $p^{\gamma-1}\geq 1$. Summing over $J$ yields $A\geq\sum_{k\in J}p_k=1$. If $\gamma<1$, the
inequality $p^{\gamma}\geq p$ is strict for every $p<1$, so $A=1$ requires $p_k=1$ for each $k\in J$. As the shares
sum to one, $J$ then has a single member, and all objects lie in one part. If $\gamma=1$, every split gives $A=1$.

\emph{The upper bound.} The function $x\mapsto x^{\gamma}$ is concave on $[0,\infty)$. Jensen's inequality over the
$L$ non-empty shares gives
\begin{equation*}
\frac{1}{L}\sum_{k\in J}p_k^{\gamma}\;\leq\;\Big(\frac{1}{L}\sum_{k\in J}p_k\Big)^{\gamma}\;=\;L^{-\gamma},
\end{equation*}
hence $A\leq L^{1-\gamma}\leq K^{1-\gamma}$, where the last step uses $L\leq K$ and $1-\gamma\geq 0$. If $\gamma<1$,
the function is strictly concave, so the first inequality is an equality only if every non-empty share equals $1/L$,
and the second only if $L=K$. The upper bound is then reached only when no part is empty and every part holds $n/K$
objects. If $\gamma=1$, both bounds equal one.
\end{proof}

The lower bound is the subadditivity of $x\mapsto x^{\gamma}$ invoked in the main text. The recursion splits a window
into $K=4$ parts, or into $K=2$ when the window is elongated (Algorithm~\ref{alg:horizon}). With the exponent of
Qwen3-VL-8B, 0.76, one split can therefore raise a compressed answer by a factor of at most $4^{0.24}\approx1.39$, or
$2^{0.24}\approx1.18$.
The bound requires an even spread of the objects and no empty part. The medians predicted for this model,
which use the true counts of the parts, lie between 1.16 and 1.33 on its two sets of crops. \appref{supp:model} lists
them, tests the model split by split and shows where its assumption fails, in particular for objects too small to be seen in a
large view.

\bibliographystyle{elsarticle-harv}
\bibliography{refs}

\clearpage
\makeatletter
\setcounter{section}{0}\setcounter{subsection}{0}\setcounter{table}{0}\setcounter{figure}{0}\setcounter{equation}{0}
\counterwithout{equation}{section}
\gdef\thesection{S\arabic{section}}
\gdef\thetable{S\arabic{table}}
\gdef\thefigure{S\arabic{figure}}
\gdef\theequation{S\arabic{equation}}
\renewcommand{\appref}[1]{\def\@tempa{#1}\def\@tempb{supp:theory}\ifx\@tempa\@tempb\ref{#1}\else Section~\ref{#1}\fi}
\makeatother
\begin{center}
{\Large\bfseries Supplementary Material}
\end{center}
\vspace{4pt}
\noindent This part holds the details that the article refers to as Supplementary Sections S1 to S8. Section, table,
figure and equation numbers without an S refer to the article.
\addtocontents{toc}{\protect\setcounter{tocdepth}{2}}
{\small\tableofcontents}
\clearpage

\section{Validation and limits of the response model}
\label{supp:model}
\paragraph{Split-level test}
We replay the recursion of the horizon measurement from the recorded answers. Every view it looked at, root crop or
sub-window, is scored against the annotated objects whose centres it contains.
The replay covers 479 of the 480 crops of the first set of DOTA crops and 458 of the second
(Section~\ref{sec:horizon:measure}; \appref{supp:horizon:crops}). Nineteen crops of the second set
coincide with crops of the first and are left out, as are four crops whose windows are not available, one of the first
set and three of the second. For the other models the replay covers 458 to 479 crops, the rest having an unanswered
or unrecorded view in their tree. The replayed total differs from the original count in 10 of the 479 crops of the first
set, which raises the recursive error at three to five objects from $+0.08$ (Table~\ref{tab:supp:models}) to $+0.19$
(Table~\ref{tab:supp:decomp}) and changes no other bin by more than 0.02.
At each split we compare the ratio of the parts' summed answers to the parent's answer with $A$ computed from the true
counts of the parts. The exponent is fitted on the unsplit crops of the same model, by least squares of $\log c$ on
$\log N$ over the crops with more than five objects and a non-zero answer.

Table~\ref{tab:supp:split} lists the medians per bin of the parent's true count. For Qwen3-VL-8B the observed and
predicted medians differ by at most 0.08 in the five bins from 6 to 160 objects on the first set.
Eq.~\eqref{eq:response} and Proposition~\ref{prop:split} were formulated after the replay of the first set, so the
second set tests them with every parameter taken from the first.
On the second set, with the exponent fixed at 0.76 from the first, they differ by at most 0.10 in four of these bins
and by 0.17 at 41 to 80 objects.
The second set shares 63 of its 130 images with the first. Restricted to its 249 replayed crops on the 67 images that
the first set does not use, the fitted exponent is again 0.76. Observed and predicted medians then differ by at most
0.06 from 6 to 40 objects, but by 0.27 at 41 to 80 and 0.14 at 81 to 160 objects, where the root crops of this
subset are answered with 94\% and 85\% of their objects at the median.
For the seven other models the ratio of observed to predicted amplification lies between 0.8 and 1.2 in 35 of their 42
bins above five objects, and between 0.72 and 1.32 in all of them.
Parents with at most five objects are split only when they are overcounted. For them the median ratio of the parts'
sum to the parent's answer is 1.12 on the first set and 1.00 on the second, whereas an exact count would need 0.30 on both.
Splitting a view with few objects thus keeps or adds to an overcount.

\begin{table}[!tp]
\centering
\caption{Amplification of one split for parents with more than five objects, per bin of the parent's true count.}
\label{tab:supp:split}
\footnotesize
\setlength{\tabcolsep}{3pt}
\resizebox{\textwidth}{!}{%
\begin{tabular}{lcccccccc}
\toprule
Model & $\gamma$ & Splits & 6--10 & 11--20 & 21--40 & 41--80 & 81--160 & $>$160 \\
\midrule
Qwen3-VL-8B, first set & 0.76 & 897 & 1.10 / 1.18 & 1.22 / 1.23 & 1.34 / 1.27 & 1.22 / 1.29 & 1.27 / 1.32 & 2.11 / 1.33 \\
Qwen3-VL-8B, second set & 0.76 & 996 & 1.10 / 1.16 & 1.20 / 1.18 & 1.35 / 1.25 & 1.11 / 1.28 & 1.23 / 1.31 & 2.14 / 1.32 \\
\ new images & 0.76 & 609 & 1.10 / 1.16 & 1.20 / 1.18 & 1.29 / 1.26 & 1.02 / 1.29 & 1.17 / 1.31 & 2.14 / 1.28 \\
Qwen3-VL-4B & 0.64 & 1{,}111 & 1.20 / 1.27 & 1.30 / 1.35 & 1.60 / 1.44 & 1.67 / 1.47 & 1.48 / 1.51 & 1.74 / 1.54 \\
Qwen3-VL-2B & 0.48 & 810 & 1.10 / 1.43 & 1.20 / 1.62 & 1.83 / 1.75 & 2.17 / 1.77 & 2.50 / 1.89 & 1.83 / 1.91 \\
Qwen2.5-VL-7B & 0.70 & 871 & 1.30 / 1.23 & 1.40 / 1.32 & 1.55 / 1.37 & 1.63 / 1.39 & 1.67 / 1.43 & 1.71 / 1.45 \\
InternVL3.5-4B & 0.67 & 1{,}124 & 1.14 / 1.26 & 1.15 / 1.34 & 1.17 / 1.41 & 1.16 / 1.43 & 1.07 / 1.49 & 1.30 / 1.49 \\
InternVL3.5-8B & 0.75 & 1{,}118 & 1.15 / 1.18 & 1.11 / 1.20 & 1.14 / 1.28 & 1.20 / 1.32 & 1.35 / 1.35 & 1.20 / 1.36 \\
LLaVA-OV-7B & 0.43 & 1{,}050 & 1.14 / 1.49 & 1.36 / 1.75 & 1.79 / 1.86 & 2.07 / 1.95 & 2.30 / 2.03 & 2.35 / 2.07 \\
Qwen3-VL-32B & 0.80 & 897 & 1.08 / 1.15 & 1.21 / 1.19 & 1.33 / 1.23 & 1.34 / 1.24 & 1.30 / 1.26 & 1.28 / 1.27 \\
\bottomrule
\end{tabular}}
\par\smallskip
\begin{minipage}{\textwidth}\footnotesize
Each cell gives the median observed ratio of the parts' summed answers to the parent's answer and, after the slash, the median predicted amplification $A$ from the true counts of the parts. $\gamma$: the exponent, fitted on each model's own unsplit crops, except for the second set of Qwen3-VL-8B, which uses the exponent of the first. New images: the crops of the second set on images that the first set does not use. Splits: number of split parents with more than five objects.
\end{minipage}
\end{table}

\paragraph{Error decomposition of the recursive count}
Let the recursion on a crop with $N$ objects end in leaves $\ell$ that hold $n_\ell$ objects and are answered with
$C_\ell$. Every object belongs to the leaf that contains its centre, so $\sum_\ell n_\ell=N$, and the count is
$T=\sum_\ell C_\ell$. Grouping the leaves by their true count gives the identity
\begin{equation}
T-N\;=\;\sum_{\ell:\,n_\ell>h}\left(C_\ell-n_\ell\right)\;+\sum_{\ell:\,1\leq n_\ell\leq h}\left(C_\ell-n_\ell\right)\;
+\sum_{\ell:\,n_\ell=0}C_\ell .
\label{eq:supp:decomp}
\end{equation}
If the leaves were fixed in advance, Eq.~\eqref{eq:response} would give the three sums the expectations
$\sum_{n_\ell>h}(a\,n_\ell^{\gamma}-n_\ell)$, $\sum_{1\leq n_\ell\leq h}(a\,n_\ell^{\gamma}-n_\ell)$ and $m_0\,s$,
where $m_0$ is the number of empty leaves. The first is the undercount left in leaves that still hold more than $h$
objects. The second is small, since $c(n)\approx n$ near the horizon, and the third is the spurious count. In the
recursion the leaves depend on the answers. A leaf with more than $h$ objects that stopped because its answer fell
within the horizon was selected for a low answer, so it falls short by more than the undercount alone.

Table~\ref{tab:supp:decomp} gives the three sums, divided by $N$ and averaged per bin, for Qwen3-VL-8B with $h=5$.
From 11 objects on, the shortfall that remains after subdivision comes from the leaves beyond the horizon, which
contribute $-0.16$ to $-0.24$ per bin on the first set.
Leaves with at most five objects hold 9\% of all objects on the first set and
8\% on the second, so about nine in ten objects end in leaves that still hold more than five.
Of the 1{,}134 leaves beyond the horizon on the first set, 885 could not be split further, at the depth limit or
below twice the minimum cell, and 249 had been answered with at most five objects.
The empty leaves add a surplus that depends on the scenes, $+0.47$ for crops with 6 to 10 objects on the first set
but $+0.05$ on the second.
Leaves within the horizon contribute between $-0.07$ and $+0.15$ per bin on the first set. On the second set, crops
with three to five objects are seldom split, and their single views already fall short by 0.27.
Subdivision removes most of the bias, while the spread of the error changes little. Above 20 objects the standard
deviation of the recursive error stays close to that of a single view, 14.5 against 16.6 objects at 21 to 40 and 316.0 against 324.0
above 160.

\begin{table}[!tp]
\centering
\caption{The error of the recursive count of Qwen3-VL-8B with $h=5$, decomposed by Eq.~\eqref{eq:supp:decomp}.}
\label{tab:supp:decomp}
\footnotesize
\setlength{\tabcolsep}{3pt}
\resizebox{\textwidth}{!}{%
\begin{tabular}{lcccccccccccc}
\toprule
 & \multicolumn{6}{c}{First set} & \multicolumn{6}{c}{Second set} \\
\cmidrule(lr){2-7}\cmidrule(lr){8-13}
$N$ & Crops & Single & Recursive & $n_\ell>5$ & $1\leq n_\ell\leq5$ & $n_\ell=0$ & Crops & Single & Recursive & $n_\ell>5$ & $1\leq n_\ell\leq5$ & $n_\ell=0$ \\
\midrule
1--2 & 60 & +0.07 & +0.17 & 0.00 & +0.07 & +0.10 & 55 & $-$0.02 & $-$0.02 & 0.00 & $-$0.02 & 0.00 \\
3--5 & 60 & 0.00 & +0.19 & 0.00 & $-$0.07 & +0.26 & 59 & $-$0.27 & $-$0.28 & 0.00 & $-$0.38 & +0.10 \\
6--10 & 60 & $-$0.25 & +0.30 & $-$0.33 & +0.15 & +0.47 & 57 & $-$0.19 & $-$0.14 & $-$0.21 & +0.01 & +0.05 \\
11--20 & 60 & $-$0.29 & $-$0.01 & $-$0.24 & +0.09 & +0.14 & 57 & $-$0.36 & $-$0.15 & $-$0.29 & +0.07 & +0.07 \\
21--40 & 60 & $-$0.35 & $-$0.13 & $-$0.19 & +0.02 & +0.04 & 57 & $-$0.30 & +0.06 & $-$0.12 & +0.06 & +0.11 \\
41--80 & 60 & $-$0.28 & $-$0.01 & $-$0.16 & +0.05 & +0.09 & 59 & $-$0.06 & +0.12 & +0.03 & +0.03 & +0.07 \\
81--160 & 60 & $-$0.42 & $-$0.16 & $-$0.18 & +0.01 & +0.02 & 59 & $-$0.25 & $-$0.03 & $-$0.06 & +0.01 & +0.02 \\
$>$160 & 59 & $-$0.59 & $-$0.22 & $-$0.23 & 0.00 & +0.01 & 55 & $-$0.58 & $-$0.15 & $-$0.19 & +0.02 & +0.02 \\
\bottomrule
\end{tabular}}
\par\smallskip
\begin{minipage}{\textwidth}\footnotesize
Mean relative error of the single view and of the recursive count, and the parts of the latter that come from leaves with more than five objects, with one to five and with none. The three parts add up to the recursive error up to rounding. Crops: replayed crops per bin of $N$.
\end{minipage}
\end{table}

\paragraph{Limits of the model}
Eq.~\eqref{eq:response} gives every view the same response, but the spurious answer and visibility depart from it.

The spurious answer $s$ is not a single constant. On the first set, sub-windows without
the queried class are answered with 2.80 trucks, 1.63 cars, 0.50 storage tanks and 0.21 ships on average. On the
second set the values are 2.00, 1.50, 0.67 and 0.09.
Views with one to three objects, most of them sub-windows, are answered with 0.71 to 1.08 objects too many. The 120
root crops with one to five objects have a mean error of 0.00 objects. Eq.~\eqref{eq:response} has no term for this
difference between sub-windows and whole crops.

Eq.~\eqref{eq:response} also assumes that the model can see every object in the view.
Above 160 objects the observed amplification of one split is 2.11 on the first set and 2.14 on the second, against
1.33 and 1.32 predicted (Table~\ref{tab:supp:split}).
A fifth of the views with more than 160 objects are answered with at most five, against 6.5\% of all views with 11 or
more objects.
Table~\ref{tab:supp:trigger} shows how often a view beyond the horizon is split. Views with 6 to 10 objects
are often answered with at most five and left unsplit, but their answers are nearly unbiased. Beyond ten objects the
views left unsplit are mostly answered with zero, and the rule for zero answers splits those that are downsampled. Of the 75
views with more than ten objects that remain unsplit, 51 are answered with zero although they are shown at full
resolution.

\begin{table}[!tp]
\centering
\caption{How often a view beyond the horizon is split, by its number of objects (Qwen3-VL-8B).}
\label{tab:supp:trigger}
\small
\begin{tabular}{lccccc}
\toprule
Objects in view & Views & Split & Zero & 1--5 & Unsplit \\
\midrule
6--7 & 283 & 0.64 & 0.07 & 0.29 & 0.36 \\
8--10 & 355 & 0.84 & 0.07 & 0.10 & 0.16 \\
11--20 & 508 & 0.93 & 0.04 & 0.03 & 0.06 \\
21--40 & 425 & 0.95 & 0.03 & 0.02 & 0.05 \\
41--80 & 241 & 0.97 & 0.03 & 0.00 & 0.02 \\
81--160 & 134 & 0.93 & 0.07 & 0.01 & 0.04 \\
$>160$ & 85 & 0.80 & 0.20 & 0.00 & 0.12 \\
\bottomrule
\end{tabular}
\par\smallskip
\begin{minipage}{\textwidth}\footnotesize
First set of DOTA crops, every view of the recursion. Split: answered with more than five. Zero and 1--5: answered
with zero or with one to five objects. Unsplit: views not split by the final rule, which also splits zeros from downsampled views.
\end{minipage}
\end{table}
Zero answers are frequent in large crops. Among the crops with more than five objects, 30 of the 56 crops of 4{,}096
pixels are answered with zero, against 9 of the 147 crops of 256 pixels.
Such answers cannot enter the fit of $a$ and $\gamma$, which is made on logarithms. On the first set they are 62 of
the 359 replayed crops with more than five objects.
At a given number of objects, the error of a single view also grows with the side of the crop
(\appref{supp:horizon:resolution}). Fitted within one crop side over both sets, the exponent of Qwen3-VL-8B lies
between 0.84 and 0.99 for crops of 256 to 2{,}048 pixels.
Larger crops hold more objects in our sample, and crops of 2{,}048 pixels fall shorter than crops of 256 pixels
in every bin. The exponents fitted over all crops, 0.76 and 0.81, are therefore lower than within any of these
sides, and they absorb part of the loss of visibility.
Visibility enters Eq.~\eqref{eq:supp:decomp} in two ways. A view whose objects are too small to be seen may be
answered within the horizon and left unsplit, so that its shortfall falls into the first sum. When such a view is
split, its parts reveal objects, and their answers exceed the parent's by more than $A$. A part is never shown at a
smaller scale than its parent, because it covers less area under the same pixel budget and is enlarged to at least
768 pixels.
The VisDrone crops with reduced detail separate this effect from the undercount of crowded views (\appref{supp:horizon:visdrone}). Splitting a
downsampled view that is answered with zero (Section~\ref{sec:method:count}) handles the views in which downsampling hides
every object.

\paragraph{Fixed grids}
Subdivision splits a view only when its count exceeds the horizon. To separate this trigger from splitting as such,
we counted every node of the full split tree of each crop, split as the executor splits down to the 256-pixel cell,
with Qwen3-VL-8B on the 937 crops of both sets (33{,}329 views) and with Qwen2.5-VL-7B on the 479 crops of the first. Every strategy is
then replayed on the same answers. Fixed strategies split every view to a given depth, from 2$\times$2 parts down to
the 256-pixel cell, or into tiles of at most 1{,}024 or 512 pixels. Subdivision splits as in
Section~\ref{sec:method:count}, with thresholds $h$ from 1 to 20, with and without splitting zeros from downsampled views. The cost is the
number of VLM calls per crop, including the views that subdivision splits. With greedy decoding, the
9{,}162 views of Qwen3-VL-8B and 3{,}956 of Qwen2.5-VL-7B that the horizon measurement also counted receive the same
answers as there.

Table~\ref{tab:supp:grid} gives the results, and Fig.~\ref{fig:grid} plots those of Qwen3-VL-8B. For this model,
subdivision at $h=5$ and the 4$\times$4 grid need 9.4 and 9.0 calls per crop, and over all crops neither is
significantly more accurate. By a paired bootstrap over images, subdivision has a median error higher by 0.02
[$-0.01$, 0.07] and 1.5 [$-1.4$, 4.2] points fewer counts within 20\%. Above 20 objects the grid is better, by 0.04
[0.00, 0.06] in median error and 4.3 [1.3, 7.4] points within 20\%, the grid taking 10.8 calls per crop against 16.0. With at
most 20 objects the grid overcounts by 85\% on average, against 1\% for subdivision, a difference of 0.84 [0.63, 1.08].
Measured in time on one RTX 3090, the grid takes 2.9 seconds per crop and subdivision 3.8, since subdivision first
views the whole crop. Splitting zeros from downsampled views as well, as GeoScope does, subdivision needs 10.4 calls per crop and reaches the
grid's median error of 0.33. On the 249 crops from images that the first set does not use, subdivision and the grid
both reach 0.33 in median error, with 36.5\% and 36.9\% of the counts within 20\%. For Qwen2.5-VL-7B the comparison over all crops is again not
significant, 0.45 against 0.43 in median error and 28\% against 26\% within 20\%. Its fixed grids overcount sparse
crops more, a 4$\times$4 grid answering 5.5 times the truth for one or two objects and a grid of 256-pixel cells 14
times. Its subdivision also falls short above 20 objects, by 32\% on average, since this model undercounts even small
views (\appref{supp:horizon:resolution}). The threshold matters little for either model. From $h=3$ to $h=8$
Qwen3-VL-8B needs 10.1 to 9.0 calls per crop and its median error changes by 0.04, since 95\% of the splits at $h=5$
on the first set follow an answer above eight.

\begin{table}[!tp]
\centering
\caption{Fixed grids against count-triggered subdivision on the DOTA crops of the horizon measurement.}
\label{tab:supp:grid}
\footnotesize
\setlength{\tabcolsep}{4pt}
\begin{tabular}{lccccccc}
\toprule
 & & \multicolumn{3}{c}{Median} & & \multicolumn{2}{c}{Signed} \\
\cmidrule(lr){3-5}\cmidrule(lr){7-8}
Strategy & Calls & All & $\leq$20 & $>$20 & Within 20\% & $\leq$20 & $>$20 \\
\midrule
\multicolumn{8}{l}{\emph{Qwen3-VL-8B, 937 crops}} \\
One view & 1.0 & 0.50 & 0.49 & 0.57 & 27 & $-$0.16 & $-$0.35 \\
Fixed 2$\times$2 & 3.2 & 0.37 & 0.44 & 0.32 & 34 & +0.30 & $-$0.12 \\
Fixed 4$\times$4 & 9.0 & 0.33 & 0.50 & 0.25 & 35 & +0.85 & +0.05 \\
Fixed 8$\times$8 & 18.8 & 0.36 & 0.58 & 0.28 & 33 & +1.45 & +0.19 \\
256-pixel cells & 25.4 & 0.36 & 0.60 & 0.28 & 33 & +1.61 & +0.24 \\
Tiles $\leq$1{,}024 px & 2.4 & 0.42 & 0.43 & 0.41 & 30 & $-$0.03 & $-$0.24 \\
Tiles $\leq$512 px & 8.1 & 0.33 & 0.43 & 0.29 & 36 & +0.39 & $-$0.02 \\
Subdivision, $h=3$ & 10.1 & 0.37 & 0.50 & 0.30 & 34 & +0.03 & $-$0.06 \\
Subdivision, $h=5$ & 9.4 & 0.35 & 0.50 & 0.29 & 34 & +0.01 & $-$0.06 \\
Subdivision, $h=8$ & 9.0 & 0.33 & 0.50 & 0.29 & 34 & 0.00 & $-$0.06 \\
Subdivision, $h=5$, zeros split & 10.4 & 0.33 & 0.50 & 0.27 & 34 & +0.02 & $-$0.04 \\
\midrule
\multicolumn{8}{l}{\emph{Qwen2.5-VL-7B, 479 crops of the first set}} \\
One view & 1.0 & 0.63 & 0.49 & 0.68 & 19 & $-$0.12 & $-$0.62 \\
Fixed 2$\times$2 & 3.2 & 0.50 & 0.57 & 0.46 & 24 & +0.51 & $-$0.43 \\
Fixed 4$\times$4 & 9.0 & 0.43 & 0.81 & 0.33 & 26 & +1.88 & $-$0.20 \\
Fixed 8$\times$8 & 19.2 & 0.46 & 0.90 & 0.32 & 27 & +4.54 & +0.02 \\
256-pixel cells & 26.6 & 0.48 & 0.90 & 0.35 & 27 & +5.77 & +0.09 \\
Tiles $\leq$1{,}024 px & 2.4 & 0.51 & 0.41 & 0.57 & 23 & +0.07 & $-$0.52 \\
Tiles $\leq$512 px & 8.1 & 0.45 & 0.50 & 0.42 & 25 & +1.06 & $-$0.31 \\
Subdivision, $h=3$ & 9.7 & 0.46 & 0.67 & 0.41 & 28 & +0.18 & $-$0.29 \\
Subdivision, $h=5$ & 8.5 & 0.45 & 0.63 & 0.40 & 28 & +0.12 & $-$0.32 \\
Subdivision, $h=8$ & 8.0 & 0.43 & 0.54 & 0.39 & 29 & +0.09 & $-$0.33 \\
Subdivision, $h=5$, zeros split & 9.7 & 0.43 & 0.60 & 0.39 & 28 & +0.18 & $-$0.29 \\
\bottomrule
\end{tabular}
\par\smallskip
\begin{minipage}{\textwidth}\footnotesize
Every strategy is replayed on one set of answers for every part of every crop. Calls: VLM calls per crop, including
the views that subdivision splits. Median: median absolute relative error over all crops, over those with at most 20
objects and over those with more. Within 20\%: share of all crops counted within 20\% of the truth. Signed: mean
signed relative error.
\end{minipage}
\end{table}

\section{Details and controls of the counting-horizon measurement}
\label{supp:horizon}
\subsection{Crops, exclusions and fitted responses}
\label{supp:horizon:crops}
\paragraph{Images}
The crops come from the training part of DOTA-v1.5~\citep{dota,dotav2}, whose annotations give every instance as a
quadrilateral. We use four classes and ask for them by everyday names. Small vehicles are asked for as cars and large
vehicles as trucks, while ships and storage tanks keep their names.
The true count $N$ of a crop is the number of annotated instances of the queried class whose centre, the mean of the
four corners, lies inside the crop.
Before sampling we remove every image whose DOTA identifier occurs in a benchmark question file, so no crop image, of
either set or of \appref{supp:horizon:prompting}, occurs in XLRS-Bench or MME-RealWorld (\appref{supp:protocol:leakage}).
Among the remaining images, those with at least three instances of the four classes form the pool. We keep the 130
with the most instances and 130 drawn at random from the rest with seed 0, so that sparse scenes are covered as well.

\paragraph{Crops}
Crops are drawn for each pair of image and class, in an order shuffled with the seed of the set. For every side of
256, 512, 1{,}024, 2{,}048 and 4{,}096 pixels that does not exceed the longer side of the image, one instance of the class is
picked at random. The crop is placed so that this instance lies between 20\% and 80\% of the side from its upper left
corner in each direction.
The crop is then shifted into the image and clipped where the image is narrower than the side. This leaves 44 crops
of the first set and 66 of the second non-square.
A crop joins the bin of its $N$ until that bin holds 60 crops, and sampling stops when all eight bins are full.
The first set uses seed 0 and the second seed 1. Each holds 480 crops, from 112 and 130 images respectively, with 63
images in common, and 19 crops of the second set coincide with crops of the first.
The first set holds 188 crops of cars, 119 of trucks, 111 of ships and 62 of storage tanks, the second 241, 125, 87
and 27.
Larger crops hold more objects. Over both sets the median $N$ is 8 for crops of 256 pixels, 33 for crops of 1{,}024
pixels and 108 for crops of 4{,}096 pixels.
The replay uses the windows of 479 of the 480 crops of the first set and 477 of the second.
InternVL3.5 and LLaVA-OneVision are evaluated on these 479 crops of the first set and have 59 crops above 160
objects. The other Qwen models use all 480 crops of the first set, drawn with the same seed.

\paragraph{Views and answers}
The native view scales a crop down to at most $2048^2$ pixels, so square crops of 4{,}096 pixels are shown at half
resolution.
A crop whose longer side is below 768 pixels is enlarged to 768 pixels, so crops of 256 and 512 pixels are shown three
and one and a half times enlarged.
For the Qwen models the processor's own size limits are lifted, so that the model receives exactly this image.
InternVL3.5 and LLaVA-OneVision resize or tile it with their own processors.
The question names the class and asks for a single integer (\appref{supp:method:prompts}). Decoding is greedy with at
most 12 new tokens. The answer is the first integer in the reply, or else the smallest number word from zero to twelve in it. A reply
with neither is unanswered and left out of the mean of its bin.
For InternVL3.5 and LLaVA-OneVision we add the default system message of each model's own chat format, which their
transformers chat templates leave out.
InternVL3.5-8B leaves 7 of its 479 answers without a number, and the other models answer every crop. InternVL3.5-2B,
run in the same way, leaves 124 of 479 crops unanswered and is not reported.
The subdivision in Fig.~\ref{fig:horizon}b follows Algorithm~\ref{alg:horizon} with $h=5$, depth limit 5 and minimum cell
256 pixels. If any part of a view is unanswered, the view's own answer is kept. The rule for zero answers from
downsampled views is not applied in these measurements.

\paragraph{Models}
Qwen3-VL-2B, 4B and 8B~\citep{qwen3vl}, Qwen2.5-VL-7B~\citep{qwen25vl}, InternVL3.5-4B and 8B~\citep{internvl35} and
LLaVA-OneVision-7B~\citep{llavaov} run locally in bfloat16, the last three from their transformers checkpoints.
Qwen3-VL-32B is queried through a commercial API with the same views and question. Only Qwen3-VL-8B is run on the
second set.

\paragraph{Per-model curves and fitted responses}
Table~\ref{tab:supp:models} gives the curves of Fig.~\ref{fig:horizon} for every model, and Table~\ref{tab:fit}
the fitted response. Taking the first bin whose mean error falls below $-0.2$ as the onset, five of the eight models undercount from
6 to 10 objects, Qwen3-VL-2B from three to five and both InternVL3.5 models from 41 to 80.
On the second set alone, Qwen3-VL-8B falls below this mark at three to five objects, with $-0.27$.
InternVL3.5-4B and 8B overcount one or two objects by 91\% and 127\% on average, and subdivision raises their surplus,
for the 8B model to $+1.07$ at 6 to 10 objects.
A quarter of the answers of Qwen3-VL-8B to crops with one or two objects are zero. Its mean error there is close to
zero because overcounts offset them.
The fit of $a$ and $\gamma$ uses the unsplit crops with more than five objects and a non-zero answer. The spurious
answer $s$ is the mean answer to the replayed views that contain no instance of the queried class. All such views are
sub-windows, since every crop holds at least one instance.
For Qwen3-VL-32B, queried through an API, $a$ and $\gamma$ are fitted on all its crops in the same way, and its
recorded answers are replayed like those of the other models.

\begin{table}[!tp]
\centering
\caption{Mean signed relative error per bin of the true count $N$ for every model, in one view (native) and with
subdivision at $h=5$ (split).}
\label{tab:supp:models}
\footnotesize
\setlength{\tabcolsep}{2.5pt}
\resizebox{\textwidth}{!}{%
\begin{tabular}{llcccccccc}
\toprule
Model & View & 1--2 & 3--5 & 6--10 & 11--20 & 21--40 & 41--80 & 81--160 & $>$160 \\
\midrule
Qwen3-VL-2B & native & $-$0.01 & $-$0.33 & $-$0.36 & $-$0.51 & $-$0.60 & $-$0.65 & $-$0.75 & $-$0.64 \\
 & split & $-$0.01 & $-$0.25 & $-$0.23 & $-$0.40 & $-$0.40 & $-$0.38 & $-$0.48 & $-$0.05 \\
Qwen3-VL-4B & native & +0.14 & +0.04 & $-$0.35 & $-$0.42 & $-$0.48 & $-$0.52 & $-$0.56 & $-$0.75 \\
 & split & +0.21 & +0.11 & +0.14 & $-$0.15 & $-$0.21 & $-$0.16 & $-$0.29 & $-$0.24 \\
Qwen3-VL-8B, first set & native & +0.07 & 0.00 & $-$0.25 & $-$0.29 & $-$0.35 & $-$0.28 & $-$0.42 & $-$0.60 \\
 & split & +0.17 & +0.08 & +0.29 & $-$0.03 & $-$0.13 & $-$0.01 & $-$0.16 & $-$0.23 \\
Qwen3-VL-8B, second set & native & 0.00 & $-$0.27 & $-$0.19 & $-$0.38 & $-$0.32 & $-$0.04 & $-$0.26 & $-$0.55 \\
 & split & 0.00 & $-$0.28 & $-$0.14 & $-$0.18 & +0.03 & +0.13 & $-$0.04 & $-$0.17 \\
Qwen3-VL-8B, both sets & native & +0.04 & $-$0.14 & $-$0.22 & $-$0.33 & $-$0.34 & $-$0.16 & $-$0.34 & $-$0.57 \\
 & split & +0.09 & $-$0.10 & +0.07 & $-$0.10 & $-$0.05 & +0.06 & $-$0.10 & $-$0.20 \\
Qwen3-VL-32B & native & +0.45 & +0.01 & $-$0.24 & $-$0.32 & $-$0.37 & $-$0.46 & $-$0.51 & $-$0.63 \\
 & split & +0.51 & +0.13 & $-$0.21 & $-$0.04 & $-$0.21 & $-$0.21 & $-$0.33 & $-$0.37 \\
Qwen2.5-VL-7B & native & +0.51 & $-$0.15 & $-$0.37 & $-$0.49 & $-$0.50 & $-$0.61 & $-$0.63 & $-$0.77 \\
 & split & +0.68 & +0.01 & $-$0.07 & $-$0.22 & $-$0.22 & $-$0.35 & $-$0.32 & $-$0.42 \\
InternVL3.5-4B & native & +0.91 & +0.09 & $-$0.04 & $-$0.18 & $-$0.18 & $-$0.28 & $-$0.33 & $-$0.49 \\
 & split & +1.00 & +0.44 & +0.11 & +0.13 & $-$0.02 & $-$0.01 & $-$0.11 & $-$0.13 \\
InternVL3.5-8B & native & +1.27 & +0.53 & +0.03 & $-$0.07 & $-$0.11 & $-$0.21 & $-$0.38 & $-$0.43 \\
 & split & +1.52 & +1.38 & +1.07 & +0.31 & +0.27 & +0.03 & +0.03 & $-$0.05 \\
LLaVA-OV-7B & native & +0.57 & +0.05 & $-$0.27 & $-$0.40 & $-$0.52 & $-$0.70 & $-$0.76 & $-$0.85 \\
 & split & +0.84 & +0.38 & +0.11 & $-$0.01 & $-$0.11 & $-$0.20 & $-$0.36 & $-$0.39 \\
\midrule
Qwen3-VL-8B, both sets & undercounts (\%) & 28 & 58 & 52 & 72 & 73 & 58 & 80 & 92 \\
 & round answers (\%) & 1 & 11 & 33 & 35 & 56 & 65 & 77 & 71 \\
 & zero answers (\%) & 24 & 22 & 21 & 18 & 13 & 8 & 11 & 24 \\
\bottomrule
\end{tabular}}
\par\smallskip
\begin{minipage}{\textwidth}\footnotesize
First set unless stated. Crops: 60 per bin and set, 59 above 160 for InternVL3.5 and LLaVA-OneVision. InternVL3.5-8B
leaves 2 crops at 41 to 80 and 5 above 160 unanswered. Bottom rows: shares of undercounts, of round answers, which are
multiples of ten from 10 on, and of zero answers, for Qwen3-VL-8B in a single view on both sets.
\end{minipage}
\end{table}
\subsection{Native versus resized views}
\label{supp:horizon:resolution}
The reduced view renders a crop with at most $1024^2$ pixels instead of $2048^2$ and is otherwise identical to the
native view.
Crops whose longer side is below 768 pixels are enlarged to 768 pixels in both views, and crops of 1{,}024 pixels fit
both budgets, so the two views coincide for crops of 256, 512 and 1{,}024 pixels. With greedy decoding, all 722 such crops of the two sets receive the same answer in
both views.
The views differ only for the 238 crops of 2{,}048 and 4{,}096 pixels, which the reduced view shows at half the scale of the
native view.

Over all crops, the reduced view changes the mean error of a bin by at most 0.07 on the first set and 0.11 on the second
(Table~\ref{tab:supp:fixed}).
On the large crops it lowers the mean error from $-0.61$ to $-0.72$, and it lowers it in seven of the eight bins.
These crops fall short already in the native view, by 0.46 at one or two objects and by 0.71 at three to five.
The near agreement of the two views over all crops therefore says little about the role of pixel size.

Table~\ref{tab:supp:side} breaks down the native view by the side of the crop. At a given number of objects the error
grows with the side. Crops of 256 pixels, shown three times enlarged, stay within $\pm0.21$ in every bin up to 160
objects. Crops of 2{,}048 pixels, shown at native resolution, fall short by 0.41 to 0.74 in every bin.
Crops of 1{,}024 and 2{,}048 pixels are shown at the same scale, and the larger ones fall short more in six of the eight
bins. At equal object size the error thus grows with the extent of the view. The reduced view shows that at equal
extent it also grows as objects get smaller.
In our crops the number of objects grows with the side, so the curves of Fig.~\ref{fig:horizon}a and b mix the
effect of the number of objects with that of the side (\appref{supp:model}).

The other Qwen models were also run in the reduced view, on the first set. On its large crops the reduced view lowers the
mean error of Qwen3-VL-2B from $-0.73$ to $-0.83$, of Qwen3-VL-4B from $-0.63$ to $-0.71$ and of Qwen3-VL-32B from
$-0.63$ to $-0.67$. It raises that of Qwen2.5-VL-7B from $-0.80$ to $-0.77$.
Qwen3-VL-32B gives different answers to 63 of the 360 small crops that the two views show
identically.

\begin{table}[!tp]
\centering
\caption{Mean signed relative error of Qwen3-VL-8B in the native view, at most $2048^2$ pixels, and in the reduced view, at most $1024^2$ pixels.}
\label{tab:supp:fixed}
\footnotesize
\setlength{\tabcolsep}{3pt}
\begin{tabular}{llcccccccc}
\toprule
Crops & View & 1--2 & 3--5 & 6--10 & 11--20 & 21--40 & 41--80 & 81--160 & $>$160 \\
\midrule
First set, all & native & +0.07 & 0.00 & $-$0.25 & $-$0.29 & $-$0.35 & $-$0.28 & $-$0.42 & $-$0.60 \\
 & fixed & +0.07 & +0.02 & $-$0.32 & $-$0.31 & $-$0.36 & $-$0.30 & $-$0.43 & $-$0.64 \\
Second set, all & native & 0.00 & $-$0.27 & $-$0.19 & $-$0.38 & $-$0.32 & $-$0.04 & $-$0.26 & $-$0.55 \\
 & fixed & $-$0.03 & $-$0.28 & $-$0.20 & $-$0.38 & $-$0.33 & $-$0.08 & $-$0.31 & $-$0.66 \\
\midrule
Both sets, large & native & $-$0.46 & $-$0.71 & $-$0.45 & $-$0.74 & $-$0.72 & $-$0.53 & $-$0.65 & $-$0.61 \\
 & fixed & $-$0.61 & $-$0.67 & $-$0.65 & $-$0.81 & $-$0.77 & $-$0.65 & $-$0.76 & $-$0.73 \\
 & crops & 14 & 21 & 21 & 28 & 17 & 32 & 30 & 75 \\
\bottomrule
\end{tabular}
\par\smallskip
\begin{minipage}{\textwidth}\footnotesize
All crops: 60 per bin and set. Large crops: the crops of 2{,}048 and 4{,}096 pixels of both sets, the only ones the two views show differently, with their number per bin.
\end{minipage}
\end{table}

\begin{table}[!tp]
\centering
\caption{Mean signed relative error of Qwen3-VL-8B in the native view by crop side (pixels), both sets.}
\label{tab:supp:side}
\footnotesize
\setlength{\tabcolsep}{3pt}
\resizebox{\textwidth}{!}{%
\begin{tabular}{lccccccccccccc}
\toprule
Side & Scale & & 1--2 & 3--5 & 6--10 & 11--20 & 21--40 & 41--80 & 81--160 & $>$160 & Zero answers & $a$ & $\gamma$ \\
\midrule
256 & 3 & error & +0.21 & +0.04 & $-$0.08 & $-$0.06 & +0.02 & +0.16 & $-$0.13 & +0.88 & 9 of 147 & 0.95 & 0.99 \\
 & & crops & 48 & 42 & 45 & 34 & 31 & 22 & 12 & 3 & & & \\
512 & 1.5 & error & +0.19 & +0.10 & $-$0.06 & $-$0.23 & $-$0.30 & 0.00 & $-$0.24 & $-$0.60 & 11 of 182 & 1.21 & 0.84 \\
 & & crops & 37 & 22 & 33 & 37 & 35 & 34 & 34 & 9 & & & \\
1{,}024 & 1 & error & $-$0.29 & $-$0.15 & $-$0.55 & $-$0.41 & $-$0.49 & $-$0.18 & $-$0.27 & $-$0.61 & 26 of 188 & 0.97 & 0.84 \\
 & & crops & 21 & 35 & 21 & 21 & 37 & 32 & 44 & 33 & & & \\
2{,}048 & 1 & error & $-$0.42 & $-$0.65 & $-$0.45 & $-$0.74 & $-$0.71 & $-$0.41 & $-$0.57 & $-$0.49 & 39 of 147 & 0.50 & 0.95 \\
 & & crops & 13 & 17 & 16 & 22 & 13 & 23 & 22 & 51 & & & \\
4{,}096 & 0.5 & error & $-$1.00 & $-$1.00 & $-$0.45 & $-$0.75 & $-$0.77 & $-$0.83 & $-$0.88 & $-$0.87 & 30 of 56 & 0.96 & 0.74 \\
 & & crops & 1 & 4 & 5 & 6 & 4 & 9 & 8 & 24 & & & \\
\bottomrule
\end{tabular}}
\par\smallskip
\begin{minipage}{\textwidth}\footnotesize
The number of crops is given underneath. Scale: magnification at which a square crop is shown. Zero answers: crops
with more than five objects answered with zero. $a$ and $\gamma$: fit of the response within the side, over the 138,
171, 162, 108 and 26 crops with more than five objects and a non-zero answer.
\end{minipage}
\end{table}

Table~\ref{tab:supp:sidemodels} repeats the breakdown by crop side for every model, over the crops with 6 to 40 objects
and sides up to 2{,}048 pixels. Every model falls short more in views of 2{,}048 pixels than of 256 pixels,
InternVL3.5-8B by 0.07 and the others by 0.27 to 0.59. Qwen3-VL-8B and
both InternVL3.5 models stay within 5\% in the smallest views, while the other models fall short by 14 to 33\% even
there.

\begin{table}[!tp]
\centering
\caption{Mean signed relative error in one view by the side of the crop, for crops with 6 to 40 objects.}
\label{tab:supp:sidemodels}
\footnotesize
\begin{tabular}{lcccc}
\toprule
Model & 256 px, $\times3$ & 512 px, $\times1.5$ & 1{,}024 px, $\times1$ & 2{,}048 px, $\times1$ \\
\midrule
Qwen3-VL-2B & $-$0.33 & $-$0.42 & $-$0.65 & $-$0.61 \\
Qwen3-VL-4B & $-$0.30 & $-$0.35 & $-$0.56 & $-$0.57 \\
Qwen3-VL-8B & $-$0.05 & $-$0.20 & $-$0.48 & $-$0.64 \\
Qwen2.5-VL-7B & $-$0.28 & $-$0.36 & $-$0.61 & $-$0.65 \\
InternVL3.5-4B & $-$0.01 & $-$0.06 & $-$0.20 & $-$0.38 \\
InternVL3.5-8B & $-$0.01 & $-$0.02 & $-$0.11 & $-$0.08 \\
LLaVA-OV-7B & $-$0.25 & $-$0.28 & $-$0.58 & $-$0.60 \\
Qwen3-VL-32B & $-$0.14 & $-$0.24 & $-$0.42 & $-$0.52 \\
\bottomrule
\end{tabular}
\par\smallskip
\begin{minipage}{\textwidth}\footnotesize
The factor is the enlargement with which a crop is shown. Crops per side: 52, 57, 40 and 23 for each model, 110, 105, 79 and 51 for Qwen3-VL-8B, whose two crop sets are pooled.
\end{minipage}
\end{table}
\subsection{Prompting variants for counting}
\label{supp:horizon:prompting}
We test two alternatives to the direct question that might count more exactly, on a separate set of 240 crops with
at most 20 objects. It holds 60 crops in each of the four bins up to 20 objects, drawn by the procedure of
\appref{supp:horizon:crops} from 87 images, and all three conditions see the same crops in the native view.
The enumerate-first prompt asks Qwen3-VL-8B to list every instance with its approximate position, one per line, and
to write the total on the last line as `Total: N'. It may generate up to 512 tokens, and the answer is the last total,
or else the last integer.
Qwen3-VL-8B-Thinking receives the direct question and may generate up to 2{,}048 tokens. Its answer is read from the
text after its reasoning.

Table~\ref{tab:supp:prompting} gives the share of exact answers and the mean signed error. Neither alternative is
significantly more exact than the direct question. Over the 240 crops, the Thinking model answers 4.6 points more crops exactly,
with a paired 95\% bootstrap interval of $[-0.4,+10.0]$, and enumerate-first 1.3 points, $[-4.6,+7.1]$.
Enumerate-first and the Thinking model undercount more at 11 to 20 objects, by 0.59 and 0.49 against 0.16 for the direct
question.
Enumerate-first leaves 24 crops without a usable number, and the Thinking model takes about ten times as long as the direct
question.

Visual programs often count by enumeration, as the length of a list of detected objects~\citep{vipergpt}. We compare
this with asking for a count on the first set, using 30 crops per bin drawn at random with seed 0.
Enumeration asks for a box around every instance of the class, in the grounding format of GeoScope. We allow up to
4{,}096 new tokens, which hold about 125 boxes, against 1{,}024 in GeoScope, and drop every box that overlaps an earlier one
with an IoU above 0.7.
The four conditions are a count in one view, enumeration in one view, the count-triggered subdivision with $h=5$, and
enumeration in every leaf of that subdivision. Table~\ref{tab:supp:enum} gives the results.

In one view, enumeration degenerates into long lists. Its output reaches the token limit for 17\% of the crops with
three to five objects and for 57\% to 73\% of the crops with more than 40.
At three to five objects, a fifth of the lists hold more than three times as many boxes as there are objects, up to
128, so the mean error of enumeration there is $+4.45$.
On the leaves, where every view is small, enumeration is still worse than counting. Above five objects its median
absolute relative error is 0.58 to 0.77, against 0.23 to 0.54 for counts on the same leaves.
GeoScope therefore asks for an integer in every leaf.

\begin{table}[!tp]
\centering
\caption{Direct and enumerate-first prompts to Qwen3-VL-8B and the direct prompt to Qwen3-VL-8B-Thinking, on 240
DOTA-v1.5 crops with at most 20 objects.}
\label{tab:supp:prompting}
\footnotesize
\setlength{\tabcolsep}{3pt}
\resizebox{\textwidth}{!}{%
\begin{tabular}{lcccccccccc}
\toprule
 & \multicolumn{5}{c}{Exact (\%)} & \multicolumn{4}{c}{Mean error} & \\
\cmidrule(lr){2-6}\cmidrule(lr){7-10}
Condition & 1--2 & 3--5 & 6--10 & 11--20 & All & 1--2 & 3--5 & 6--10 & 11--20 & Unanswered \\
\midrule
Direct question & 33 & 15 & 7 & 10 & 16.3 & +0.08 & $-$0.15 & $-$0.21 & $-$0.16 & 0 \\
Enumerate-first & 40 & 20 & 8 & 2 & 17.5 & +0.50 & $-$0.26 & $-$0.42 & $-$0.59 & 24 \\
Qwen3-VL-8B-Thinking & 48 & 20 & 12 & 3 & 20.8 & +0.10 & $-$0.16 & $-$0.32 & $-$0.49 & 1 \\
\bottomrule
\end{tabular}}
\par\smallskip
\begin{minipage}{\textwidth}\footnotesize
60 crops per bin. Exact: share of crops answered exactly, with unanswered crops counted as wrong. Mean error: mean
signed relative error over the answered crops. Unanswered: crops without a usable number.
\end{minipage}
\end{table}

\begin{table}[!tp]
\centering
\caption{Counting and enumerating by grounding with Qwen3-VL-8B, 30 crops of the first set per bin.}
\label{tab:supp:enum}
\footnotesize
\setlength{\tabcolsep}{3pt}
\resizebox{\textwidth}{!}{%
\begin{tabular}{llcccccccc}
\toprule
Measure & Condition & 1--2 & 3--5 & 6--10 & 11--20 & 21--40 & 41--80 & 81--160 & $>$160 \\
\midrule
Mean signed relative error & count, one view & $-$0.13 & $-$0.01 & $-$0.18 & $-$0.29 & $-$0.39 & $-$0.21 & $-$0.48 & $-$0.64 \\
 & enumerate, one view & +0.50 & +4.45 & +1.55 & +1.30 & +0.58 & +0.19 & $-$0.37 & $-$0.75 \\
 & count, leaves & $-$0.13 & +0.26 & +0.36 & $-$0.13 & $-$0.23 & +0.01 & $-$0.27 & $-$0.24 \\
 & enumerate, leaves & +0.50 & +4.77 & +3.43 & +0.94 & +0.47 & +1.41 & +0.28 & +0.38 \\
\midrule
Median absolute relative error & count, one view & 0.00 & 0.55 & 0.42 & 0.22 & 0.53 & 0.77 & 0.34 & 0.83 \\
 & enumerate, one view & 0.00 & 0.45 & 0.83 & 0.62 & 0.94 & 0.77 & 0.50 & 0.90 \\
 & count, leaves & 0.00 & 0.55 & 0.54 & 0.23 & 0.28 & 0.31 & 0.31 & 0.40 \\
 & enumerate, leaves & 0.00 & 0.50 & 0.77 & 0.71 & 0.58 & 0.71 & 0.69 & 0.73 \\
\midrule
Token limit (\%) & enumerate, one view & 0 & 17 & 23 & 30 & 33 & 70 & 73 & 57 \\
 & enumerate, leaves & 0 & 17 & 20 & 17 & 13 & 47 & 60 & 67 \\
\bottomrule
\end{tabular}}
\par\smallskip
\begin{minipage}{\textwidth}\footnotesize
One view: the native view of the crop. Leaves: the leaves of the count-triggered subdivision with $h=5$. Token limit: share of crops for which an enumeration reached the limit of 4{,}096 new tokens.
\end{minipage}
\end{table}
\subsection{Drone imagery}
\label{supp:horizon:visdrone}
VisDrone2019-DET~\citep{visdrone} holds drone images of Chinese cities. Its vehicles are larger in pixels than those
of our DOTA crops, with a median size of 30 against 18 pixels.
We ask for cars, which merge the cars and vans of VisDrone as DOTA's small vehicles do, and for trucks. Of the images
with at least three such vehicles we take the 150 densest and 150 at random, with seed 0.
Crops are drawn as on DOTA with sides of 256 to 2{,}048 pixels, except that a crop touching an annotated ignore region is
discarded, since vehicles there are not annotated.
No crop of 2{,}048 pixels remained, and the bins above 40 objects could not be filled. The 373 crops, 287 of cars and 86
of trucks, come from 179 images. The bins up to 40 objects hold 60 crops each, the bin of 41 to 80 holds 41 and that
of 81 to 160 holds 32.
Qwen3-VL-8B answers the same question in the native view, with and without subdivision at $h=5$.
To reduce the detail per object, each image is downsampled by a factor and upsampled back to its size with bicubic
interpolation before the crops are cut. Windows and views stay the same. A factor of 1.66 brings the median vehicle
size to that of the DOTA crops, and a factor of 3 is a stronger test.
Table~\ref{tab:supp:visdrone} also gives the DOTA crops of cars and trucks with the same sides, 256 to 1{,}024 pixels,
from both sets.

At full detail the single view falls short by at most 0.06 up to 20 objects, by 0.14 at 21 to 40 and by 0.23 at 81 to
160. Undercounts rise from 17\% of the answers at one or two objects to 69\% at 81 to 160. Round numbers are absent up
to five objects and make up 22\% of the answers at 6 to 10 and 81\% at 81 to 160.
Against the DOTA vehicle crops of the same sides, the single view falls short less from 6 to 40 objects, by 0.03 to
0.14 against 0.13 to 0.35. It falls short more at 41 to 80 objects, 0.20 against 0.04, and about as much at 81 to
160, 0.23 against 0.29.
Subdivision brings every bin within $\pm0.09$ and lowers the share of round answers to at most 22\%.

Less detail moves the shortfall to fewer objects and deepens it. At three to five objects the single view falls short
by 0.01 at full detail and by 0.09 and 0.11 at factors of 1.66 and 3. At 81 to 160 objects it falls short by 0.23,
0.25 and 0.45.
Round answers do not follow the detail. At every level they make up about a fifth of the answers at 6 to 10
objects and 81\% to 97\% at 81 to 160.
Subdivision removes the shortfall at a factor of 1.66, leaving errors within $\pm0.03$ from 6 to 80 objects. At a
factor of 3 it leaves $-0.08$ to $-0.30$ from 6 to 160 objects, and crops with three to five objects, which are not
split, already fall short by 0.11.
Subdivision thus removes the shortfall at full detail and at a factor of 1.66, and part of it at a factor of 3. A factor of 1.66
moves the shortfall
only a little towards fewer objects. From 6 to 40 objects the single view then falls short by 0.07 to 0.19, against
0.13 to 0.35 for the DOTA vehicle crops of the same sides.
Object size alone therefore does not explain the later onset of the shortfall on drone images.

\begin{table}[!tp]
\centering
\caption{Mean signed relative error of Qwen3-VL-8B on VisDrone crops at full and reduced detail, in one view (native)
and with subdivision at $h=5$ (split).}
\label{tab:supp:visdrone}
\footnotesize
\setlength{\tabcolsep}{3pt}
\resizebox{\textwidth}{!}{%
\begin{tabular}{llccccccc}
\toprule
Crops & View & 1--2 & 3--5 & 6--10 & 11--20 & 21--40 & 41--80 & 81--160 \\
\midrule
VisDrone, full detail & native & +0.09 & $-$0.01 & $-$0.03 & $-$0.06 & $-$0.14 & $-$0.20 & $-$0.23 \\
 & split & +0.09 & $-$0.01 & +0.06 & +0.06 & +0.06 & +0.03 & $-$0.08 \\
VisDrone, factor 1.66 & native & +0.08 & $-$0.09 & $-$0.07 & $-$0.11 & $-$0.19 & $-$0.21 & $-$0.25 \\
 & split & +0.08 & $-$0.09 & 0.00 & $-$0.01 & $-$0.03 & $-$0.02 & $-$0.17 \\
VisDrone, factor 3 & native & $-$0.06 & $-$0.11 & $-$0.14 & $-$0.17 & $-$0.33 & $-$0.36 & $-$0.45 \\
 & split & $-$0.06 & $-$0.11 & $-$0.09 & $-$0.08 & $-$0.15 & $-$0.21 & $-$0.30 \\
DOTA, cars and trucks & native & +0.24 & +0.03 & $-$0.13 & $-$0.24 & $-$0.35 & $-$0.04 & $-$0.29 \\
 & split & +0.32 & +0.01 & $-$0.01 & $-$0.02 & $-$0.11 & +0.14 & $-$0.10 \\
\midrule
VisDrone crops & & 60 & 60 & 60 & 60 & 60 & 41 & 32 \\
DOTA crops & & 68 & 67 & 70 & 65 & 79 & 59 & 66 \\
VisDrone, full detail & undercounts (\%) & 17 & 27 & 37 & 53 & 70 & 63 & 69 \\
 & round answers (\%) & 0 & 0 & 22 & 25 & 52 & 71 & 81 \\
\bottomrule
\end{tabular}}
\par\smallskip
\begin{minipage}{\textwidth}\footnotesize
Detail is reduced by factors of 1.66 and 3. DOTA: crops of cars and trucks with sides of 256 to 1{,}024 pixels from
both sets. No VisDrone crop holds more than 160 objects.
\end{minipage}
\end{table}
\subsection{FAIR1M and DIOR}
\label{supp:horizon:sources}
From the public part of FAIR1M~\citep{fair1m} we use the 1{,}732 annotated images, from DIOR~\citep{dior} the 11{,}725
images of its training and validation archive. From each we take 300 images with at least three target objects, half
the densest and half at random. On FAIR1M the crops are centred on cars, trucks and ships of every type. Its cars are the
small cars and vans, and its trucks the cargo trucks, dump trucks and truck tractors. On DIOR they are centred on
vehicles, ships and storage tanks. Bins, the question and the model are those of the DOTA measurement, and subdivision
uses $h=5$. Crops of 256 to 2{,}048 pixels that fit in the image are drawn as on DOTA, and the whole image is added as
one more crop. The images are small, mostly 1{,}000 pixels on a side on FAIR1M and 800 on DIOR, so the crops are 256
to 1{,}024 pixels wide or whole images of 800 to 1{,}500 pixels.
Pooled over classes, the model falls short on FAIR1M by 38 to 59\% in every bin from three objects on except 41 to
80, and on DIOR it overcounts small sets and is close to unbiased up to 160 objects. Table~\ref{tab:supp:sources} gives
the error per class. This breakdown and the checks below were made after the pooled results were seen.
We looked at 16 random crops of each dataset with the annotated objects marked. Most FAIR1M truck annotations are small
vehicles that look like cars from above, in crops where the model, asked for trucks, answers zero or one. On DIOR,
vehicle crops near marinas appear to be answered with the number of boats, and some crops of coarse images are too
blurred for the object to be seen. A few FAIR1M and DIOR crops also hold visible objects of the queried class that are not annotated.
For cars on FAIR1M and ships on DIOR, whose names match their appearance, the errors follow the pattern described in
Section~\ref{sec:horizon:measure}, with a later onset than on DOTA.

\begin{table}[!tp]
\centering
\caption{Mean signed relative error per class on FAIR1M and DIOR crops, Qwen3-VL-8B, single view (native) and with
subdivision at $h=5$ (split); crops per bin in parentheses.}
\label{tab:supp:sources}
\footnotesize
\setlength{\tabcolsep}{2.5pt}
\resizebox{\textwidth}{!}{%
\begin{tabular}{llcccccccc}
\toprule
Class & View & 1--2 & 3--5 & 6--10 & 11--20 & 21--40 & 41--80 & 81--160 & $>$160 \\
\midrule
FAIR1M car & native & +52.00 (4) & $-$0.16 (8) & $-$0.01 (12) & $-$0.11 (33) & $-$0.31 (49) & +0.04 (58) & $-$0.36 (58) & $-$0.55 (57) \\
 & split & +7.50 (4) & $-$0.16 (8) & $-$0.03 (12) & $-$0.03 (33) & 0.00 (49) & +0.12 (58) & $-$0.11 (58) & $-$0.19 (57) \\
FAIR1M truck & native & $-$0.51 (40) & $-$0.52 (38) & $-$0.78 (41) & $-$0.84 (24) & $-$0.95 (11) & $-$0.84 (1) & -- & $-$0.67 (2) \\
 & split & $-$0.51 (40) & $-$0.46 (38) & $-$0.69 (41) & $-$0.85 (24) & $-$0.94 (11) & $-$0.79 (1) & -- & $-$0.31 (2) \\
FAIR1M ship & native & $-$0.22 (16) & $-$0.15 (14) & $-$0.53 (7) & $-$0.80 (3) & -- & +0.92 (1) & $-$0.93 (2) & $-$1.00 (1) \\
 & split & $-$0.22 (16) & $-$0.29 (14) & $-$0.53 (7) & $-$0.80 (3) & -- & +0.92 (1) & +0.27 (2) & $-$1.00 (1) \\
\midrule
DIOR ship & native & +0.08 (6) & $-$0.12 (9) & +0.11 (12) & +0.08 (38) & +0.02 (57) & $-$0.01 (55) & $-$0.05 (60) & $-$0.47 (25) \\
 & split & +0.08 (6) & $-$0.12 (9) & +0.15 (12) & +0.10 (38) & +0.10 (57) & +0.07 (55) & +0.06 (60) & +0.09 (25) \\
DIOR vehicle & native & +0.58 (46) & +0.30 (43) & $-$0.08 (44) & $-$0.12 (18) & $-$0.62 (3) & $-$0.24 (5) & -- & -- \\
 & split & +0.73 (46) & +0.65 (43) & +0.60 (44) & +0.35 (18) & +0.08 (3) & $-$0.27 (5) & -- & -- \\
DIOR storage tank & native & +0.50 (8) & $-$0.18 (8) & +0.27 (4) & +0.18 (4) & -- & -- & -- & -- \\
 & split & +0.50 (8) & $-$0.18 (8) & +0.38 (4) & +0.62 (4) & -- & -- & -- & -- \\
\bottomrule
\end{tabular}}
\end{table}

\subsection{Detectors and fixed grids on FAIR1M and DIOR}
\label{supp:horizon:detectors}
A remote sensing system could also count by detecting the objects and counting the boxes. We ran three
open-vocabulary detectors on every crop of the horizon measurement: the second DOTA set (458 crops), FAIR1M (480) and
DIOR (445), and the first DOTA set (479) to choose their thresholds. The detectors are MM-Grounding-DINO-L~\citep{mmgdino},
whose released checkpoint was trained on Objects365, OpenImages and GoldG according to its model card, OWLv2-L~\citep{owlv2} and LLMDet-L~\citep{llmdet}. None of them was
trained on DOTA, FAIR1M or DIOR. Each crop is queried with the class phrase the VLM is asked about, in overlapping
tiles of 400 pixels (25\% overlap) enlarged to 800, the setting that did best for counting on our development
questions. Boxes are merged across tiles by non-maximum suppression and a containment test, and the score threshold of
each detector, from 0.10 to 0.50 in steps of 0.05, is chosen once on the first DOTA set by the median absolute
relative error. The VLM strategies are replayed as in Section~\ref{supp:model} on the answers for every part of
every crop, which we counted for FAIR1M and DIOR as for DOTA (3{,}877 views, 0 unparsed). Counting models trained on counting data, such as CountGD and
CountGD++~\citep{countgd,countgdpp}, are not part of this comparison.

Table~\ref{tab:detectors} gives the results. With the chosen thresholds (0.15 for MM-Grounding-DINO and OWLv2,
0.20 for LLMDet) every detector is less accurate than subdivision as GeoScope uses it on the second DOTA set and on
FAIR1M cars. By a paired bootstrap over images, the median error of subdivision is lower by 0.39 to 0.54 on DOTA and
0.34 to 0.55 on FAIR1M cars, and 17 to 23 and 24 to 40 points more of its counts lie within 20\% of the truth. The
detectors overcount views with few objects. At one to five objects on DOTA they answer 4.1 to 9.9 times the truth on
average. In views with more than 160 objects the two Grounding-DINO-based detectors miss about half or more of them on every set,
OWLv2 only on DIOR. Chosen on each set itself, the best threshold still leaves the best detector at 0.63 on DOTA and
0.33 on FAIR1M cars, against 0.33 and 0.24 for subdivision. On DIOR ships, in images of 800 pixels, OWLv2 is level with
subdivision in median error, 0.14 against 0.11 (paired difference $-0.02$ [$-0.06$, 0.01]), but its total error is more than twice as
large, 0.38 against 0.16 of the true count, because it misses most ships in the densest crops. Over all classes the
detectors also do worse in median error, OWLv2 with 0.25 on DIOR against 0.17 for subdivision and
MM-Grounding-DINO, the best detector there, with 0.70 on FAIR1M against 0.42.

\begin{figure}[!tp]
\centering
\includegraphics[width=\linewidth]{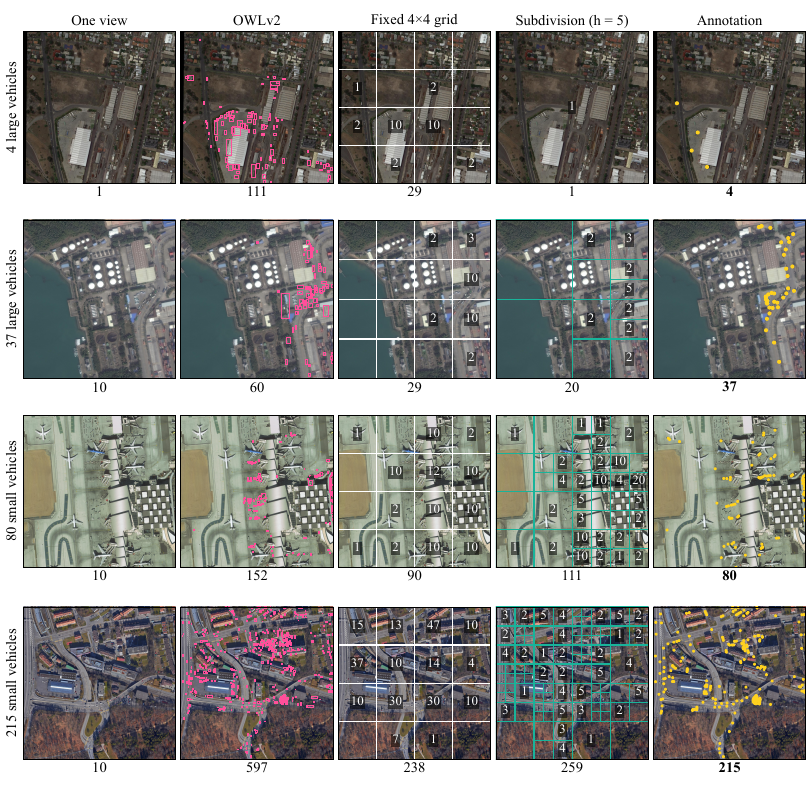}
\caption{Four DOTA crops counted by Qwen3-VL-8B in one view, by OWLv2, with a fixed 4$\times$4 grid and with
count-triggered subdivision, beside the annotation. Cell numbers: the VLM's count of that view; totals below each
crop.}
\label{fig:countgrid}
\end{figure}

Fig.~\ref{fig:countgrid} shows four crops of the second DOTA set, one per range of true counts, each the crop of at
least 1{,}024 pixels whose error under subdivision is closest to the median of its range. On the three denser crops
one view answers 10, while the grid and subdivision come within 11 to 46\% of the truth. The detector overcounts all
four crops, and the grid overcounts the sparse one sevenfold.

The FAIR1M and DIOR crops also repeat the fixed-grid comparison of Section~\ref{supp:model} outside DOTA. Their
images are small, so a 4$\times$4 grid usually stops at the 256-pixel cell after one split. On FAIR1M cars and DIOR
ships it costs 3.2 and 3.0 calls per crop, against 3.9 and 3.6 for subdivision, and the two are equally accurate,
0.24 and 0.11 in median error. Over all FAIR1M classes, which include the trucks whose annotations do not match the question, the
grid's median error is lower, 0.33 against 0.42.

\section{Implementation details of GeoScope}
\label{supp:method}
\subsection{Program grammar}
\label{supp:method:grammar}
The synthesiser writes one JSON object with two fields. The checker and the executor read only its field
\texttt{program}, and the field \texttt{answer\_type} is ignored. The checker accepts the programs of the grammar
below. Braces, colons and quoted keys are JSON syntax, and the members of an object may appear in any order. Square
brackets mark optional members and vertical bars separate alternatives.
\begin{promptbox}{Program grammar}
\begin{verbatim}
Program   ::= { "answer_type": Answer, "program": Top }
Answer    ::= "count" | "direction" | "position" | "color"
Top       ::= Count | Direction | Position | Color
Count     ::= { "op": "count", "child": Set }
Direction ::= { "op": "direction", "target": Set, "reference": Set }
Position  ::= { "op": "position", "target": Set }
Color     ::= { "op": "color", "target": Set }
Set       ::= Detect | Refer | InRegion | Filter | Near | Extreme | Top1
Detect    ::= { "op": "detect", "cls": Phrase }
Refer     ::= { "op": "refer", "expr": Text [, "cls": String] [, "plural": Bool]
                [, "within": (Part | Refer)] }
InRegion  ::= { "op": "in_region", "child": Set, "region": Region }
Filter    ::= { "op": "filter", "child": Set, "pred": Dir, "anchor": Set }
Near      ::= { "op": "near", "child": Set, "anchor": Set }
Extreme   ::= { "op": ("argmax" | "argmin"), "child": Set, "key": Key }
Top1      ::= { "op": "top1", "child": Set }
Region    ::= { "op": "image" } | { "op": "red_circle" } | Part | Around
Part      ::= { "op": "part", "part": Dir }
Around    ::= { "op": "around", "child": Set }
Dir       ::= "TOP" | "BOTTOM" | "LEFT" | "RIGHT" | "UL" | "UR" | "LL" | "LR" | "CENTER"
Key       ::= "X" | "Y" | "AREA" | "LENGTH"
Phrase    ::= a string of one to eight words
Text      ::= a string with at least one character other than a space
String    ::= any string
Bool      ::= true | false
\end{verbatim}
\end{promptbox}
Besides the structure, the checker requires one to eight words in the class phrase of every \texttt{detect}, a
non-empty expression in every \texttt{refer} and a nesting depth of at most eight below the top operator. It ignores
members that the grammar does not name. For UHR-Micro the checker also accepts the region
\verb|{"op": "box", "box": [x0, y0, x1, y1]}|, a pixel box taken from the question (\appref{supp:umicro}).

The two synthesis prompts offer different parts of the grammar (\appref{supp:method:prompts}). The counting prompt,
used for RC, XLRS-Bench overall counting and MME-RealWorld counting, offers \texttt{count} and \texttt{direction} with
every set and region operator except \texttt{refer}. The relation prompt, used for OSR, offers only \texttt{direction}
between two \texttt{refer} nodes. All 80 RC test programs count a class phrase inside the red circle.
Programs of type \texttt{position} and \texttt{color} come from a third prompt, written for the position and colour
questions of MME-RealWorld. GeoScope leaves these questions to the VLM (Section~\ref{sec:discussion:scope}). Questions that
state a pixel box are not synthesised (Section~\ref{sec:method:projection}).

\begin{table}[!tp]
\centering
\caption{Operators and their evaluation by the executor.}
\label{tab:supp:ops}
\footnotesize
\begin{tabular}{p{0.24\textwidth}p{0.69\textwidth}}
\toprule
Operator & Evaluation \\
\midrule
\raggedright \texttt{detect}$(c)$ & \raggedright boxes of $c$ grounded in the current window, coarse to fine (\appref{supp:method:ladder}) \tabularnewline
\raggedright \texttt{refer}$(e,c,\mathit{pl},w)$ & \raggedright $e$ grounded as a whole, in the whole image or, if $w$ is a place, in a window
around it that is 15\% larger than the place on each side and at least 1{,}024 pixels wide; an image part $w$ is
appended to $e$ as a phrase. Each box is grounded again with $c$ in a close-up of at least 1{,}024 pixels and three
times its size. One box is kept, or up to four if $\mathit{pl}$ is true \tabularnewline
\raggedright \texttt{in\_region}$(s,R)$ & \raggedright members of $s$ whose centre lies in $R$; $s$ is grounded in the box of $R$ enlarged by
15\% on each side \tabularnewline
\raggedright \texttt{filter}$(s,d,a)$ & \raggedright members of $s$ within $\rho$ of the centre of $a$ whose direction from it lies within
$67.5^\circ$ of $d$, or within $\rho/2$ for \texttt{CENTER} \tabularnewline
\raggedright \texttt{near}$(s,a)$ & \raggedright members of $s$ within $\rho$ of the centre of $a$ \tabularnewline
\raggedright \texttt{top1}$(s)$ & \raggedright the first member of $s$ in the order of the VLM's answer \tabularnewline
\raggedright \texttt{argmax}, \texttt{argmin}$(s,k)$ & \raggedright the member with the largest or smallest centre $x$, centre $y$, area or
longer side, for \texttt{X}, \texttt{Y}, \texttt{AREA} and \texttt{LENGTH} \tabularnewline
\raggedright \texttt{image} & \raggedright the whole image \tabularnewline
\raggedright \texttt{red\_circle} & \raggedright the ellipse fitted to the red ring drawn on the image \tabularnewline
\raggedright \texttt{part}$(d)$ & \raggedright the upper, lower, left or right half for \texttt{TOP}, \texttt{BOTTOM}, \texttt{LEFT} and
\texttt{RIGHT}, a quadrant for \texttt{UL}, \texttt{UR}, \texttt{LL} and \texttt{LR}, and the middle half of both
axes for \texttt{CENTER} \tabularnewline
\raggedright \texttt{around}$(s)$ & \raggedright the box around all members of $s$, enlarged by 5\% on each side \tabularnewline
\raggedright \texttt{count}$(s)$ & \raggedright the VLM's count in the window of $s$ by Algorithm~\ref{alg:horizon}, or the number of members
of $s$ \tabularnewline
\raggedright \texttt{direction}$(t,r)$ & \raggedright the direction-relation matrix~\citep{cdc} of the boxes around $t$ and $r$, or the vector
between their centres if $t$ lies inside $r$. If $r$ is a single box, target candidates that overlap it with an IoU
above 0.5 are discarded unless all do \tabularnewline
\bottomrule
\end{tabular}
\par\smallskip
\begin{minipage}{\textwidth}\footnotesize
$W\times H$: image size; $\rho=\max\big(4\,\mathrm{diag}(a),\,0.05\min(W,H)\big)$, with $\mathrm{diag}(a)$ the
diagonal of the box around the anchor $a$. The centre of a group is the mean of the centres of its members.
\end{minipage}
\end{table}

Table~\ref{tab:supp:ops} lists how the executor evaluates each operator. A \texttt{count} is evaluated in one of two
ways. If its set is a \texttt{detect} of a class $c$, an \texttt{in\_region} of such a set within a region $R$, or a
\texttt{filter} or \texttt{near} of one of these two, the VLM counts $c$ in a window by Algorithm~\ref{alg:horizon}. The window is the
whole image, the bounding box of $R$, or a square of side $2\rho$ at the anchor, cut to the box of $R$ if there is
one. The square is centred on the anchor for \texttt{near} and for the direction \texttt{CENTER}, and otherwise its
centre lies $\rho$ away from the centre of the anchor in direction $d$. A window other than the whole image is
enlarged by 15\% of its size on each side if no side exceeds the cell size, 7{,}000 pixels for MME-RealWorld and
2{,}000 for XLRS-Bench overall counting. A larger window, the whole image included, is cut into equal cells of at most
that size, which are counted separately and summed. For RC the
window is the bounding box of the drawn circle enlarged by 15\%, counted in one view at 1\,M pixels with a question
about the objects inside the red circle. Any other set is grounded, and the count is the number of its boxes
(\appref{supp:method:ladder}).

The mapping from question phrases to operators is not hard-coded. The synthesiser follows the rules and worked examples of
the prompts, summarised in Table~\ref{tab:supp:phrases}. Before the check, direction words that are not among the nine
tokens are mapped onto them. \texttt{TL}, \texttt{TR}, \texttt{BL} and \texttt{BR}, and corner names such as
\texttt{UPPER\_LEFT} or \texttt{BOTTOM LEFT}, become the corner tokens. \texttt{UPPER}, \texttt{ABOVE}, \texttt{UP}
and \texttt{NORTH} become \texttt{TOP}; \texttt{LOWER}, \texttt{BELOW}, \texttt{DOWN} and \texttt{SOUTH} become
\texttt{BOTTOM}; \texttt{WEST} and \texttt{EAST} become \texttt{LEFT} and \texttt{RIGHT}; and \texttt{MIDDLE},
\texttt{CENTRE} and \texttt{MID} become \texttt{CENTER}.

\begin{table}[!tp]
\centering
\caption{Question phrases and the program parts the two prompts map them to, with examples from the prompts.}
\label{tab:supp:phrases}
\footnotesize
\begin{tabular}{p{0.27\textwidth}p{0.33\textwidth}p{0.30\textwidth}}
\toprule
Question phrase & Program & Example \\
\midrule
\raggedright object noun with its colour and shape words & \raggedright class phrase \texttt{cls}, singular, without positions, counts or
relations & \raggedright \emph{white storage tank} \tabularnewline
\raggedright \emph{in the entire image} & \raggedright \texttt{detect} as the counted set & \raggedright \emph{How many airplanes are there in the entire
image?} \tabularnewline
\raggedright \emph{inside} or \emph{within the red circle} & \raggedright \texttt{in\_region} with \texttt{red\_circle} & \raggedright \emph{How many wind
turbines are there inside the red circle?} \tabularnewline
\raggedright position in the image frame & \raggedright \texttt{in\_region} with \texttt{part}; relation prompt: \texttt{within} a
\texttt{part} & \raggedright \emph{in the upper left corner of the picture}, \texttt{UL} \tabularnewline
\raggedright position relative to another object & \raggedright \texttt{filter} with that object as anchor & \raggedright \emph{below the long white
building}, \texttt{BOTTOM} \tabularnewline
\raggedright \emph{near} another object & \raggedright \texttt{near}; relation prompt: part of \texttt{expr} & \raggedright \emph{the yellow bus near the
gas station} \tabularnewline
\raggedright \emph{in}, \emph{on} or \emph{along} a place & \raggedright \texttt{in\_region} with \texttt{around} the place; relation prompt:
\texttt{within} a \texttt{refer} & \raggedright \emph{in the parking lot to the right of the gray stadium} \tabularnewline
\raggedright singular referent & \raggedright \texttt{top1}, unless a superlative makes it single; relation prompt: \texttt{plural} false &
\emph{the gray stadium} \tabularnewline
\raggedright plural referent & \raggedright a set located at its centre; relation prompt: \texttt{plural} true & \raggedright \emph{the two baseball
fields} \tabularnewline
\raggedright \emph{leftmost}, \emph{rightmost}, \emph{topmost}, \emph{bottommost} & \raggedright \texttt{argmin} or \texttt{argmax} of
\texttt{X} or \texttt{Y} & \raggedright -- \tabularnewline
\raggedright \emph{largest}, \emph{biggest}, \emph{smallest}, \emph{longest} & \raggedright \texttt{argmax} or \texttt{argmin} of
\texttt{AREA}, \texttt{argmax} of \texttt{LENGTH}; relation prompt: part of \texttt{expr} & \raggedright \emph{the largest orange
warehouse} \tabularnewline
\raggedright where A is located relative to B & \raggedright \texttt{direction} with target A and reference B & \raggedright \emph{where is the round white
storage tank located in relation to the rectangular pier} \tabularnewline
\raggedright scene phrase that applies to both objects & \raggedright the same constraint on both referents; relation prompt: the same
\texttt{within} & \raggedright \emph{On the runway in the middle of the picture,} \ldots \tabularnewline
\bottomrule
\end{tabular}
\end{table}
\FloatBarrier
\subsection{Prompts and the coverage check}
\label{supp:method:prompts}
Each prompt is sent to Qwen3-4B as one user turn, with the question in place of \texttt{<{}<QUESTION>{}>}. The model
decodes greedily with thinking disabled and at most 512 new tokens, and the first complete JSON object of its reply is
parsed. Neither prompt contains the answer options. We reproduce both prompts word for word. Line breaks were added to
fit the page, so a line that starts with spaces continues the line above, and runs of spaces that align columns are
shortened.

The counting prompt holds ten worked examples, six about counting and four about directions, and we print the second,
fourth and fifth. The first, third and sixth ask for the airplanes in the entire image, the patches of farmland within
a red-circled area and the green cars below a long white building in the bottom left area. The last four ask the first
four questions of the relation prompt, written with the operators of the counting prompt.
\begin{promptbox}{Counting prompt of the synthesiser}
\begin{verbatim}
You translate a multiple-choice question about a large remote-sensing image into a STRICT
  JSON program (no markdown, no commentary, no <think>). You see only the question text,
  never the image and never the answer options. Output exactly one JSON object:
{"answer_type": "count" | "direction", "program": <top>}

<top> is one of:
- {"op": "count", "child": <set>}   how many objects are in <set>
- {"op": "direction", "target": <set>, "reference": <set>}   where <target> is located
  relative to <reference>

<set> (a scored set of detected objects) is one of:
- {"op": "detect", "cls": "<noun phrase>"}   all objects matching the phrase
- {"op": "in_region", "child": <set>, "region": <region>}   objects whose centre lies
  inside the region
- {"op": "filter", "child": <set>, "pred": "<DIR>", "anchor": <set>} objects lying in
  direction DIR of the anchor object
- {"op": "near", "child": <set>, "anchor": <set>}   objects close to the anchor object
- {"op": "argmax", "child": <set>, "key": "<KEY>"} / {"op": "argmin", "child": <set>,
  "key": "<KEY>"}   the single most/least extreme object
- {"op": "top1", "child": <set>}   the single best-matching object (use for a singular
  "the X")

<region> is one of:
- {"op": "image"}   the whole image
- {"op": "red_circle"}   the area inside a red circle drawn on the image
- {"op": "part", "part": "<DIR>"}   a part of the image frame
- {"op": "around", "child": <set>}   the area covered by some objects (e.g. inside a
  parking lot, on a field)

DIR: TOP, BOTTOM, LEFT, RIGHT, UL, UR, LL, LR, CENTER   (UL = upper/top left, UR =
  upper/top right, LL = lower/bottom left, LR = lower/bottom right)
KEY: X (left->right), Y (top->bottom), AREA (size), LENGTH (longest side)
Superlatives: leftmost = argmin X, rightmost = argmax X, topmost = argmin Y, bottommost =
  argmax Y, largest/biggest = argmax AREA, smallest = argmin AREA, longest = argmax
  LENGTH.

RULES:
- "cls" is a short detector phrase: the object noun plus its colour/shape words ("red
  car", "L-shaped black roof building", "white storage tank"). Use the singular noun.
  Never put positions, counts or relations inside "cls".
- Position words about the IMAGE frame ("in the top left corner of the picture", "on the
  right side of the image", "in the middle") become in_region with a part region. A
  position relative to ANOTHER object ("below the white building", "to the left of the
  bridge") becomes filter with that object as anchor.
- "inside / within the red circle (red-circled area)" becomes in_region with {"op":
  "red_circle"}.
- "in / on / along <a place such as a parking lot, field, road, river, harbor>" becomes
  in_region with an "around" region of that place.
- A singular referent ("the blue building", "the bridge") is wrapped in top1 unless a
  superlative already makes it single. A plural referent ("the three sports fields", "the
  cluster of houses") is NOT wrapped: its position is the centre of the group.
- For "where is A relative to / in relation to B", "what is the
  direction/position/orientation of A relative to B": target = A, reference = B.
- Keep context phrases that locate the scene ("In the parking lot under the black roof
  building ...") as in_region / filter constraints on BOTH target and reference when they
  apply to both.
- For whole-image counts ("in the entire image / picture") the set is just detect.

EXAMPLES:
Q: How many wind turbines are there inside the red circle?
{"answer_type": "count", "program": {"op": "count", "child": {"op": "in_region", "child":
  {"op": "detect", "cls": "wind turbine"}, "region": {"op": "red_circle"}}}}
Q: How many white trucks are there in the upper left corner of the picture?
{"answer_type": "count", "program": {"op": "count", "child": {"op": "in_region", "child":
  {"op": "detect", "cls": "white truck"}, "region": {"op": "part", "part": "UL"}}}}
Q: How many buses are parked in the parking lot to the right of the gray stadium?
{"answer_type": "count", "program": {"op": "count", "child": {"op": "in_region", "child":
  {"op": "detect", "cls": "bus"}, "region": {"op": "around", "child": {"op": "filter",
  "child": {"op": "detect", "cls": "parking lot"}, "pred": "RIGHT", "anchor": {"op":
  "top1", "child": {"op": "detect", "cls": "gray stadium"}}}}}}}

Q: <<QUESTION>>
\end{verbatim}
\end{promptbox}
The relation prompt holds six worked examples, and we print the third and the fourth. The first two ask for the round
white storage tank relative to the rectangular pier and for the largest orange warehouse relative to the two baseball
fields. The fifth asks for the small blue greenhouse in the lower right corner relative to the cluster of factory
buildings. The sixth asks for the white yacht in the marina on the left side of the image relative to the long wooden
jetty.
\begin{promptbox}{Relation prompt of the synthesiser}
\begin{verbatim}
You translate a question about the relative position of two objects in a large
  remote-sensing image into a STRICT JSON program (no markdown, no commentary, no
  <think>). You see only the question text, never the image and never the answer options.
  Output exactly one JSON object:
{"answer_type": "direction", "program": {"op": "direction", "target": <ref>, "reference":
  <ref>}}

The question asks where the TARGET is located relative to the REFERENCE ("where is A
  relative to / in relation to B", "what is the position/direction/orientation of A
  relative to B", "A is located in which direction of B": target = A, reference = B).

<ref> = {"op": "refer", "expr": "<referring expression>", "cls": "<head phrase>",
  "plural": true | false, "within": <container>}
- "within" (optional): the larger area that CONTAINS the object, when the question gives
  one. It is either an image-frame part {"op": "part", "part": "<DIR>"} ("in the top left
  corner of the picture", "on the right side of the image") or a place {"op": "refer",
  "expr": "<place expression with all of its own locating words>", "cls": "<place head
  phrase>", "plural": false} ("in the parking lot beneath the white building", "on the
  runway in the middle of the picture", "in the residential area on the left"). If a scene
  phrase applies to both objects ("In the parking lot under the black roof building, ...",
  "In the city on the right side of the image, ..."), give BOTH refs the same "within".
  Omit "within" when the question gives no containing area.
- "expr": the object's own description, copied from the question words: its category,
  colour, shape and size words ("largest", "longer") and its relations to nearby objects
  ("near the river", "below the yellow buildings"), but WITHOUT the containing area
  already given in "within". Never put the other object's relation to this one into expr
  ("relative to B" is not part of A's expr).
- "cls": the short category phrase with colour/shape words only, singular ("white car",
  "L-shaped black roof building", "tennis court").
- "plural": true only when the expression denotes a group ("the two adjacent tennis
  courts", "the cluster of houses", "the three sports fields"); otherwise false.

DIR: TOP, BOTTOM, LEFT, RIGHT, UL, UR, LL, LR, CENTER   (UL = upper/top left, UR =
  upper/top right, LL = lower/bottom left, LR = lower/bottom right)

EXAMPLES:
Q: What is the position of the red bus at the top of the picture relative to the yellow
  bus near the gas station?
{"answer_type": "direction", "program": {"op": "direction", "target": {"op": "refer",
  "expr": "the red bus", "cls": "red bus", "plural": false, "within": {"op": "part",
  "part": "TOP"}}, "reference": {"op": "refer", "expr": "the yellow bus near the gas
  station", "cls": "yellow bus", "plural": false}}}
Q: On the runway in the middle of the picture, in which direction is the white airplane
  located relative to the gray airplane?
{"answer_type": "direction", "program": {"op": "direction", "target": {"op": "refer",
  "expr": "the white airplane", "cls": "white airplane", "plural": false, "within": {"op":
  "refer", "expr": "the runway in the middle of the picture", "cls": "runway", "plural":
  false}}, "reference": {"op": "refer", "expr": "the gray airplane", "cls": "gray
  airplane", "plural": false, "within": {"op": "refer", "expr": "the runway in the middle
  of the picture", "cls": "runway", "plural": false}}}}

Q: <<QUESTION>>
\end{verbatim}
\end{promptbox}
The coverage check compares a counting program with the words of its question and never reads the options. It looks
for the words and phrases of Table~\ref{tab:supp:coverage}, ignoring case. In the reported configuration it runs on
MME-RealWorld counting only. A program that misses any of these words, or that the checker rejects, is synthesised
once more, with one of two notes appended to the question on a line of its own:
\begin{promptbox}{Notes appended for a second synthesis}
\begin{verbatim}
NOTE: a previous program for this question ignored these constraints: "<word>", "<word>".
  Every colour/shape word and every place or position phrase in the question must be
  represented in the program: inside a "cls" phrase, as an image part, or as an anchor /
  place set.

NOTE: a previous program for this question did not follow the grammar above; output one
  valid JSON object that follows it exactly.
\end{verbatim}
\end{promptbox}
The first note lists the words that the first program missed. The second program replaces the first if it is well
typed and either the first was ill typed or it misses fewer words. RC and OSR programs are executed as first written, and an
ill-typed one is answered zero-shot.

\begin{table}[!tp]
\centering
\caption{Words and phrases of the question that the coverage check looks for, ignoring case, and when a program
covers them. Corner positions map to \texttt{UL}, \texttt{UR}, \texttt{LL} and \texttt{LR}.}
\label{tab:supp:coverage}
\footnotesize
\begin{tabular}{p{0.11\textwidth}p{0.53\textwidth}p{0.26\textwidth}}
\toprule
Kind & Words or patterns & Covered if \\
\midrule
\raggedright Colour & \raggedright red, blue, white, black, gray, grey, green, yellow, orange, brown, purple, pink, silver, golden, dark,
light, beige, cyan & \raggedright the word occurs in a class phrase or referring expression; gray and grey count as one word \tabularnewline
\raggedright Shape and size & \raggedright square, rectangular, circular, round, triangular, L-shaped, T-shaped, U-shaped, oval, long, large,
small, big, tall, irregular, curved, straight & \raggedright the word occurs in a class phrase or referring expression \tabularnewline
\raggedright Place & \raggedright parking lot, parking space, road, highway, street, river, bridge, lake, pond, harbor, harbour, port, dock,
pier, field, park, square, runway, stadium, court, playground, forest, grassland, greenland, grass, lawn, water,
shore, beach, yard, roundabout, intersection, wall, roof, building, factory, residential area, town, village,
farmland; also in the plural & \raggedright the noun occurs within a class phrase or referring expression \tabularnewline
\raggedright Position & \raggedright upper or top left; upper or top right; lower or bottom left; lower or bottom right; \emph{left} or
\emph{right} before side, edge, part, area or half, or after \emph{on the}, for \texttt{LEFT} and \texttt{RIGHT};
\emph{upper} or \emph{top} before side, edge, part, area, half, \emph{of the picture} or \emph{of the image}, or
\emph{at the top}, for \texttt{TOP}, and likewise \emph{lower} or \emph{bottom} for \texttt{BOTTOM}; center, centre,
middle or central, for \texttt{CENTER} & \raggedright the token is a part or predicate of the program, both tokens of a corner
are, or the matched words occur in a referring expression \tabularnewline
\bottomrule
\end{tabular}
\end{table}

The executor asks the VLM four kinds of question. Each view is a crop of the original image, reduced to the pixel
budget $B$ if it is larger. A small crop is enlarged, by at most four times, until its longer side has 768 pixels.
Both sides are rounded to multiples of 32 pixels, and the model receives exactly this image. Qwen3-VL decodes
greedily, without a system turn. The count question replaces \texttt{<where>} by \emph{inside the red circle} for RC
and by \emph{in this image} otherwise. Its reply, at most 12 tokens, is read as the first integer it contains, or else
as the smallest number word from zero to twelve that it contains. The class phrase is inserted as the program writes
it, usually in the singular.
\begin{promptbox}{Count question to the VLM}
\begin{verbatim}
This is a crop of a high-resolution aerial image. How many <phrase> are there <where>?
  Count each one carefully. Answer with a single integer only.
\end{verbatim}
\end{promptbox}
The two grounding questions ask for boxes, the first for a class phrase and the second for a referring expression,
with the boxes labelled by its class phrase. A plural expression is asked for as \emph{Locate all of} \ldots, with \emph{bounding
boxes}. Each reply is cut at 1{,}024 tokens, about 30 boxes. The boxes are read in the 0 to 1{,}000 frame of the view,
and their order in the reply serves as their score. The fourth kind is the multiple-choice question itself, asked with
the views of a pixel box (\appref{supp:results:xlrs}) and answered by the first-token scores of the option letters.
\begin{promptbox}{Grounding questions to the VLM}
\begin{verbatim}
Locate every <phrase> in the image. Output the bounding boxes as a JSON list of
  {"bbox_2d": [x1, y1, x2, y2], "label": "<phrase>"}. If there is none, output [].

Locate <expr> in the image. Output the bounding box as a JSON list of
  {"bbox_2d": [x1, y1, x2, y2], "label": "<cls>"}. If there is none, output [].
\end{verbatim}
\end{promptbox}
\subsection{Fallback steps and RC re-counting}
\label{supp:method:ladder}
Execution degrades in steps, following the fallback design of GeoSelect~\citep{geoselect}. Table~\ref{tab:supp:ladder} lists every step of
the reported configuration with its trigger and the number of test questions in which it fired.

\begin{table}[!tp]
\centering
\caption{Fallback steps on the test splits: questions in which each step fired.}
\label{tab:supp:ladder}
\footnotesize
\begin{tabular}{lp{0.43\textwidth}rrr}
\toprule
Step & Trigger & RC & OSR & Count \\
\midrule
\multicolumn{5}{l}{\emph{Synthesis}} \\
Repair & \raggedright a known slip makes the program ill typed & 0 & 3 & 92 \\
Second synthesis & \raggedright the program is ill typed after repair & -- & -- & 13 \\
 & \raggedright the program misses a word of the question & -- & -- & 458 \\
\midrule
\multicolumn{5}{l}{\emph{Execution}} \\
Red circle & \raggedright no red ring is found & 3 & -- & -- \\
Class phrase & \raggedright the referring expression yields no box & -- & 108 & -- \\
Whole image & \raggedright the place, or the referent inside it, is not found & -- & 10 & -- \\
Relaxation & \raggedright an anchor or place is not found & -- & -- & 54 \\
Counting boxes & \raggedright the counted set has no region form & -- & -- & 51 \\
Re-counting & \raggedright the count equals no option value & 24 & -- & -- \\
\midrule
\multicolumn{5}{l}{\emph{Answer}} \\
Executed program & \raggedright & 77 & 382 & 962 \\
Zero-shot answer & \raggedright GeoScope abstains & 3 & 18 & 19 \\
\quad of which correct & \raggedright & 1 & 3 & 8 \\
\bottomrule
\end{tabular}
\par\smallskip
\begin{minipage}{\textwidth}\footnotesize
Out of 80 RC, 400 OSR and 981 MME-RealWorld counting questions. --: not part of the task's configuration.
\end{minipage}
\end{table}

\paragraph{Synthesis}
A repair step rewrites common slips of the synthesiser, by fixed rules that keep the meaning, before the program is
checked. Direction words are mapped onto the nine tokens (\appref{supp:method:grammar}), and a \texttt{filter}
anchored on the image itself becomes \texttt{in\_region} with the matching part. A \texttt{filter} whose predicate
names proximity, such as \texttt{NEAR}, \texttt{NEXT\_TO}, \texttt{BESIDE} or \texttt{ALONG}, becomes \texttt{near},
and one whose predicate is no direction at all is removed. An \texttt{argmax} or \texttt{argmin} with an unknown key
becomes \texttt{top1}, and a set given as a region becomes \texttt{around} that set.
On the test splits, repair changed 3 OSR and 92 counting programs, each of which the checker would otherwise have
rejected. In 43 of the counting programs it turned a proximity predicate into \texttt{near}.
For MME-RealWorld counting the checker then rejected 13 programs and the coverage check flagged 458, and each of these
questions was synthesised once more. The second program replaced 4 of the rejected programs and 93 of the flagged
ones, 58 of which then missed no word. Of the other flagged programs, 350 were kept because the second program missed
as many words or more, and 15 because it was ill typed. The missing words were place nouns in 257 of the 458
questions, positions in 255, colour words in 49 and shape words in 40.

\paragraph{Grounding}
A class phrase is grounded coarse to fine when the pixel budget shrinks the sides of its window to 60\% of their
native length or less. The first four boxes of the reduced view are grounded again in close-ups of at least 1{,}024
pixels and three times their size. A box that its close-up does not confirm is dropped, unless every box would be, and
if the reduced view yields no box at all, the window is searched in tiles of about 3{,}000 pixels.
A referring expression that yields no box is replaced by its class phrase, which happened for 122 of the 793 OSR
referents grounded on the test split, in 108 questions. If a place container is not found, or the referent is not
found in the window around it, the referent is searched for in the whole image with the place phrase appended to its
expression. On OSR this happened in 10 test questions. In the OSR configuration an image-part container is always
handled this way, because image parts are visible in the whole-image view.

\paragraph{Relaxation}
If an anchor or place of a counting program is not found, it is searched for again without its colour and shape words,
and then also without its own spatial constraints. Relaxation was tried for 55 anchors and places in 54 counting questions.
Ten were found without their colour and shape words, 41 more without their constraints as well, and 4 not at all, so
that 50 of the questions could be executed. In 3 more questions an anchor or place was not found and had nothing to
drop. On the development split, removing the coverage check and relaxation together lowered counting accuracy from 43.3
to 42.4 and raised the share of abstentions from 2.4 to 6.1\%.

\paragraph{Counting}
When the set of a counting program is a \texttt{detect}, optionally inside an \texttt{in\_region} and that
optionally inside a \texttt{filter} or \texttt{near}, the executor turns it into a class phrase and one window, which
Algorithm~\ref{alg:horizon} counts. Of the 962 counting programs executed on the test split, 911 had such a set, and for 474
of these at least one view was split. The other 51, of which 43 nest a \texttt{filter} or \texttt{near} inside an
\texttt{in\_region}, are grounded, and the count is the number of boxes. Coarse-to-fine grounding keeps at most four
boxes of a reduced view. The 51 counts lie between zero and four, and 10 of their answers are correct. An unparseable
count in a sub-window leaves the count of its parent view in place, and an unparseable count of a whole window abstains,
which did not happen on the test split.

\paragraph{RC}
The drawn circle is found in a copy of the image reduced to 2{,}500 pixels on its longer side. Pixels with a red value
above 150, green and blue values below 90 and red exceeding both by more than 90 are kept. An ellipse is fitted to the
largest connected group of them, preferring groups that fill less than 60\% of their bounding box, as a ring does. The
circle was not found in 3 of the 80 RC test images, and GeoScope abstained on these questions.
A count that equals no option value is taken twice more. The first extra view shows the bounding box of the circle
enlarged by 30\% on each side, and the second shows the original window enlarged, by at most four times, until its longer
side has 1{,}536 pixels. The median of the three counts is used. Re-counting uses the values of the options, never
their order or letters, and the VLM never sees them. It fired for 24 of the 77 executed RC questions. It changed the
count in two of them and the chosen option in one, which was wrong before and after, so RC accuracy is the same
without it.

\paragraph{Abstention}
When every step fails, GeoScope abstains and returns the answer of the zero-shot VLM at 4\,M pixels
(Section~\ref{sec:exp:baselines}). On OSR it abstained on 14 questions because a referent was not found, 3 because both
referents were grounded on the same box, which leaves no direction, and 1 because its program was ill typed. On
MME-RealWorld counting, 9 programs remained ill typed, 7 anchors or places were not found, and 3 regions were empty.
Two of them were relative regions less than 8 pixels wide inside the region containing them, and one did not overlap
the region containing it. The zero-shot answers are correct for 1 of the 3 RC, 3 of the 18 OSR and 8 of the 19
counting questions on which it abstained. With LLaVA-OV-7B as the perception model, GeoScope abstains on 182 of
the 400 OSR test questions, mostly because one of the two referents is not found.
\subsection{Program correctness and coverage}
\label{supp:method:correctness}
We wrote reference programs by hand for 150 questions, 30 test questions per task and 30 development questions each
for OSR and MME-RealWorld counting. For each we judged whether the synthesised program expresses every constraint of
the question, without looking at the answer. All 30 RC programs are acceptable, as are 34 of the 60 OSR programs and
27 of the 60 counting programs. Most faults are simplifications. Of the 33 faulty counting programs, 16 drop the place
and keep the image part, so that the parking lot in the lower right becomes the lower right, and 10 write a relation
the question does not state. OSR programs most often attach a scene condition to only one of the two objects, in 9
cases, or add a direction the question does not name, in 7. Questions with acceptable and unacceptable
programs are answered correctly at similar rates, 41\% and 42\% for counting and 44\% and 38\% for OSR. Replacing the synthesised programs by the hand-written ones on the test questions changes accuracy by at most two
of thirty per task, from 19, 11 and 11 correct answers to 19, 13 and 11 for RC, OSR and counting.

Of the 245 development counting questions, 43 use constructions the grammar lacks, mostly words for the state or
action of an object, ordinals and \emph{between}. Their simplified programs are answered correctly in 51.2\% of the
cases, against 41.6\% for the other questions.
\subsection{Prompt sensitivity and choice of synthesiser}
\label{supp:method:synth}
\paragraph{Prompt variants}
We changed each synthesis prompt in two ways and ran the development split with the rest of the configuration
unchanged. The first variant removes the guidance, which is the RULES block of the counting prompt and the four lines
that explain the fields of a referring expression in the relation prompt. The second keeps every other worked example,
five of ten and three of six.
Table~\ref{tab:supp:prompts} gives the results. RC programs did not change at all. For OSR and counting, 33 to
61\% of the programs stayed identical, and two of the four variants were significantly less accurate: OSR with every
other example by 10.0 points and counting without the guidance by 6.1. The guidance of the counting prompt and the
examples of the relation prompt are thus needed, and the programs depend on the wording of the prompt.

\begin{table}[!tp]
\centering
\caption{Prompt variants on the development split.}
\label{tab:supp:prompts}
\footnotesize
\setlength{\tabcolsep}{4pt}
\resizebox{\textwidth}{!}{%
\begin{tabular}{llccccc}
\toprule
Task & Prompt & \begin{tabular}[c]{@{}c@{}}Well\\typed\end{tabular} & Identical &
\begin{tabular}[c]{@{}c@{}}Same\\structure\end{tabular} & Accuracy & $\Delta$ \\
\midrule
RC & final & 100.0 & -- & -- & 60.0 & -- \\
 & no guidance & 100.0 & 100.0 & 100.0 & 60.0 & \dci{0.0}{0.0}{0.0} \\
 & every other example & 100.0 & 100.0 & 100.0 & 60.0 & \dci{0.0}{0.0}{0.0} \\
\midrule
OSR & final & 100.0 & -- & -- & 56.0 & -- \\
 & no guidance & 100.0 & 33.0 & 57.0 & 50.0 & \dci{-6.0}{-14.0}{+2.0} \\
 & every other example & 95.0 & 40.0 & 56.0 & 46.0 & \dci{-10.0}{-16.0}{-5.0} \\
\midrule
MME-RS count & final & 99.2 & -- & -- & 43.3 & -- \\
 & no guidance & 97.1 & 60.8 & 62.4 & 37.1 & \dci{-6.1}{-10.2}{-2.0} \\
 & every other example & 98.8 & 55.5 & 57.6 & 41.6 & \dci{-1.6}{-6.1}{+2.9} \\
\bottomrule
\end{tabular}}
\par\smallskip
\begin{minipage}{\textwidth}\footnotesize
Identical and same structure: share of programs equal to those of the final prompt as JSON, or in their operators and
direction tokens. $\Delta$: accuracy minus that of the final prompt, paired on the same questions, with its 95\%
bootstrap interval underneath.
\end{minipage}
\end{table}

\paragraph{Choice of synthesiser}
Table~\ref{tab:supp:synth} replaces Qwen3-4B by five other synthesisers and keeps everything else, the prompts
included. Two are vision--language models, Qwen3-VL-4B and the perception model Qwen3-VL-8B, each given the question
alone or the question and the image at 4\,M pixels. The other three are text models from other families, the dense
transformers Phi-4-mini~\citep{phi4mini} and Llama-3.2-3B, a smaller model of the Llama 3 family~\citep{llama3}, and the state-space model
Falcon3-Mamba-7B, a continued pre-training of Falcon Mamba 7B~\citep{falconmamba}. An ill-typed program goes through the same fallback steps as in GeoScope (\appref{supp:method:ladder}).
No synthesiser does better than Qwen3-4B. The vision--language models lose 4 to 7 points on OSR, none significantly on
100 questions, and 3.7 to 5.7 points on counting. The loss on counting is significant for Qwen3-VL-8B, with
differences of $-5.7$ [$-10.6$, $-0.8$] points without the image and $-5.3$ [$-10.2$, $-0.8$] with it. Showing the
image changes accuracy by $+0.4$ to $+3.0$ points, and none of the four paired intervals excludes zero, the largest
gain being $+3.0$ [$-1.0$, $+8.0$] on OSR for Qwen3-VL-8B.
The image raises the share of well-typed programs of Qwen3-VL-4B from 79 to 89\% on OSR and from 86 to 93\% on
counting. Without the image, 74 of the 245 counting programs of Qwen3-VL-8B are ill typed. Phi-4-mini writes
well-typed programs for 98, 91 and 100\% of the OSR, counting and RC questions, Llama-3.2-3B for 57, 56 and 20\%, and
Falcon3-Mamba-7B for 77, 94 and 5\%. On RC, whose programs are simple, every synthesiser reaches the 60.0 of Qwen3-4B
except these two, which fall to 45.0 and 40.0.
With Qwen3-VL-4B as the perception model, letting it also write the programs lowers OSR from 50.0 to 40.0 and counting
from 35.5 to 33.5, and leaves RC at 45.0. Among Qwen3 text models of 0.6B, 1.7B, 4B and 8B parameters, the 4B model
writes the best programs, with 47, 49, 56 and 46 on OSR and 38.4, 41.6, 43.3 and 40.8 on counting. The prompt variants
and the five synthesisers were evaluated after the configuration had been frozen, and they did not change it.

\begin{table}[!tp]
\centering
\caption{Development accuracy (\%) of GeoScope with other synthesisers, on 100 OSR, 245 MME-RealWorld counting and 20 RC questions.}
\label{tab:supp:synth}
\footnotesize
\begin{tabular}{llccc}
\toprule
Synthesiser & Input & OSR & MME-RS count & RC \\
\midrule
\multicolumn{5}{l}{\emph{Perception by Qwen3-VL-8B}} \\
Qwen3-4B & question & 56.0 & 43.3 & 60.0 \\
Qwen3-VL-4B & question & 49.0 & 38.8 & 60.0 \\
Qwen3-VL-4B & question and image & 51.0 & 39.6 & 60.0 \\
Qwen3-VL-8B & question & 49.0 & 37.6 & 60.0 \\
Qwen3-VL-8B & question and image & 52.0 & 38.0 & 60.0 \\
Phi-4-mini & question & 48.0 & 36.3 & 60.0 \\
Llama-3.2-3B & question & 36.0 & 38.8 & 45.0 \\
Falcon3-Mamba-7B & question & 40.0 & 40.8 & 40.0 \\
\midrule
\multicolumn{5}{l}{\emph{Perception by Qwen3-VL-4B}} \\
Qwen3-4B & question & 50.0 & 35.5 & 45.0 \\
Qwen3-VL-4B & question & 40.0 & 33.5 & 45.0 \\
\bottomrule
\end{tabular}
\par\smallskip
\begin{minipage}{\textwidth}\footnotesize
The first row of each block is GeoScope's synthesiser. All synthesisers use the same prompts, and a synthesiser that receives the image gets it at 4\,M pixels.
\end{minipage}
\end{table}

\section{Evaluation protocol}
\label{supp:protocol}
\subsection{Splits}
\label{supp:protocol:splits}
Each task is split once by a fixed rule, and the split is stored with its questions. The question identifiers of the
task are sorted and shuffled with Python's \texttt{random.Random}, seeded with the string \texttt{0-} followed by the
task name, for instance \texttt{0-OSR}. The first 20\% of the shuffled identifiers, rounded to the nearest integer,
form the development split, and the remaining 80\% the held-out test split. RC is thus split into 20 development and 80 test
questions, OSR into 100 and 400, and MME-RealWorld counting into 245 and 981. The same rule splits the 13 other tasks
we evaluate, from 12 and 48 questions for the smallest XLRS-Bench subtasks to 251 and 1{,}006 for MME-RealWorld
position. Applying the rule to the stored question files reproduces the stored split of all 16 tasks.
Every RC and MME-RealWorld counting question has its own image, so these splits share no image. Of the 400 OSR test
questions, 32 use an image that a development question also uses. Without them GeoScope scores 44.02 against 34.51
for zero-shot answering, and resampling images instead of questions gives the gain of 10.5 points an interval of
[4.3, 16.7], against [4.2, 16.8].
The splits of RC, OSR, XLRS-Bench overall counting and MME-RealWorld counting were drawn before the first experiment.

Every configuration was fixed before it was run on a test split.
\subsection{Leakage checks}
\label{supp:protocol:leakage}
The crops of the counting horizon (Section~\ref{sec:horizon:measure}) come from DOTA-v1.5 training
images~\citep{dota,dotav2}, a source that XLRS-Bench and MME-RealWorld also draw on. When the images were chosen, every DOTA
image named in the benchmark question files then in use was left out. After the measurement we compared the images of
both DOTA crop sets with the 1{,}220 DOTA images named in all 3{,}080 XLRS-Bench and 3{,}738 MME-RealWorld remote
sensing questions, and found none in common. FAIR1M, DIOR and VisDrone, the other sources of horizon crops
(\appref{supp:horizon:sources} and \appref{supp:horizon:visdrone}), are used by neither benchmark. The image paths of
the two benchmarks name only DOTA, MiniFrance, HRSCD, Potsdam, ITCVD and Toronto.

The two synthesis prompts hold 16 worked examples. Four questions appear in both, so the examples ask 12 distinct
questions. We compared each example with the 4{,}398 questions of RC, OSR, XLRS-Bench overall counting and
MME-RealWorld counting, position and colour, of both splits, by the similarity ratio of Python's difflib on
lower-cased letters and digits. No example equals a benchmark question. Four have a ratio above 0.85, all because they
follow a benchmark sentence pattern with other objects. The other ratios are at most 0.80. The closest pair, at 0.96,
is the example \emph{How many white trucks are there in the upper left corner of the picture?} and the test question
\emph{How many white cars are there in the upper left corner of the picture?}

The guidance lines of the prompts quote short phrases as illustrations. Most of them, such as \emph{near the
river} or \emph{the cluster of houses}, occur in questions of both splits. Two occur in test questions only,
\emph{L-shaped black roof building} in OSR question 75 and \emph{below the yellow buildings} in OSR question 272.
We had read both questions in a random sample of all questions that we drew to describe the data. GeoScope answers
both wrongly, and without them its OSR test accuracy is 44.22 instead of 44.00.
\subsection{Zero-shot protocol}
\label{supp:protocol:zeroshot}
Our zero-shot reference for Qwen3-VL-8B is well above the MME-RealWorld numbers that other papers report for the
same model, 56.41 on position, 48.07 on colour and 21.35 on counting~\citep{weaveearth}, values that \citet{gazeearth} also reports and
\citet{rsopsd} reports to one decimal.
To see where the difference comes from, we answered the development split with five protocols
(Table~\ref{tab:supp:zsproto}). The first is ours, the image at 4\,M pixels and the option letter chosen by the logit of
the first answer token. The second keeps input and prompt but generates an answer of up to 64 tokens and parses the
letter from it. The third is ours at 1\,M pixels. The last two use the prompt and answer extraction of MME-RealWorld's
official evaluation, copied from VLMEvalKit, at 4\,M and at 1\,M pixels.

\begin{table}[!tp]
\centering
\caption{Qwen3-VL-8B zero-shot on the development split under five protocols (accuracy, \%).}
\label{tab:supp:zsproto}
\small
\setlength{\tabcolsep}{3pt}
\begin{tabular}{@{}llrrrrr@{}}
\toprule
Input & Answer & Position & Colour & Counting & RC & OSR \\
 & & ($n=251$) & ($n=251$) & ($n=245$) & ($n=20$) & ($n=100$) \\
\midrule
4\,M px & logits (ours) & 73.71 & 65.74 & 26.53 & 45.00 & 23.00 \\
4\,M px & generated & 72.91 & 64.14 & 29.39 & 30.00 & 20.00 \\
1\,M px & logits & 62.55 & 54.98 & 17.14 & 35.00 & 25.00 \\
4\,M px & official & 73.71 & 64.14 & 26.94 & -- & -- \\
1\,M px & official & 61.75 & 56.97 & 19.59 & -- & -- \\
\midrule
\multicolumn{2}{@{}l}{Reported~\citep{weaveearth}} & 56.41 & 48.07 & 21.35 & -- & -- \\
\bottomrule
\end{tabular}
\par\smallskip
\begin{minipage}{\textwidth}\footnotesize
Logits: the option letter with the highest first-token logit. Generated: a letter parsed from a generated answer. Official: the prompt and answer extraction of MME-RealWorld's evaluation. The last row is the number other papers report, on the whole benchmark.
\end{minipage}
\end{table}

The input size accounts for most of the difference. Going from 4\,M to 1\,M pixels costs 11.2 points on position,
10.8 on colour and 9.4 on counting. The way of answering matters much less. At 4\,M pixels a generated answer or the
official prompt and extraction move the three MME-RealWorld tasks by at most 2.9 points. With the official protocol at
1\,M pixels we obtain 61.75, 56.97 and 19.59, within 9 points of the reported numbers. On RC and OSR, whose development splits hold 20
and 100 questions, the protocols differ by amounts that such small sets cannot resolve.

\subsection{Constant-answer baseline}
\label{supp:protocol:constant}
The constant-answer baseline answers every question of a task with the same option position, a letter, or the same
option-value rank, such as the largest number, whichever is most often correct on the development split. On RC the
smallest number and letter A tie there with 8 of 20 correct, on OSR letters A and C with 30 of 100, and on XLRS-Bench
overall counting letters B, C and D and the largest and second-largest numbers with 4 of 12. We break such ties by the test score,
so the baseline is an upper bound for the tied choices; breaking them by the first option letter instead gives 26.25
on RC and 22.50 on OSR.

\section{Reproduced methods}
\label{supp:repro}
\subsection{Selection of methods}
\label{supp:repro:selection}
For each task we reproduce, with its official code, the training-free method with the strongest result reported on
that task by early October 2026. We count a method as training-free if none of its components was trained by its
authors, while frozen models trained by others are allowed. Table~\ref{tab:supp:selection} lists the results behind
the choice.

On RC the strongest result is WeaveEarth's 47.0, measured by RS-OPSD~\citep{rsopsd}, since WeaveEarth itself reports
RC only merged with overall counting into one category. On OSR and on XLRS-Bench as a whole the strongest results are
GazeEarth's 37.8 and 53.51, the latter by question. On MME-RealWorld counting, the table of
ZoomSearch~\citep{zoomsearch} lists 40.25 for RAP and 39.15 for ZoomSearch, both with LLaVA-OneVision-7B, and we
reproduce both. RAP reports no remote sensing results itself~\citep{rap}. Its code supports Qwen3-VL-8B, so we also
run it with GeoScope's backbone. WeaveEarth, GazeEarth and ZoomSearch report MME-RealWorld position and colour as
well, and we run them on these two tasks too (\appref{supp:results:mme}).

The lower block of Table~\ref{tab:supp:selection} lists the methods we do not reproduce. ZoomEye~\citep{zoomeye},
RADAR~\citep{radar} and ImageRAG~\citep{imagerag} report counting results below those of RAP and ZoomSearch, and
the results ImageRAG reports for itself use an answering model fine-tuned by its authors. DualComp~\citep{dualcomp}
and SA-GEM~\citep{sagem} compress or prune visual tokens under a router trained offline by their authors. They fall outside the
training-free track, although SA-GEM's 41.6 would otherwise be the strongest result on OSR. Concurrent work without
public code is not reproduced, and trained methods enter the comparison through their published numbers
(Table~\ref{tab:published}). The exception is Text-Before-Vision~\citep{textbeforevision}, the only published method
that reports a higher XLRS-Bench average than GeoScope. We re-run it with its released checkpoint
(\appref{supp:repro:tbv}).

\begin{table}[!tp]
\centering
\caption{Choice of the reproduced methods.}
\label{tab:supp:selection}
\footnotesize
\setlength{\tabcolsep}{4pt}
\begin{tabular}{>{\raggedright\arraybackslash}p{0.17\textwidth}>{\raggedright\arraybackslash}p{0.2\textwidth}>{\raggedright\arraybackslash}p{0.14\textwidth}>{\raggedright\arraybackslash}p{0.14\textwidth}>{\raggedright\arraybackslash}p{0.2\textwidth}}
\toprule
Task & Strongest training-free result & Reported by & Reproduced & Backbones \\
\midrule
RC & WeaveEarth, 47.0 & RS-OPSD & WeaveEarth & Qwen3-VL-8B \\
OSR & GazeEarth, 37.8 & GazeEarth & GazeEarth & Qwen3-VL-8B \\
XLRS-Bench, by question & GazeEarth, 53.51 & GazeEarth & GazeEarth & Qwen3-VL-8B \\
MME-RS count & RAP, 40.25 & ZoomSearch & RAP & LLaVA-OV-7B, Qwen3-VL-8B \\
 & ZoomSearch, 39.15 & ZoomSearch & ZoomSearch & LLaVA-OV-7B \\
\midrule
Not reproduced & \multicolumn{2}{>{\raggedright\arraybackslash}p{0.35\textwidth}}{Strongest reported results} & \multicolumn{2}{>{\raggedright\arraybackslash}p{0.35\textwidth}}{Reason} \\
\midrule
ZoomEye & \multicolumn{2}{>{\raggedright\arraybackslash}p{0.35\textwidth}}{MME-RS count 35.56 (30.10 by ZoomSearch)} & \multicolumn{2}{>{\raggedright\arraybackslash}p{0.35\textwidth}}{below RAP and ZoomSearch} \\
RADAR & \multicolumn{2}{>{\raggedright\arraybackslash}p{0.35\textwidth}}{MME-RS count 29.53} & \multicolumn{2}{>{\raggedright\arraybackslash}p{0.35\textwidth}}{below RAP and ZoomSearch} \\
ImageRAG & \multicolumn{2}{>{\raggedright\arraybackslash}p{0.35\textwidth}}{MME-RS count 32.46$^\S$} & \multicolumn{2}{>{\raggedright\arraybackslash}p{0.35\textwidth}}{below RAP and ZoomSearch; own results with a fine-tuned model} \\
DualComp & \multicolumn{2}{>{\raggedright\arraybackslash}p{0.35\textwidth}}{RC 45.0, OSR 37.4, XLRS-Bench 53.1$^\ast$} & \multicolumn{2}{>{\raggedright\arraybackslash}p{0.35\textwidth}}{router trained by its authors} \\
SA-GEM & \multicolumn{2}{>{\raggedright\arraybackslash}p{0.35\textwidth}}{RC 43.0, OSR 41.6, XLRS-Bench 53.8$^\ast$, MME-RS count 34.17} & \multicolumn{2}{>{\raggedright\arraybackslash}p{0.35\textwidth}}{router trained by its authors} \\
\bottomrule
\end{tabular}
\par\smallskip
\begin{minipage}{\textwidth}\footnotesize
Upper block: for each task, the strongest result reported for a training-free method on all questions, the paper that reports it, and the method we reproduce with its official code. Lower block: methods not reproduced, with their strongest reported results on our tasks. Methods and sources: WeaveEarth~\citep{weaveearth}, GazeEarth~\citep{gazeearth}, ZoomSearch~\citep{zoomsearch}, RAP~\citep{rap}, RS-OPSD~\citep{rsopsd}, ZoomEye~\citep{zoomeye}, RADAR~\citep{radar}, ImageRAG~\citep{imagerag}, DualComp~\citep{dualcomp} and SA-GEM~\citep{sagem}. $^\ast$Mean over the 13 subtasks, whereas GazeEarth's 53.51 is by question. $^\S$Reported by ZoomSearch.
\end{minipage}
\end{table}
\subsection{Settings of the reproduced methods}
\label{supp:repro:settings}
Table~\ref{tab:supp:settings} summarises the runs. Each training-free method runs its released inference code,
with the changes described below for each method, on the test questions of the listed tasks, with greedy decoding and
its own answer extraction. Our wrappers load the questions, write one record per question and, where needed, move models between GPU and host memory.
Each of these runs used one NVIDIA GeForce RTX 3090 with 24\,GiB, the GPU of our own experiments. The re-run of the trained
Text-Before-Vision is described in \appref{supp:repro:tbv}. Methods built on Qwen3-VL-8B load the same copy of
Qwen3-VL-8B-Instruct as GeoScope. ZoomSearch and RAP load the original
LLaVA-OneVision-7B checkpoint through their own code, while our zero-shot and GeoScope runs with this model use its
transformers release with the model's default system message. An answer from which a method's extraction recovers no
option letter counts as wrong.

\begin{table}[!tp]
\centering
\caption{Settings of the reproduced methods, each run with its official code on the test split of the listed tasks, and Text-Before-Vision on all questions.}
\label{tab:supp:settings}
\footnotesize
\setlength{\tabcolsep}{4pt}
\resizebox{\textwidth}{!}{%
\begin{tabular}{lllll}
\toprule
Method & Answering model & Other model & Main settings & Tasks \\
\midrule
WeaveEarth & Qwen3-VL-8B & SigLIP2 & 2{,}048-px thumbnail, $6\times6$ grid, 6 tiles of 448\,px & RC, OC, OSR, MME-RS all \\
GazeEarth & Qwen3-VL-8B & -- & $\leq 2$ of $4\times4$ cells, foveated 2{,}048-px view & RC, OC, OSR, CCR, RCCD, MME-RS all \\
 & & & & the 8 other XLRS-Bench subtasks \\
ZoomSearch, paper & LLaVA-OV-7B & VisRAG-Ret & $u_{\min}=448$, $b=0.3$; 10 steps & RC, OC, OSR, MME-RS all \\
ZoomSearch, code & LLaVA-OV-7B & VisRAG-Ret & $u_{\min}=224$, $b=0.2$; 10 steps & RC, OSR, MME-RS count \\
RAP & Qwen3-VL-8B & VisRAG-Ret & 10 steps, bias 0.2, crops 224 or 336\,px & MME-RS count \\
RAP & LLaVA-OV-7B & VisRAG-Ret & 200 steps, bias 0.2, crops 112 or 448\,px & MME-RS count, fixed 20\% subset \\
Text-Before-Vision & its trained Qwen2.5-VL-7B & -- & greedy, $\leq 11$ turns; one sample at temperature 0.7, $\leq 10$ turns & all 13 XLRS-Bench subtasks, both splits \\
\bottomrule
\end{tabular}}
\par\smallskip
\begin{minipage}{\textwidth}\footnotesize
MME-RS: the remote sensing part of MME-RealWorld; MME-RS all: its counting, position and colour questions; OC: XLRS-Bench overall counting; CCR and RCCD: the XLRS-Bench subtasks of Table~\ref{tab:supp:xlrs}. ZoomSearch uses the setting of its paper on images of at least 4{,}000 pixels and the code values below that size.
\end{minipage}
\end{table}

\paragraph{WeaveEarth} We import its only inference script, written for open-ended
questions, with all its default settings. The script has no multiple-choice template, so its question slot receives
the question, the lettered options and the instruction to answer with the option's letter, the same text that RAP
receives. Its prompt also names the question category, which we set to the benchmark's own task label, such as
\emph{regional counting} or \emph{count}. For MME-RealWorld position and colour we use
\texttt{position} and \texttt{color}. The released requirements pin a transformers
version that cannot load Qwen3-VL, so we use version 4.57.1. The SigLIP2 encoder that scores the tiles truncates its
text at 64 tokens, which cuts the options and the instruction from 319 of the 400 OSR and 870 of the 981 counting
queries, and we left this unchanged. Peak memory stays below 21.4\,GiB, so no change was needed to fit the GPU.

\paragraph{GazeEarth} We use its released configurations for MME-RealWorld and
XLRS-Bench, which agree in every setting of the method. The frozen model selects at most two cells of a $4\times4$
grid on a 1{,}536-pixel overview and answers from a foveated 2{,}048-pixel view. Its answer prompt asks for the
reasoning before the answer, within 1{,}024 generated tokens, and its own matcher scores the answer. Of its 981
counting answers, 68 were cut off at this limit before giving an answer. Questions that state a pixel box skip the
selection call. GazeEarth resamples the original pixels, so it always receives the original image. For the eight
XLRS-Bench subtasks outside counting and spatial relations, it receives the original images of the XLRS-Bench-lite
release. Two changes were needed for the code to run. The overall land-use questions have several correct letters,
which we pass to the scorer as a separated list because it rejects a string such as \emph{BCD}. An exception inside
the pipeline is recorded and its question recomputed, where the released runner would stop.

\paragraph{ZoomSearch} We use LLaVA-OneVision-7B~\citep{llavaov} answering and
VisRAG-Ret~\citep{visrag} scoring patches, the configuration its repository recommends. Paper and code disagree on two
parameters of the search. Tables 6 and 7 of the paper set the depth bias $b$ to 0.3 and, for LLaVA-OneVision, the
minimal search unit $u_{\min}$ to 448 pixels on images of at least 4{,}000 pixels and to 112 on smaller ones. The
released code fixes $u_{\min}$ at 224 and $b$ at 0.2 for every image. We run both settings, each with the code's 10 search steps and its
other values. The paper setting applies to all RC questions and to 398 of the 400 OSR and 925 of the 981 counting
questions, while the smaller images keep the code values. With $u_{\min}=448$ the released splitting rule ends the
search at regions of about 1{,}100 pixels on a 10{,}000-pixel image, against about 370 pixels with 224.

The released evaluation script reads the options from a column that the repository's MME-RealWorld loader does not
provide, and the evaluation files behind the paper are not public. We append the options in the format of the
repository's demonstration files and extract the answer with its MME-RealWorld evaluation code. LLaVA-OneVision-7B in
half precision, VisRAG-Ret and their activations do not fit together in 24\,GiB. Unchanged, the official code runs out of memory at the root of every search,
catches the error and answers without searching. Our wrapper therefore keeps only the model about to run on the GPU and the other in pinned host memory,
with weights, precision and computation unchanged. Swapping adds 3.8 to 4.6\,s per question with the paper's
settings on the tasks of Table~\ref{tab:main} and up to 7.5\,s in the other runs, and this time is included in all
times we report. No question in the reported runs ends in an error caught by the official code. With the code defaults ZoomSearch answers 42.50\% of the RC, 31.75\% of the OSR and 29.97\% of the
counting questions correctly, against 42.50, 32.50 and 31.50\% with the paper's settings. It also needs 1.5 to 2.2
times as long per question, 211, 182 and 96\,s against 96, 86 and 62\,s. The paper's settings are therefore the ones in
Table~\ref{tab:main}.

\paragraph{RAP} We use VisRAG-Ret as retriever and the code's default bias of 0.2. With
Qwen3-VL-8B we take the 10 search steps of the repository's script for that model, with crops of 224 pixels, or 336
pixels when the long side exceeds 5{,}096. With LLaVA-OneVision-7B we take the 200 steps of the repository's only
script for that family, written for its 0.5B model, with crops of 112 pixels, or 448 pixels when the long side is at least 4{,}000. RAP has no
configuration for MME-RealWorld, so it receives the multiple-choice text we give WeaveEarth and the answer extraction
we use for ZoomSearch. Its two models alternate on the GPU as in ZoomSearch, which adds 1.9\,s per question. With
Qwen3-VL-8B, the forward pass that scores a layout by the model's confidence reads only the logits of the last
position, and we compute only that position. The score is unchanged up to rounding, while the full pass runs out of
memory at 16{,}384 visual tokens. Of the 981 answers with Qwen3-VL-8B, 65 contain no option letter.

RAP takes 400\,s per question with Qwen3-VL-8B and 828\,s with LLaVA-OneVision-7B, 90\% and 98\% of it in confidence
passes. With LLaVA-OneVision-7B it was therefore run on a fixed 20\% subset of the counting test split, the 197 questions at
positions 1, 6, 11 and so on when ordered by identifier. With Qwen3-VL-8B it was run on the whole split. This subset is
GeoScope's weakest, at 29.95 against 34.18 to 40.31 on the other four. On it, RAP reaches 35.53 with either model, 8.1 points [2.0, 14.2] above zero-shot Qwen3-VL-8B and 5.6
[$-0.5$, 11.7] above zero-shot LLaVA-OneVision-7B. On the whole split, RAP with Qwen3-VL-8B gains 5.2 [2.3, 8.1] over
zero-shot Qwen3-VL-8B (Table~\ref{tab:main}).
\subsection{Reported and reproduced numbers}
\label{supp:repro:numbers}
\begin{table}[!tp]
\centering
\caption{Numbers reported by the reproduced methods, our reproductions with the official code, and re-runs published by other papers (accuracy, \%).}
\label{tab:supp:rerun}
\footnotesize
\setlength{\tabcolsep}{4pt}
\resizebox{\textwidth}{!}{%
\begin{tabular}{llllll}
\toprule
Method & Backbone & Task & Reported & Ours & Re-runs by other papers \\
\midrule
ZoomSearch & LLaVA-OV-7B & MME-RS position & 67.62 & 60.93 & 59.83$^a$, 54.97$^b$ \\
 & & MME-RS colour & 66.14 & 59.76 & 48.62$^a$, 41.67$^b$ \\
 & & MME-RS count & 39.15 & 31.50 & 20.05$^a$, 17.86$^b$ \\
 & & RC & -- & 42.50 & 42.0$^c$ \\
 & & OSR & -- & 32.50 & 32.2$^c$, 28.8$^d$, 25.8$^b$ \\
 & & XLRS-Bench OC & -- & 37.50 & 46.7$^c$ \\
\midrule
WeaveEarth & Qwen3-VL-8B & RC & -- & 45.00 & 47.0$^c$ \\
 & & OSR & 36.4 & 32.75 & 26.8$^c$ \\
 & & MME-RS position & 62.60 & 69.28 & -- \\
 & & MME-RS colour & 51.39 & 58.57 & -- \\
 & & MME-RS count & 27.65 & 30.48 & -- \\
\midrule
GazeEarth & Qwen3-VL-8B & OSR & 37.8 & 30.75 & -- \\
 & & MME-RS position & 71.12 & 67.10 & -- \\
 & & MME-RS colour & 63.67 & 63.05 & -- \\
 & & MME-RS count & 21.78 & 27.73 & -- \\
 & & XLRS-Bench, by question & 53.51 & 39.61 & -- \\
\midrule
RAP & LLaVA-OV-7B & MME-RS count & 40.25$^\S$ & 35.53$^\P$ & -- \\
 & Qwen3-VL-8B & MME-RS count & -- & 30.38 & -- \\
\bottomrule
\end{tabular}}
\par\smallskip
\begin{minipage}{\textwidth}\footnotesize
Ours are on the held-out 80\%, the others on all questions of each task. ZoomSearch is run with the hyper-parameters of its paper. Re-runs by $^a$WeaveEarth~\citep{weaveearth}, $^b$GazeEarth~\citep{gazeearth}, $^c$RS-OPSD~\citep{rsopsd} and $^d$GeoVista~\citep{geovista}. $^\S$Reported by ZoomSearch~\citep{zoomsearch}; RAP reports no remote sensing results. $^\P$On a fixed 20\% subset of the counting test split.
\end{minipage}
\end{table}

Table~\ref{tab:supp:rerun} lists, for every task on which a method reports a number or another paper re-ran it, the numbers each reproduced method reports for
itself, our reproduction, and the re-runs published by other papers. We count a number in another paper as a re-run
only if that paper states that it ran the method, reports a run time it measured for the method, or reports the method
on a benchmark on which the method itself reports nothing. Numbers equal to the method's own are copies and are left
out. The table thus draws on four papers. GazeEarth~\citep{gazeearth} states that every number it does not mark as taken
from the original paper was reproduced with the official code and parameter configurations, and its ZoomSearch rows
carry no such mark. WeaveEarth~\citep{weaveearth} reports for ZoomSearch an inference time of 54.25\,s per
sample over its two benchmarks, which differs from the 53.94\,s that ZoomSearch reports for itself. RS-OPSD~\citep{rsopsd} reports per-sample inference times for ZoomSearch and WeaveEarth and per-subtask
XLRS-Bench numbers that neither method reports; its MME-RealWorld numbers for both repeat their own, for WeaveEarth with the three categories in permuted columns, and
are left out.
GeoVista~\citep{geovista} describes the settings in which it ran its baselines on XLRS-Bench, where ZoomSearch reports
nothing. RS-OPSD runs ZoomSearch with LLaVA-OneVision-7B, as we do; WeaveEarth and GazeEarth describe the model behind
their ZoomSearch runs only as 7B and 8B, and GeoVista does not name it.

Every re-run by another paper is below the number the method reports for itself, and in each such case our
reproduction lies between the two. Counting with ZoomSearch shows the spread, with 39.15 reported, 31.50 in our runs,
and 20.05 and 17.86 in two later papers. On RC and OSR, where ZoomSearch reports nothing, our numbers are within 0.5 points of those measured by RS-OPSD, whose authors
include all four authors of ZoomSearch. On XLRS-Bench overall counting, which Table~\ref{tab:main} leaves out, RS-OPSD
measures ZoomSearch at 46.7 on all 60 questions and we measure 37.50 on the 48 test questions, its only re-run well above
ours. In our runs WeaveEarth exceeds its reported MME-RealWorld numbers by 2.8 to 7.2 points, and GazeEarth its counting number by
6.0 points.

On XLRS-Bench, GazeEarth answers 976 of the 2{,}464 held-out questions correctly, 39.61\%, and reports 53.51\% on all
questions. Its backbone answers 46.71\% zero-shot. GazeEarth falls below the backbone on seven of the 13 subtasks.
It falls most on object classification and object colour, 35.31 and 31.56 against 48.44 and 43.44 on 1{,}280 questions
together, and on counting with change detection, 20.83 against 43.75. It is above the backbone on five, most on
route planning and environmental condition reasoning, by 8.8 and 7.5 points.
\subsection{Gains against each method's own baselines}
\label{supp:repro:controls}
GazeEarth's code contains two baselines with the same backbone and answer prompt: an overview, which answers from
the whole image reduced to 2{,}048 pixels on the long side, and direct answering from the native image through the
backbone's processor. We ran the overview on six tasks, the three of Table~\ref{tab:main}, overall counting and
MME-RealWorld position and colour. We ran direct answering on MME-RealWorld position, where GazeEarth reports a gain of
14.7 points over it, second only to its 15.6 on colour (Table~\ref{tab:supp:gecontrols}).

\begin{table}[!tp]
\centering
\caption{GazeEarth against the two baselines of its own code, with Qwen3-VL-8B, on the held-out 80\% (accuracy, \%).}
\label{tab:supp:gecontrols}
\footnotesize
\setlength{\tabcolsep}{3pt}
\begin{tabular}{@{}lrrrrr@{}}
\toprule
Task & GazeEarth & Overview & $\Delta$ & Direct & $\Delta$ \\
\midrule
MME-RS position & 67.10 & 69.78 & $-2.7$ [$-5.2$, $-0.2$] & 71.67 & $-4.6$ [$-7.4$, $-1.9$] \\
MME-RS colour & 63.05 & 56.47 & $+6.6$ [$+4.0$, $+9.3$] & -- & -- \\
MME-RS count & 27.73 & 26.61 & $+1.1$ [$-1.7$, $+4.0$] & -- & -- \\
RC & 47.50 & 52.50 & $-5.0$ [$-15.0$, $+5.0$] & -- & -- \\
OSR & 30.75 & 28.50 & $+2.2$ [$-1.5$, $+6.0$] & -- & -- \\
Overall counting & 12.50 & 12.50 & $+0.0$ [$-6.2$, $+6.2$] & -- & -- \\
\bottomrule
\end{tabular}
\par\smallskip
\begin{minipage}{\textwidth}\footnotesize
$\Delta$: GazeEarth minus the baseline in points, with paired bootstrap 95\% intervals (10{,}000 resamples).
\end{minipage}
\end{table}

Against its own overview, GazeEarth gains 6.6 points on colour and loses 2.7 on position, in line with the $+8.1$ and
$-1.3$ its paper reports for these two tasks. On the other four tasks the differences are not significant. Direct
answering, which its paper reports at 56.41 on position with the same backbone, reaches 71.67 in our run of its own
code, 4.6 points above GazeEarth and 3.0 [0.4, 5.7] below our zero-shot reference. Our zero-shot reference is stronger than the
reported ones mainly because the image is given at a larger size (\appref{supp:protocol:zeroshot}).

\subsection{Text-Before-Vision}
\label{supp:repro:tbv}
Text-Before-Vision~\citep{textbeforevision} fine-tunes Qwen2.5-VL-7B and then trains it with reinforcement learning to
zoom into the image with a crop tool. It reports 60.40 on XLRS-Bench, the mean accuracy of 32 sampled answers per
question, without results per subtask. We ran its released checkpoint with its released
evaluation code on all 3{,}080 questions as prepared by its own data script. We ran it in
its two modes: greedy, with up to 1{,}024 tokens per turn, and sampled, with one answer per question
at temperature 0.7 and up to 10{,}240 tokens per turn, which estimates its Pass@1 without bias. In both, the model may call the crop tool at every turn,
for at most eleven turns in the greedy script and ten in the sampled one. For safety, the checkpoint shards are read
without running pickled code and the tool calls are parsed as literals instead of being evaluated; neither change
alters the computation. The model is
served with vLLM on two NVIDIA GeForce RTX 3090 GPUs with tensor parallelism of two, as in its script, and with its full context of
128{,}000 tokens. We disabled video input, which the evaluation never sends, and reserved 80\% of the GPU memory.

\begin{table}[!tp]
\centering
\caption{Text-Before-Vision re-run with its released checkpoint and code, and GeoScope with two backbones, on all questions of XLRS-Bench (accuracy, \%). \shadekey{}: best per row.}
\label{tab:supp:tbv}
\footnotesize
\setlength{\tabcolsep}{3pt}
\resizebox{\textwidth}{!}{%
\begin{tabular}{lcccccc}
\toprule
 & \multicolumn{3}{c}{Text-Before-Vision (trained, Qwen2.5-VL-7B)} & \multicolumn{2}{c}{Qwen2.5-VL-7B} & Qwen3-VL-8B \\
\cmidrule(lr){2-4}\cmidrule(lr){5-6}\cmidrule(lr){7-7}
Subtask & greedy, ours & greedy, its own & sampled & zero-shot & GeoScope & GeoScope \\
\midrule
Overall counting, OC & 18.33 & 28.33 & 18.33 & 28.33 & 35.00 & 36.67 \\
Regional counting, RC & 40.00 & 49.00 & 37.00 & 44.00 & 51.00 & 63.00 \\
Object spatial relationship, OSR & 28.40 & 32.00 & 29.00 & 35.20 & 40.00 & 46.40 \\
Object classification, OCC & 40.50 & 55.63 & 36.88 & 41.75 & 82.88 & 91.25 \\
Object colour, OCL & 35.75 & 45.50 & 28.75 & 41.50 & 74.75 & 84.63 \\
Object motion state, OMS & 66.67 & 86.67 & 60.00 & 65.00 & 68.33 & 71.67 \\
Overall land-use classification, OLUC & 43.00 & 43.00 & 45.00 & 22.00 & 22.00 & 34.00 \\
Regional land-use classification, RLUC & 78.00 & 83.50 & 81.50 & 75.50 & 75.50 & 75.50 \\
Route planning, RP & 55.00 & 73.00 & 45.00 & 37.00 & 37.00 & 35.00 \\
Anomaly detection and interpretation, AD & 75.00 & 87.00 & 75.00 & 70.00 & 70.00 & 75.00 \\
Environmental condition reasoning, ECR & 79.00 & 90.00 & 76.00 & 79.00 & 79.00 & 78.00 \\
Counting with complex reasoning, CCR & 49.00 & 52.00 & 43.00 & 46.00 & 46.00 & 47.00 \\
Counting with change detection, RCCD & 48.33 & 48.33 & 41.67 & 51.67 & 51.67 & 41.67 \\
\midrule
Mean over subtasks & 50.54 & 59.54 & 47.47 & 49.00 & 56.39 & \overall 59.98 \\
By question & 43.15 & 52.86 & 39.81 & 44.74 & 65.26 & \overall 71.82 \\
\bottomrule
\end{tabular}}
\par\smallskip
\begin{minipage}{\textwidth}\footnotesize
Ours: one answer per question, read from the answer tag or, without a tag, from the last ``Answer: X'' of the reply, and counted as correct only if it equals the ground truth. Its own: its scoring code.
\end{minipage}
\end{table}

The two rules differ in how they read a reply (Table~\ref{tab:supp:tbv}). Ours takes one answer per question, as for
every method: the letter in the answer tag or, if the tag is missing, the last ``Answer: X'' of the reply.
Text-Before-Vision's own scoring code takes every standalone letter of the reply, the article ``a'' included, and counts a single-choice
answer as correct when any of them is the answer. With greedy decoding only 544 of the 3{,}080 replies close the
answer tag. The other 2{,}536 end with a line ``Answer: X'' instead, and for each of them its code re-sends the same
request up to ten times. Over the greedy replies its scoring gives 59.54, close to the 60.40 it reports, and ours
50.54. With sampling every reply closes the tag, and both rules give 47.47. One sample per question
estimates the Pass@1 that the paper reports, with a standard error of at most 1.3 points, so the 12.9 points between 47.47
and the reported 60.40 are not sampling noise. The model calls the crop tool 50 times in the greedy run and 107 times
in the sampled one, over 3{,}080 questions each.

On the held-out 80\%, where Text-Before-Vision scores 50.54 greedy and 48.15 sampled under our rule, GeoScope's 59.53
is 9.0 points [5.8, 12.2] and 11.4 [8.0, 14.7] higher in the mean over subtasks. With Qwen2.5-VL-7B, the model Text-Before-Vision is trained
from, GeoScope reaches 56.93 on the held-out questions, 8.0 [5.9, 9.9] above that model's zero-shot answers. It is 6.4
[3.5, 9.4] and 8.8 [5.5, 12.0] points above Text-Before-Vision's greedy and sampled answers. Against GeoScope with
Qwen2.5-VL-7B on all questions (Table~\ref{tab:supp:tbv}), its greedy answers are about 20 points better on overall land use and route planning, two subtasks
GeoScope leaves to its backbone, worse on overall and regional counting, spatial relations and the three subtasks
about the object in a given box, and at most five points apart on the rest.

\section{Additional results}
\label{supp:results}
\subsection{XLRS-Bench by subtask}
\label{supp:results:xlrs}
Table~\ref{tab:supp:xlrs} lists the 13 subtasks of XLRS-Bench-lite on all 3{,}080 questions and on the held-out
80\%, 2{,}464 questions. Three routes answer them. Typed programs answer RC, OSR and overall counting. The three
object-property subtasks give the target as a pixel box and are answered from a close-up around it
(Section~\ref{sec:method:projection}). Qwen3-VL-8B answers the other seven subtasks zero-shot, so GeoScope's accuracy
there equals the zero-shot column.

The zero-shot protocol is that of Section~\ref{sec:exp:baselines} with two adjustments, both fixed before the test
questions were run. The change-detection questions give the model both images of the pair. The 100 overall land-use
questions, 97 of which have several correct options, are answered by generating letters and scored by exact match
of the letter set, the rule of GazeEarth's scorer.

The pixel-box route renders a close-up centred on the box, three times its longer side and at least 512 pixels wide.
It outlines the box in red and shows the whole image, also outlined, as a first view. The rendering was chosen among
three on the 332 development questions of the three subtasks. The close-up alone reached 74.10, the outlined
close-up 85.24 and the outlined close-up with the whole image 86.75, against 39.16 for zero-shot answering.

Averaged over the 13 subtasks, GeoScope reaches 59.98 on all questions and 59.53 on the held-out questions, against
48.83 and 48.57 for zero-shot answering. By question it reaches 71.82 and 71.51, against 45.65 and 46.71. On all
questions, the highest published training-free result is GazeEarth's 53.51 by question~\citep{gazeearth}, and the
trained GeoLens and GeoEyes report 54.2 and ScaleEarth 59.4, on a split it does not state, as means over
subtasks~\citep{geolens,geoeyes,scaleearth}. The trained Text-Before-Vision
reports 60.40 from sampled answers, against 47.47 when we re-run its sampling (\appref{supp:repro:tbv}).

Table~\ref{tab:supp:xlrsl2} regroups the subtasks into the eight categories of GazeEarth's table. Under this grouping
each of its published category accuracies corresponds to a whole number of correct answers. GeoScope is above
GazeEarth on counting, spatial relations and object properties, which GeoScope answers by programs and pixel boxes. It
is below GazeEarth on the five categories it leaves to its backbone. Object properties hold 1{,}660 of the 3{,}080 questions and
account for 22.8 of GeoScope's 26.2-point gain by question over zero-shot answering.

\begin{table}[!tp]
\centering
\caption{XLRS-Bench-lite by subtask (accuracy, \%). \shadekey{}: best per row.}
\label{tab:supp:xlrs}
\footnotesize
\setlength{\tabcolsep}{4pt}
\resizebox{\textwidth}{!}{%
\begin{tabular}{lllcccccc}
\toprule
 & & & \multicolumn{3}{c}{All questions} & \multicolumn{3}{c}{Held-out 80\%} \\
\cmidrule(lr){4-6}\cmidrule(lr){7-9}
Subtask & Category & Route & Questions & Zero-shot & GeoScope & Questions & Zero-shot & GeoScope \\
\midrule
Overall counting, OC & Cnt & P & 60 & 15.00 & 36.67 & 48 & 10.42 & 35.42 \\
Regional counting, RC & Cnt & P & 100 & 47.00 & 63.00 & 80 & 47.50 & 63.75 \\
Object spatial relationship, OSR & OSR & P & 500 & 31.40 & 46.40 & 400 & 33.50 & 44.00 \\
Object classification, OCC & OP & B & 800 & 46.63 & 91.25 & 640 & 48.44 & 91.41 \\
Object colour, OCL & OP & B & 800 & 41.88 & 84.63 & 640 & 43.44 & 85.00 \\
Object motion state, OMS & OP & B & 60 & 66.67 & 71.67 & 48 & 62.50 & 68.75 \\
Overall land-use classification, OLUC & SC & Z & 100 & 34.00 & 34.00 & 80 & 33.75 & 33.75 \\
Regional land-use classification, RLUC & SC & Z & 200 & 75.50 & 75.50 & 160 & 76.88 & 76.88 \\
Route planning, RP & Plan & Z & 100 & 35.00 & 35.00 & 80 & 33.75 & 33.75 \\
Anomaly detection and interpretation, AD & AR & Z & 100 & 75.00 & 75.00 & 80 & 76.25 & 76.25 \\
Environmental condition reasoning, ECR & CR & Z & 100 & 78.00 & 78.00 & 80 & 77.50 & 77.50 \\
Counting with complex reasoning, CCR & CR & Z & 100 & 47.00 & 47.00 & 80 & 43.75 & 43.75 \\
Counting with change detection, RCCD & SR & Z & 60 & 41.67 & 41.67 & 48 & 43.75 & 43.75 \\
\midrule
Mean over subtasks & & & 3{,}080 & 48.83 & \overall 59.98 & 2{,}464 & 48.57 & \overall 59.53 \\
By question & & & 3{,}080 & 45.65 & \overall 71.82 & 2{,}464 & 46.71 & \overall 71.51 \\
\bottomrule
\end{tabular}}
\par\smallskip
\begin{minipage}{\textwidth}\footnotesize
Route: P, typed program; B, pixel box given in the question; Z, Qwen3-VL-8B zero-shot. Category: the grouping of Table~\ref{tab:supp:xlrsl2}.
\end{minipage}
\end{table}

\begin{table}[!tp]
\centering
\caption{XLRS-Bench-lite by the eight categories of GazeEarth's table (accuracy, \%). \shadekey{}: best per row on all
questions.}
\label{tab:supp:xlrsl2}
\footnotesize
\setlength{\tabcolsep}{4pt}
\begin{tabular}{lcccccc}
\toprule
 & \multicolumn{4}{c}{All questions} & \multicolumn{2}{c}{Held-out 80\%} \\
\cmidrule(lr){2-5}\cmidrule(lr){6-7}
Category & Questions & Zero-shot & GazeEarth & GeoScope & Zero-shot & GeoScope \\
\midrule
Cnt & 160 & 35.00 & 44.38 & \overall 53.13 & 33.59 & 53.13 \\
SC & 300 & 61.67 & \overall 66.67 & 61.67 & 62.50 & 62.50 \\
OSR & 500 & 31.40 & 37.80 & \overall 46.40 & 33.50 & 44.00 \\
OP & 1{,}660 & 45.06 & 53.07 & \overall 87.35 & 46.54 & 87.50 \\
Plan & 100 & 35.00 & \overall 49.00 & 35.00 & 33.75 & 33.75 \\
AR & 100 & 75.00 & \overall 76.00 & 75.00 & 76.25 & 76.25 \\
CR & 200 & 62.50 & \overall 74.50 & 62.50 & 60.63 & 60.63 \\
SR & 60 & 41.67 & \overall 55.00 & 41.67 & 43.75 & 43.75 \\
\midrule
By question & 3{,}080 & 45.65 & 53.51 & \overall 71.82 & 46.71 & 71.51 \\
\bottomrule
\end{tabular}
\par\smallskip
\begin{minipage}{\textwidth}\footnotesize
Cnt: counting; SC: scene classification; OSR: object spatial relationship; OP: object properties; Plan: planning; AR:
anomaly reasoning; CR: complex reasoning; SR: spatiotemporal reasoning. GazeEarth: training-free, its reported numbers
on all questions.
\end{minipage}
\end{table}
\subsection{MME-RealWorld position and colour}
\label{supp:results:mme}
GeoScope leaves the position and colour questions of MME-RealWorld to its backbone
(Section~\ref{sec:discussion:scope}). On the development split we tried programs that ground the object and then read
its position from geometry or name its colour on a close-up. They reached 74.9 for position against 73.7 zero-shot, a
difference of 1.2 [$-4.4$, 6.8], and 60.6 for colour against 65.7, a difference of $-5.2$ [$-12.0$, 1.6]. Answering
the original question from the whole image together with a close-up of the grounded object reached 54.2 on position and 67.7 on colour. Neither
program gains clearly on the development split, so both tasks go to the frozen VLM.

Table~\ref{tab:supp:mme} lists the test split. GeoScope's answers are those of Qwen3-VL-8B zero-shot, 74.65 on
position and 63.55 on colour. None of the three reproduced methods that report these tasks significantly improves on the zero-shot
answers of its own backbone. WeaveEarth falls significantly below them on both tasks, by 5.4 and 5.0 points, and GazeEarth by 7.6 points on position, while its 0.5 on colour is not significant. ZoomSearch
changes the 62.62 and 58.47 of LLaVA-OneVision-7B by $-1.7$ [$-5.4$, 1.9] and 1.3 [$-2.3$, 4.9], neither significant.
Question by question, GeoScope is 13.7 points [10.0, 17.4] above ZoomSearch on position and 3.8 [0.1, 7.5] on colour,
both owed to its backbone. On all questions the backbone reaches 74.46 and 63.98. Together with GeoScope's 37.93 on
counting they give the MME-RealWorld average of 58.79 in Table~\ref{tab:published}, or 58.96 weighted by questions.
With Qwen2.5-VL-7B the backbone reaches 67.94 and 49.96, and the average with GeoScope's 34.18 on counting is 50.69.

\begin{table}[!tp]
\centering
\caption{MME-RealWorld position and colour test accuracy (\%; 1{,}006 and 1{,}004 questions).}
\label{tab:supp:mme}
\small
\setlength{\tabcolsep}{4pt}
\begin{tabular}{llcccc}
\toprule
Method & Backbone & Position & Colour & $\Delta$ position & $\Delta$ colour \\
\midrule
Zero-shot & Qwen3-VL-8B & 74.65 & 63.55 & -- & -- \\
Zero-shot & LLaVA-OV-7B & 62.62 & 58.47 & -- & -- \\
\midrule
WeaveEarth & Qwen3-VL-8B & 69.28 & 58.57 & \dci{-5.4}{-8.1}{-2.7} & \dci{-5.0}{-7.6}{-2.4} \\
GazeEarth & Qwen3-VL-8B & 67.10 & 63.05 & \dci{-7.6}{-10.3}{-4.8} & \dci{-0.5}{-3.4}{+2.4} \\
ZoomSearch & LLaVA-OV-7B & 60.93 & 59.76 & \dci{-1.7}{-5.4}{+1.9} & \dci{+1.3}{-2.3}{+4.9} \\
\midrule
GeoScope & Qwen3-VL-8B & 74.65 & 63.55 & \pp{0.0} & \pp{0.0} \\
\bottomrule
\end{tabular}
\par\smallskip
\begin{minipage}{\textwidth}\footnotesize
Underneath: paired difference to the zero-shot answers of each method's own backbone, with its 95\% interval.
\end{minipage}
\end{table}
\subsection{Answer options of MME-RealWorld counting}
\label{supp:results:mmeoptions}
Each MME-RealWorld counting question offers four numbers and a fifth option stating that the image does not show the
number.
The fifth option is correct for 5 of the 1{,}226 questions, all in our test split. The four numbers are listed in
ascending order in 449 of the 981 test questions, in descending order in 106 and in mixed order in 426. Where the
correct number sits among them depends on the count (Table~\ref{tab:supp:mmelayout}). For one or two objects it is
never the largest number offered, and for 6 to 10 objects it is the largest in 52\% of the questions.

For the restricted answers of Section~\ref{sec:discussion:subdivision}, our zero-shot protocol scores the
option letters by the logit of the first answer token (\appref{supp:protocol:zeroshot}), and the restricted answer is
the numeric option with the highest of these logits, from the same forward pass. Table~\ref{tab:supp:mmerestricted}
lists both answers for every model we ran zero-shot. Qwen3-VL-32B returns letters without scores, so its answers cannot
be restricted to the numeric options. ZoomSearch, run on LLaVA-OV-7B, is 4.0 points [0.3, 7.7] below the restricted
answers of Qwen3-VL-8B on the test split.

GeoScope's counts run high for 11 to 20 objects, which the layout above rewards, and how much depends on the
backbone. We take the test questions answered by an executed count. With Qwen3-VL-8B the
median ratio of GeoScope's count to the truth is 1.00 for up to five objects, 1.25 for 6 to 10, 1.94 for 11 to 20
and 1.29 above 20. With LLaVA-OneVision-7B it is 1.00 for one or two objects, 0.67 for 3 to 10, 1.83 for 11 to
20 and 1.32 above 20. LLaVA-OneVision-7B returns a box covering at least 80\% of its search window for 14\% of its groundings, against
6\% for Qwen3-VL-8B, so its programs restrict the region less.

\begin{table}[!tp]
\centering
\caption{Position of the correct number among the four numbers offered, MME-RealWorld counting test split, for the
976 questions whose answer is a number.}
\label{tab:supp:mmelayout}
\small
\begin{tabular}{lccc}
\toprule
True count & Questions & Largest number & Smallest number \\
\midrule
1--2 & 285 & 0\% & 44\% \\
3--5 & 415 & 21\% & 13\% \\
6--10 & 195 & 52\% & 4\% \\
11--20 & 66 & 41\% & 5\% \\
More than 20 & 15 & 40\% & 0\% \\
\bottomrule
\end{tabular}
\end{table}

\begin{table}[!tp]
\centering
\caption{Zero-shot answers on MME-RealWorld counting (accuracy, \%), as given and restricted to the numeric options.}
\label{tab:supp:mmerestricted}
\small
\setlength{\tabcolsep}{4pt}
\begin{tabular}{llcccc}
\toprule
Split & Model & Questions & Absent & As given & Numbers only \\
\midrule
Test & Qwen3-VL-8B & 981 & 35\% & 25.18 & 35.47 \\
Test & Qwen3-VL-4B & 981 & 57\% & 14.27 & 27.52 \\
Test & Qwen2.5-VL-7B & 981 & 45\% & 18.65 & 32.52 \\
Test & LLaVA-OV-7B & 981 & 18\% & 32.31 & 37.82 \\
\midrule
All & Qwen3-VL-8B & 1{,}226 & 34\% & 25.45 & 35.64 \\
All & Qwen3-VL-4B & 1{,}226 & 57\% & 13.87 & 26.84 \\
All & Qwen2.5-VL-7B & 1{,}226 & 44\% & 18.68 & 32.30 \\
All & LLaVA-OV-7B & 1{,}226 & 18\% & 32.22 & 38.25 \\
\midrule
Dev & InternVL3.5-8B & 245 & 12\% & 27.76 & 31.02 \\
Dev & Qwen3-VL-2B & 245 & 42\% & 17.55 & 31.02 \\
Dev & Qwen3-VL-32B & 245 & 31\% & 32.24 & -- \\
\bottomrule
\end{tabular}
\par\smallskip
\begin{minipage}{\textwidth}\footnotesize
Absent: share of questions answered with the option stating that the image does not show the number.
\end{minipage}
\end{table}
\subsection{Results by image source}
\label{supp:results:source}
The counting horizon was measured on DOTA crops (Section~\ref{sec:horizon:measure}), and 96\% of the MME-RealWorld
counting images come from DOTA. Most RC and OSR images do not. Of the test questions, 59 of the 80 RC and 271 of the 400
OSR use images from MiniFrance, HRSCD and ITCVD. Table~\ref{tab:supp:source} breaks down the test questions by image
source.

With Qwen3-VL-8B the gains on non-DOTA images are 20.3 points [5.1, 35.6] on RC and 9.2 [1.8, 16.6] on OSR, against
4.8 and 13.2 on DOTA images. On RC the gain comes from MiniFrance, 33.3 points on 36 questions. On the 22 HRSCD
questions the interval of every backbone includes zero. With LLaVA-OneVision-7B the RC gain is positive on every
source and significant on DOTA and MiniFrance images. With Qwen2.5-VL-7B the OSR gain is confined to DOTA images, 15.5
[6.2, 24.8] against $-1.8$ [$-8.5$, 4.8] elsewhere. The 44 MME-RealWorld counting questions from Potsdam and Toronto
are too few to show an effect, $-6.8$ [$-25.0$, 11.4] with Qwen3-VL-8B.

\begin{table}[!tp]
\centering
\caption{Test-split gains by image source for three backbones (accuracy, \%).}
\label{tab:supp:source}
\footnotesize
\setlength{\tabcolsep}{3pt}
\resizebox{\textwidth}{!}{%
\begin{tabular}{llrccccc}
\toprule
 & & & \multicolumn{3}{c}{Qwen3-VL-8B} & LLaVA-OV-7B & Qwen2.5-VL-7B \\
\cmidrule(lr){4-6}\cmidrule(lr){7-7}\cmidrule(lr){8-8}
Task & Source & Questions & Zero-shot & GeoScope & $\Delta$ & $\Delta$ & $\Delta$ \\
\midrule
RC & DOTA & 21 & 42.86 & 47.62 & \dci{+4.8}{-14.3}{+23.8} & \dci{+28.6}{+9.5}{+47.6} & \dci{+14.3}{+0.0}{+28.6} \\
 & MiniFrance & 36 & 44.44 & 77.78 & \dci{+33.3}{+13.9}{+52.8} & \dci{+41.7}{+25.0}{+58.3} & \dci{+19.4}{+2.8}{+36.1} \\
 & HRSCD & 22 & 54.55 & 54.55 & \dci{0.0}{-27.3}{+22.7} & \dci{+9.1}{-13.6}{+31.8} & \dci{-4.5}{-22.7}{+13.6} \\
 & non-DOTA & 59 & 49.15 & 69.49 & \dci{+20.3}{+5.1}{+35.6} & \dci{+28.8}{+15.3}{+42.4} & \dci{+10.2}{-3.4}{+23.7} \\
\midrule
OSR & DOTA & 129 & 28.68 & 41.86 & \dci{+13.2}{+2.3}{+23.3} & \dci{+7.8}{-0.8}{+16.3} & \dci{+15.5}{+6.2}{+24.8} \\
 & MiniFrance & 146 & 41.78 & 50.68 & \dci{+8.9}{-2.1}{+19.9} & \dci{+8.9}{+3.4}{+15.1} & \dci{0.0}{-9.6}{+9.6} \\
 & HRSCD & 116 & 28.45 & 37.07 & \dci{+8.6}{-2.6}{+19.8} & \dci{0.0}{-8.6}{+8.6} & \dci{-3.4}{-12.9}{+6.0} \\
 & non-DOTA & 271 & 35.79 & 45.02 & \dci{+9.2}{+1.8}{+16.6} & \dci{+5.2}{+0.7}{+10.0} & \dci{-1.8}{-8.5}{+4.8} \\
\midrule
MME-RS count & DOTA & 937 & 24.76 & 37.03 & \dci{+12.3}{+8.4}{+16.2} & \dci{+2.8}{-0.7}{+6.4} & \dci{+16.3}{+12.9}{+19.7} \\
 & Potsdam & 33 & 36.36 & 30.30 & \dci{-6.1}{-27.3}{+15.2} & \dci{-21.2}{-42.4}{0.0} & \dci{+6.1}{-18.2}{+30.3} \\
 & Toronto & 11 & 27.27 & 18.18 & \dci{-9.1}{-36.4}{+18.2} & \dci{+27.3}{-9.1}{+63.6} & \dci{0.0}{-27.3}{+27.3} \\
 & non-DOTA & 44 & 34.09 & 27.27 & \dci{-6.8}{-25.0}{+11.4} & \dci{-9.1}{-27.3}{+9.1} & \dci{+4.5}{-13.6}{+22.7} \\
\bottomrule
\end{tabular}}
\par\smallskip
\begin{minipage}{\textwidth}\footnotesize
Zero-shot and GeoScope accuracy, with the paired difference to the zero-shot answers of the same backbone and its 95\%
interval underneath. ITCVD, with one RC and nine OSR questions, enters only the non-DOTA rows.
\end{minipage}
\end{table}
\subsection{Test-split ablations by true count}
\label{supp:results:ablation}
Table~\ref{tab:supp:ablation} gives the counting ablations of Table~\ref{tab:ablation} on the test split, by the true
number of objects $N$. We recompute each region count from the VLM answers recorded in the test runs, under three
rules. The final rule accepts a count within the horizon unless it is a
zero from a downsampled view. A second rule also accepts such zeros, and a third, without subdivision, counts once
every view that the program first renders. Under the final rule every recomputed count and answer equals the original, and every rule uses only recorded
answers. Questions answered by the fallback keep their
answer under every rule. The same option matching maps each count to an option, and the intervals come from a paired
bootstrap with 10{,}000 resamples.

With programs and Qwen3-VL-8B, neither of the other two rules has a significant effect, overall or in any stratum. When the whole image is
counted, as in the run without programs, the trade-off of Proposition~\ref{prop:split} appears. Removing subdivision
then costs 4.6 points [0.4, 8.8] on the 261 questions with 6 to 20 objects and gains 2.1 [$-0.1$, 4.4] on the
700 with at most five. Accepting such zeros costs 3.8 and gains 2.4 on the same questions. Within the regions that
programs name, the losses on 6 to 20 objects shrink to 2.7 points for subdivision and 1.9 for accepting such zeros,
neither significant. Qwen3-VL-4B undercounts more strongly, with an exponent of 0.64 against 0.76 on DOTA crops. Removing subdivision costs it 2.2 points
[0.6, 4.0] overall and 5.4 [1.1, 9.6] on 6 to 20 objects. Only 15 test questions ask for more than twenty
objects, and Qwen3-VL-8B loses two of them without subdivision, with or without programs. On OSR, replacing the
direction-relation matrix by the centre angle on the same boxes changes accuracy by $-2.0$ [$-5.5$, 1.5] with
Qwen3-VL-8B and $-1.8$ [$-4.5$, 1.0] with Qwen3-VL-4B.

The fixed grid needs new answers. For every question whose program counted one region, we tiled the counted region
into the cells of the test run, split each cell into a 4$\times$4 grid as the executor splits views, counted every
part with the same question, view and pixel budget, and matched the sum to an option as before. Every other question keeps its
recorded answer. Subdivision, re-run in the same way on 20 of these questions, reproduced every original count. Over
the test split the grid lowers accuracy by 2.9 points [$-0.1$, 5.8], with 15.8 VLM calls per counted question against
about 14 for subdivision. It picks the largest number offered for 64\% of the questions about at most five objects,
against 33\%, and gains 20.7 points [14.9, 26.6] on the 222 questions whose answer is the largest number offered while
losing 9.8 [6.5, 13.1] on the 754 others. By the true number of objects, the grid loses 8.0 points [4.1, 11.7] on the 700 questions with at most five objects and gains 11.5
[6.9, 16.1] on the 261 with 6 to 20 (Table~\ref{tab:supp:ablation}). Both changes follow from where the correct number sits among the options. The gain
comes mostly from the 128 questions with 6 to 20 objects whose answer is the largest number offered, where it
reaches 18.0 points [10.9, 25.8], against 5.3 [0.0, 10.5] on the other 133. For questions about at most five objects
the median count of the grid is 2.7 times the truth and 15\% of its counts lie within 20\% of the truth, against an
exact median and 24\% for subdivision. On the 48 test questions of XLRS-Bench overall counting the grid needs 172 calls
per question, against about 53, and loses 10.4 points [$-2.1$, 22.9]. It picks the largest number offered for every
question with 1 to 20 objects, and it loses 22.2 points [8.3, 36.1] on the 36 questions whose answer is a smaller one.

\begin{table}[!tp]
\centering
\caption{Counting ablations on the MME-RealWorld test split (accuracy, \%), with the paired difference to the final
rule and its 95\% interval underneath.}
\label{tab:supp:ablation}
\footnotesize
\setlength{\tabcolsep}{3pt}
\resizebox{\textwidth}{!}{%
\begin{tabular}{llrccccccc}
\toprule
 & & & Final & \multicolumn{2}{c}{Zeros accepted} & \multicolumn{2}{c}{No subdivision} & \multicolumn{2}{c}{Fixed 4$\times$4 grid} \\
\cmidrule(lr){5-6}\cmidrule(lr){7-8}\cmidrule(lr){9-10}
Setting & $N$ & Questions & Acc. & Acc. & $\Delta$ & Acc. & $\Delta$ & Acc. & $\Delta$ \\
\midrule
Qwen3-VL-8B, programs & all & 981 & 36.60 & 36.70 & \dci{+0.1}{-1.1}{+1.3} & 35.98 & \dci{-0.6}{-2.3}{+1.1} & 33.74 & \dci{-2.9}{-5.8}{+0.1} \\
 & $\leq 5$ & 700 & 37.29 & 38.14 & \dci{+0.9}{-0.6}{+2.3} & 37.71 & \dci{+0.4}{-1.1}{+2.0} & 29.29 & \dci{-8.0}{-11.7}{-4.1} \\
 & 6--20 & 261 & 35.25 & 33.33 & \dci{-1.9}{-4.6}{+0.4} & 32.57 & \dci{-2.7}{-7.3}{+1.9} & 46.74 & \dci{+11.5}{+6.9}{+16.1} \\
 & $>20$ & 15 & 40.00 & 40.00 & \dci{0.0}{0.0}{0.0} & 26.67 & \dci{-13.3}{-46.7}{+20.0} & 26.67 & \dci{-13.3}{-33.3}{0.0} \\
\midrule
Qwen3-VL-8B, whole image & all & 981 & 32.62 & 33.33 & \dci{+0.7}{-1.1}{+2.5} & 32.72 & \dci{+0.1}{-1.9}{+2.1} & -- & -- \\
 & $\leq 5$ & 700 & 30.43 & 32.86 & \dci{+2.4}{+0.1}{+4.7} & 32.57 & \dci{+2.1}{-0.1}{+4.4} & -- & -- \\
 & 6--20 & 261 & 39.08 & 35.25 & \dci{-3.8}{-7.3}{-0.4} & 34.48 & \dci{-4.6}{-8.8}{-0.4} & -- & -- \\
 & $>20$ & 15 & 33.33 & 33.33 & \dci{0.0}{0.0}{0.0} & 20.00 & \dci{-13.3}{-40.0}{+13.3} & -- & -- \\
\midrule
Qwen3-VL-4B, programs & all & 981 & 36.80 & 35.58 & \dci{-1.2}{-2.5}{0.0} & 34.56 & \dci{-2.2}{-4.0}{-0.6} & -- & -- \\
 & $\leq 5$ & 700 & 39.00 & 37.71 & \dci{-1.3}{-3.0}{+0.4} & 37.86 & \dci{-1.1}{-2.9}{+0.6} & -- & -- \\
 & 6--20 & 261 & 32.57 & 31.42 & \dci{-1.1}{-3.1}{+0.4} & 27.20 & \dci{-5.4}{-9.6}{-1.1} & -- & -- \\
 & $>20$ & 15 & 20.00 & 20.00 & \dci{0.0}{0.0}{0.0} & 20.00 & \dci{0.0}{-26.7}{+26.7} & -- & -- \\
\bottomrule
\end{tabular}}
\par\smallskip
\begin{minipage}{\textwidth}\footnotesize
Replayed from the VLM answers recorded in the test runs; the fixed grid uses new answers on the same regions. $N$:
true number of objects; the five questions whose answer is the option stating that the image does not show the number belong
to no stratum. Programs: the regions named by the programs; whole image: the run in which every program is replaced by
a count over the whole image, made with Qwen3-VL-8B only.
\end{minipage}
\end{table}
\subsection{Direct answers on the program's window}
\label{supp:results:window}
To separate where a program looks from what it computes there, the backbone answered each test question directly on
the window its program chose, with the question and its options and the first-token scores of the option letters.
The windows are those recorded in the test runs. On RC the window is the crop that
was counted, at 1\,M pixels. On OSR it is the union of the boxes of the two referents, enlarged to 1.3 times its
longer side and to at least 768 pixels, at 2.5\,M pixels. On MME-RealWorld counting it is the counted region with its
margin, at 2.5\,M pixels; the 51 programs that count grounded boxes have no region and are left out. Only questions
whose program executed are compared, with GeoScope's executed answer before any fallback (Table~\ref{tab:supp:window}).
A direct answer takes one VLM call, against a median of two, five and seven calls for GeoScope on RC, OSR and
counting (Section~\ref{sec:exp:cost}). On MME-RealWorld the direct answer picks the option stating that the image does not show
the number for 31\% of the questions, so the table also gives the highest-scoring numeric option.

\begin{table}[!tp]
\centering
\caption{GeoScope's executed answer against the backbone's direct answer on the same window (Qwen3-VL-8B, test split,
accuracy, \%). Difference: GeoScope minus the direct answer, with its 95\% interval (paired bootstrap).}
\label{tab:supp:window}
\small
\setlength{\tabcolsep}{4pt}
\begin{tabular}{lcccc}
\toprule
 & Questions & GeoScope & Direct answer & Difference \\
\midrule
RC & 77 & 64.94 & 68.83 & $-3.9$ [$-14.3$, 6.5] \\
OSR & 382 & 45.29 & 38.48 & 6.8 [0.8, 12.8] \\
MME-RS count & 911 & 37.43 & 24.92 & 12.5 [8.6, 16.5] \\
\ numeric options only & 911 & 37.43 & 34.25 & 3.2 [$-0.9$, 7.2] \\
\bottomrule
\end{tabular}
\end{table}

\subsection{InternVL3.5-8B as the perception model}
\label{supp:results:internvl}
To see whether GeoScope depends on the Qwen family, we replaced its perception model by
InternVL3.5-8B~\citep{internvl35} on the development split. The synthesiser, prompts, parsers and parameters stay as
they are, and boxes use the same coordinates from 0 to 1{,}000. InternVL3.5 receives the default system message of its
own chat template. Its zero-shot answers at 4\,M pixels, scored by first-token letter logits, serve as reference and
fallback. Table~\ref{tab:supp:internvl} gives the results.

With InternVL3.5-8B, GeoScope gains 13.1 points [6.1, 19.6] on counting, and its accuracy is not significantly different from
that of GeoScope with Qwen3-VL-8B, $-2.4$ [$-8.6$, 3.7], although GeoScope was tuned on this split with Qwen3-VL-8B. On OSR its gain of 5.0 is not significant, and it
abstains on 15 of the 100 questions, 13 of them because no direction could be computed. On counting it abstains on 45
of the 245 questions, 28 of them because the count InternVL3.5 returned could not be parsed. On RC it gains nothing and
stays 20 points below GeoScope with Qwen3-VL-8B. RC asks mostly about small sets, with a median of three objects, and on DOTA crops InternVL3.5-8B
overcounts one or two objects by 127\% on average (Table~\ref{tab:supp:models}).

\begin{table}[!tp]
\centering
\caption{InternVL3.5-8B as the perception model on the development split (accuracy, \%).}
\label{tab:supp:internvl}
\small
\setlength{\tabcolsep}{4pt}
\begin{tabular}{lccc}
\toprule
 & RC (20) & OSR (100) & MME-RS count (245) \\
\midrule
Zero-shot, Qwen3-VL-8B & 45.0 & 23.0 & 26.5 \\
GeoScope, Qwen3-VL-8B & 60.0 & 56.0 & 43.3 \\
\ abstained & 0\% & 2\% & 2\% \\
\midrule
Zero-shot, InternVL3.5-8B & 40.0 & 37.0 & 27.8 \\
GeoScope, InternVL3.5-8B & 40.0 & 42.0 & 40.8 \\
\ abstained & 0\% & 15\% & 18\% \\
\ $\Delta$ to its zero-shot & \dci{0.0}{-25.0}{+25.0} & \dci{+5.0}{-8.0}{+18.0} & \dci{+13.1}{+6.1}{+19.6} \\
\ $\Delta$ to Qwen3-VL-8B & \dci{-20.0}{-45.0}{0.0} & \dci{-14.0}{-27.0}{-1.0} & \dci{-2.4}{-8.6}{+3.7} \\
\bottomrule
\end{tabular}
\par\smallskip
\begin{minipage}{\textwidth}\footnotesize
$\Delta$: paired difference of GeoScope with InternVL3.5-8B to its own zero-shot answers and to GeoScope with
Qwen3-VL-8B, with the 95\% interval underneath. Abstained: share of questions answered by the zero-shot fallback.
\end{minipage}
\end{table}
\subsection{Sensitivity to pixel budget, horizon and margin}
\label{supp:results:budget}
All variants of the pixel budget $B$ and the horizon $h$ were run on the development split only
(Table~\ref{tab:supp:budget}), and the test runs of OSR and counting use $B=2.5$\,M pixels and $h=5$. On OSR the
development accuracy peaks at the chosen budget, 56 against 42 at 1\,M and 43 at 4\,M pixels. On counting the curve is
flatter, with 38.8, 43.3 and 41.2. The held-out OSR accuracy of the chosen budget, 44.00, is close to the development accuracy of the other two
budgets, so most of the peak is the noise of choosing among variants on 100 questions. The zero-shot reference is at its best at 4\,M pixels or within noise of it,
with 45, 23 and 26.5 on RC, OSR and counting against 40, 25 and 19.2 at 2.5\,M and 35, 25 and 17.1 at 1\,M pixels.

The horizon matters little. On the development counting questions, with the configuration that still accepted zeros
from downsampled views, $h=3$, 5 and 8 give 40.4, 41.2 and 41.2. On the first set of DOTA crops of Section~\ref{sec:horizon:measure},
the recursive errors for the three values nearly coincide beyond ten objects. The number of views falls only from
4{,}660 to 4{,}063 as $h$ grows from 3 to 8, because most splits are decided by large answers and by the minimum cell
rather than by the threshold. A larger $h$ splits fewer small sets and lowers the surplus at 6 to 10 objects, from
31\% at $h=3$ to 24\% at $h=8$, as Proposition~\ref{prop:split} predicts for splits below the horizon. For Qwen3-VL-2B, whose
shortfall begins at three to five objects, $h=2$ reduces the error at 6 to 10 objects from $-23$\% to $-12$\% but
leaves $-38$\% at 11 to 20 objects.

When every accepted count must come from a native-resolution view instead of being triggered by the answer, development accuracy falls from 41.2 to 37.1 and the median relative count error rises from
0.71 to 1.00, with more overcounts.

The 15\% margin around a counted region was also chosen on the development split. With a margin of 5\% or none,
development counting accuracy falls from 43.3 to 40.0 and 37.6, the latter by 5.7 points [$-9.4$, $-2.4$]. Among the
questions counted in one view of their region, 80 or 81 counts are too high at every margin, while those too low rise
from 50 to 59 and 67 and the exact ones fall from 34 to 25 and 19. A narrower window thus cuts off objects on the
border of a coarsely grounded region and does not remove overcounts. Counts of whole images, which have no margin, are
the same in all three runs.

\begin{table}[!tp]
\centering
\caption{Pixel budget per view $B$ and horizon $h$ on the development split (accuracy, \%), all other settings at their final values. \shadekey{}: best per row.}
\label{tab:supp:budget}
\small
\setlength{\tabcolsep}{4pt}
\begin{tabular}{lccc}
\toprule
$B$ & 1\,M & 2.5\,M & 4\,M \\
\midrule
GeoScope, OSR (100) & 42 & \overall 56 & 43 \\
GeoScope, MME-RS count (245) & 38.8 & \overall 43.3 & 41.2 \\
Zero-shot, RC (20) & 35 & 40 & \overall 45 \\
Zero-shot, OSR (100) & \overall 25 & \overall 25 & 23 \\
Zero-shot, MME-RS count (245) & 17.1 & 19.2 & \overall 26.5 \\
\midrule
$h$ & 3 & 5 & 8 \\
\midrule
GeoScope, MME-RS count (245) & 40.4 & 41.2 & 41.2 \\
\ median relative count error & 0.80 & 0.71 & 0.71 \\
DOTA crops, views & 4{,}660 & 4{,}275 & 4{,}063 \\
\ error, 6--10 objects & +0.31 & +0.30 & +0.24 \\
\ error, 81--160 objects & $-$0.15 & $-$0.16 & $-$0.16 \\
\ error, more than 160 objects & $-$0.21 & $-$0.22 & $-$0.22 \\
\bottomrule
\end{tabular}
\par\smallskip
\begin{minipage}{\textwidth}\footnotesize
The horizon was varied while zeros from downsampled views were still accepted; that configuration scores 41.2 at $h=5$, the final one 43.3. DOTA crops: mean signed relative error of the recursive count on the first set of crops of Section~\ref{sec:horizon:measure}, Qwen3-VL-8B, and the number of views it needs.
\end{minipage}
\end{table}

\subsection{UHR-Micro data, protocol and breakdown}
\label{supp:umicro}
\paragraph{Data} We use the test split of the five task types in Table~\ref{tab:umicro}, 1{,}369 questions, and the
validation split of the same task types, 100 questions each, whose images the test split does not use. The public
release lacks 105 of the SODA-A images these questions use. We took them from the SODA-A release~\citep{sodaa} by
their file number; three images present in both releases are identical pixel by pixel. All questions are evaluated.

\paragraph{Zero-shot answers} The benchmark's code for direct answering runs on several large GPUs and its output
instructions are not released. We re-run Qwen3-VL-8B on one RTX 3090 with its released image preprocessing (up to
16{,}777{,}216 pixels), its conversion of pixel coordinates in a question to a 0--1{,}000 grid, the question followed
by its closing instruction, greedy decoding, and its released parser and metric. Every answer is parsed. The re-run
lies within 3.4 points of the reported scores on four task types and 7.5 points above on conditional counting
(Table~\ref{tab:umicro}). InternVL3.5-8B is re-run with its own preprocessing, up to 12 tiles of $448\times448$
pixels and a thumbnail, and otherwise in the same way; it lies within 4.4 points of the reported scores on all five
task types. We compare GeoScope with the re-run of its backbone, question by question.

\paragraph{GeoScope} The counting programs run with the frozen configuration of MME-RealWorld counting. Of the 544
regional counting questions, 461 give the region as a pixel box. For them the program counts the class in that box,
which needed one addition to the program language, a region given by its corners; the other 83 describe the region
in words and are synthesised like every counting question. On the validation split, the usual 15\% view margin
around a given box scored 48.30 against 45.34 without it on the 84 questions with a box, and was kept. The
direction and distance questions mark the reference object in red and the others in blue, green, yellow or purple,
each in one fixed colour. The executor takes the bounding box of the largest connected outline of each colour and
answers from the boxes, the direction by the direction-relation matrix as for OSR and the distance between box
centres. On the validation split these rules agree with the reference answer on all 200 questions, as do the centre
angle and the distance between box edges. The configuration was fixed on the validation split before the test split
was evaluated. Three regional and 19 conditional counting questions fell back to the zero-shot answer. With
InternVL3.5-8B the same frozen configuration runs with that model as the VLM, set up as in
\appref{supp:results:internvl}, and 2 global, 10 regional and 58 conditional counting questions fall back to its
zero-shot answer. For 44 of the 58 the place or anchor that the question names is not found or is grounded with an
empty box; the other 14 have programs that are not well typed, as with Qwen3-VL-8B.

\paragraph{Horizon of InternVL3.5-8B} The shortfall of InternVL3.5-8B begins beyond 40 objects
(Table~\ref{tab:fit}), and the upper end of the bin before it gives $h=40$. A larger threshold only removes splits,
so the counts with $h=40$ follow from the answers already recorded with $h=5$. On the test split, global counting rises from 42.23 to 44.01 (+1.8 [$-0.2$, 3.8]),
regional counting falls from 58.77 to 56.88 ($-1.9$ [$-3.2$, $-0.5$]) and conditional counting is 26.40 against 26.19
(+0.2 [$-1.2$, 1.4]). On the validation split the differences are +5.6, $-2.7$ and +0.6. The answers that change
show why. In global counting, 92 of the 108 changed answers with $h=5$ are too high, the spurious counts of splits
below the horizon; InternVL3.5-8B answers an empty view with 2.4 objects on average, the most of the models in
Table~\ref{tab:fit} (column $s$). With $h=40$ fewer views are split, and on the 83 of these questions about at most 20 objects the score
rises. Of the 25 about more objects, 19 answers with $h=40$ are too low against 5 with $h=5$, and in regional counting
the 164 changed answers about more than 20 objects include 90 that are too low against 76. On these views the
single-view answers of InternVL3.5-8B already fall short below 40 objects, as the horizon depends on the view
(Section~\ref{sec:horizon:measure}), and $h=40$ accepts them where $h=5$ splits. With $h=40$ the global
count would be the highest in Table~\ref{tab:umicro}, above the 43.19 of MAP, and the regional count below its 57.81.
Neither threshold is better on both tasks, and the table keeps $h=5$ for both backbones.

\paragraph{Breakdown by question type} On the test split, GeoScope scores 53.30 against 11.76 on the regional counting
questions with a box and 27.36 against 20.72 on those that describe the region. On conditional counting, the 93
questions about a fine-grained aircraft type score 8.3 against 3.3, and the other 109 score 25.3 against 40.6; their
median signed relative error is zero, but the counts scatter widely, as for any question whose place is grounded
wrongly.

\paragraph{Without subdivision} To measure what subdivision adds, we take the recorded answer of Qwen3-VL-8B
on the first, unsplit view of every counted region in the test run, the view a count without subdivision would use. Abstained questions keep the zero-shot answer, and the eight global counting questions whose image
is cut into cells keep their full count, which slightly understates the effect. On global counting the full system
scores 40.49, the unsplit view alone 24.87 and zero-shot answering 23.71, so subdivision adds 15.6 points
[11.7, 19.6], while counting the whole image in one view adds 1.2 [$-1.4$, 3.8] over zero-shot answering. On regional
counting the three score 49.34, 42.43 and 13.13. Restricting the count to the given box adds 29.3 points [26.3, 32.4]
and subdivision 6.9 [4.6, 9.2].

\section{Cost breakdown}
\label{supp:cost}
Table~\ref{tab:supp:costall} lists the mean seconds per question that Fig.~\ref{fig:cost} plots against
accuracy, and also those of RAP with LLaVA-OV-7B on the fixed 20\% subset of the counting split, which the figure
leaves out.

\begin{table}[!tp]
\centering
\caption{Mean seconds per question of every method on one NVIDIA GeForce RTX 3090.}
\label{tab:supp:costall}
\small
\setlength{\tabcolsep}{5pt}
\begin{tabular}{lccc}
\toprule
Method & RC & OSR & MME-RS count \\
\midrule
Zero-shot (Qwen3-VL-8B) & 2.8 & 3.1 & 2.8 \\
WeaveEarth & 6.9 & 6.3 & 5.6 \\
GazeEarth & 12.9 & 18.5 & 16.0 \\
ZoomSearch & 96.3 & 86.0 & 62.1 \\
RAP (Qwen3-VL-8B) & -- & -- & 400.1 \\
RAP (LLaVA-OV-7B) & -- & -- & 827.6 \\
GeoScope & 3.0 & 28.3 & 22.4 \\
\bottomrule
\end{tabular}
\end{table}

Table~\ref{tab:supp:cost} breaks down the cost of GeoScope with Qwen3-VL-8B on 20 RC, 60 OSR and 60 counting questions
of the development split. All runs use one NVIDIA GeForce RTX 3090 with 24\,GiB from a fresh start, and times exclude model loading. Every VLM call
records its answer when it returns. Within a question, the time between consecutive records is charged to the later
call, and the time after the last call to execution. Visual tokens count one token per $32\times32$ pixels of every
image the VLM receives. Synthesis was timed separately with Qwen3-4B from a fresh start, one question at a time and in
batches of 16, the batch size of our runs.

Nearly all of the time goes to VLM calls. Grounding is the expensive kind, at 3.6 to 9\,s per call, because the VLM
generates a list of boxes, while a count costs about 0.45\,s. An OSR question spends 24.3\,s on 6.7 grounding calls on
average, and a counting question 12.6\,s on 1.4 grounding calls and 7.3\,s on 16.4 counts. RC needs no grounding,
since the red circle is found by its colour, and its crop is counted 2.9 times on average. Executing a program takes
less than 10\,ms. Counting has a long tail, with a median of 6.7\,s and a maximum of 185\,s for a region that needs
deep subdivision. Peak GPU memory is 19.3\,GiB for perception and 14.5\,GiB for synthesis.

The same programs were also served with vLLM~\citep{vllm}, on the same GPU and questions, with the independent calls of
a question issued concurrently: the parts of a split view, the cells of a long window and the zoom windows of a
coarse-to-fine search. The mean time per question was 2.3, 22.2 and 13.6\,s for RC, OSR and counting, against 1.8,
24.3 and 19.8\,s, and the longest counting question took 67\,s instead of 185\,s. Served one call at a time, the
times were 2.3, 23.3 and 22.9\,s. The served model gave a different answer on 8 of the 140 questions, mostly
through different grounding outputs, so all reported results use the original implementation.

The reproduced methods were timed on the test questions, on the same GPU model, from their per-question records.
WeaveEarth scores tiles with SigLIP2 and calls its VLM once per question, in 5.6 to 6.9\,s. GazeEarth makes two calls
per counting question, with 5{,}435 visual tokens and 329 generated tokens on average, in 16.0\,s. ZoomSearch scores
20.1 to 24.0 views per question with the confidence of its VLM, and RAP 30.9 layouts with Qwen3-VL-8B and 379.8 with
LLaVA-OneVision-7B, whose setting allows 200 steps. For reference, WeaveEarth reports
7.59\,s per sample for itself on an NVIDIA A100, averaged over the two benchmarks it evaluates, and ZoomSearch reports 60.02\,s per MME-RealWorld remote sensing sample
without naming the GPU, against 62.1\,s in our runs.

\begin{table}[!tp]
\centering
\caption{Cost of GeoScope with Qwen3-VL-8B per question, on development questions, one RTX 3090, from a fresh start, model loading excluded.}
\label{tab:supp:cost}
\footnotesize
\setlength{\tabcolsep}{4pt}
\begin{tabular}{lccc}
\toprule
 & RC (20) & OSR (60) & MME-RS count (60) \\
\midrule
Synthesis, single question (s) & 1.17 & 3.95 & 2.55 \\
Synthesis, batch of 16 (s) & 0.40 & 1.02 & 1.26 \\
Grounding (s) & -- & 24.3 [6.7] & 12.6 [1.4] \\
Counting (s) & 1.8 [2.9] & -- & 7.3 [16.4] \\
Execution (s) & $<$0.01 & $<$0.01 & $<$0.01 \\
Wall time, median / mean (s) & 1.3 / 1.8 & 17.2 / 24.3 & 6.7 / 19.8 \\
Wall time, p90 / maximum (s) & 3.0 / 10.9 & 48.6 / 86.9 & 42.9 / 185.1 \\
VLM calls, median / mean & 2 / 2.9 & 5 / 6.7 & 7 / 17.9 \\
Visual tokens, median / mean & 1.7k / 2.2k & 11.0k / 13.0k & 13.4k / 17.0k \\
Total, single question (s) & 3.0 & 28.3 & 22.4 \\
\midrule
Zero-shot, mean (s) & 2.8 & 3.1 & 2.8 \\
Zero-shot, visual tokens, median & 3.8k & 3.8k & 3.8k \\
\bottomrule
\end{tabular}
\par\smallskip
\begin{minipage}{\textwidth}\footnotesize
Synthesis: seconds per question, synthesising one question at a time or batches of 16. Grounding and counting: mean seconds, with the mean number of calls in brackets. Wall time: time per question without synthesis; p90, its 90th percentile. Total: single-question synthesis plus mean wall time. Visual tokens: one per $32\times32$ pixels of every image the VLM receives.
\end{minipage}
\end{table}

\section{Failure analysis and examples}
\label{supp:failures}
Hand-written programs change at most two of thirty answers per task (\appref{supp:method:correctness}), so we look
at what perception gets wrong on the test split. Of the 209 OSR questions answered wrongly by an executed program, the computed direction lies
more than 45 degrees from the correct option in 179, and is roughly opposite to it in 48. With both referents grounded
on the right objects, the computed direction would lie near the correct option, so errors this large mean that a
referent was grounded on a different object, or that the annotation disagrees with the geometry. Only 30 wrong
directions lie within 45 degrees of the correct option, near the boundary between two options. Of the 605 counting
questions answered wrongly with a numeric answer, 243 are off by one or two objects and 249 by more than five. The count is
too low in 315 and too high in 290.

XLRS-Bench overall counting asks for all instances of a class in the image, mostly ships, cars and boats, which make
up 38 of the 48 test questions. The true counts have a median of 61.5 and reach 734. Adjacent options differ by a
median of 17\% of the smaller one, so a count must lie within about 8\% of the truth to select the correct option.
GeoScope cuts the image into cells of at most 2{,}000 pixels and counts each with count-triggered subdivision. Its
counts are nearly unbiased, with a median signed error of $-0.9$\%, 25 too low and 21 too high. Their median absolute
error is 17.7\%, however, and only 9 of the 48 lie within 8\% of the truth. Table~\ref{tab:supp:oc} compares the
methods we ran. GeoScope gains 25.0 points over its zero-shot backbone and is 2.1 points below both ZoomSearch and
always choosing the second-largest number, with intervals of about 20 points either way.

\begin{table}[!tp]
\centering
\caption{XLRS-Bench overall counting, test split, 48 questions (accuracy, \%).}
\label{tab:supp:oc}
\small
\setlength{\tabcolsep}{4pt}
\begin{tabular}{llcc}
\toprule
Method & Backbone & Accuracy & $\Delta$ \\
\midrule
Constant answer D & -- & 31.25 & -- \\
Second-largest number & -- & 37.50 & -- \\
Zero-shot & Qwen3-VL-8B & 10.42 & -- \\
Zero-shot & LLaVA-OV-7B & 22.92 & -- \\
\midrule
WeaveEarth & Qwen3-VL-8B & 6.25 & \dci{-4.2}{-12.5}{+4.2} \\
GazeEarth & Qwen3-VL-8B & 12.50 & \dci{+2.1}{-6.2}{+10.4} \\
ZoomSearch & LLaVA-OV-7B & 37.50 & \dci{+14.6}{-2.1}{+31.2} \\
GeoScope & Qwen3-VL-8B & 35.42 & \dci{+25.0}{+8.3}{+41.7} \\
\bottomrule
\end{tabular}
\par\smallskip
\begin{minipage}{\textwidth}\footnotesize
Underneath: paired difference to the zero-shot answers of each method's own backbone, with its 95\% interval. Constant
answers: the best letter, D, and the best rank, the second-largest of the four numbers, among those tied on the
development split, both chosen with the test split and therefore upper bounds
(Section~\ref{supp:protocol:constant}).
\end{minipage}
\end{table}

Table~\ref{tab:supp:logs} shows two executions on MME-RealWorld counting test questions, both with a program that
counts a class in one quarter of a 4{,}096-pixel DOTA image. In the first, the lower-left quarter, widened by 15\% on
each side, is too large for one view at 2.5\,M pixels, and the model answers 7. Since 7 exceeds the horizon, the view
is split into four parts at native resolution, whose counts of 0, 0, 4 and 5 are accepted and sum to the true 9. The
zero-shot model chose 3. In the second, the first view answers 10, and one part again exceeds the horizon. The leaves
sum to 11 ships where the annotation has 2, and option matching selects the largest option, 4, while the zero-shot model
answers correctly. The recorded answers do not show whether the extra ships are spurious answers or real ships that lie in the quarter
but outside the corner the question means. For one or two objects the correct number is never the largest offered
(Table~\ref{tab:supp:mmelayout}), so such an overshoot always fails.

\begin{table}[!tp]
\centering
\caption{Two executions on MME-RealWorld counting test questions (DOTA images of about 4{,}096 pixels on a side).}
\label{tab:supp:logs}
\footnotesize
\begin{verbatim}
(a) How many gray planes are there in the lower left side of the picture?
    A 7   B 8   C 3   D 9   E          truth D        zero-shot C
    count(in_region(detect("gray plane"), part(LL)))
    (0,1741)-(2356,4097)          scale 0.67   count  7 > h, split
      (0,1741)-(1178,2919)        native       count  0
      (1178,1741)-(2356,2919)     native       count  0
      (0,2919)-(1178,4097)        native       count  4
      (1178,2919)-(2356,4097)     native       count  5
    sum 9 -> D                                 correct

(b) How many ships are there in the bottom right corner of the picture?
    A 2   B 3   C 4   D 1   E          truth A        zero-shot A
    count(in_region(detect("ship"), part(LR)))
    (1740,1740)-(4096,4096)       scale 0.67   count 10 > h, split
      (1740,1740)-(2918,2918)     native       count  7 > h, split
        (1740,1740)-(2329,2329)   scale 1.30   count  1
        (2329,1740)-(2918,2329)   scale 1.30   count  1
        (1740,2329)-(2329,2918)   scale 1.30   count  0
        (2329,2329)-(2918,2918)   scale 1.30   count  5
      (2918,1740)-(4096,2918)     native       count  0
      (1740,2918)-(2918,4096)     native       count  2
      (2918,2918)-(4096,4096)     native       count  2
    sum 11 -> C, the largest option            wrong
\end{verbatim}
\par\smallskip
\begin{minipage}{\textwidth}\footnotesize
Windows: corners $(x_0,y_0)$--$(x_1,y_1)$ in pixels of the original image. Scale: the factor at which a window is
shown to the VLM at 2.5\,M pixels; windows below 768 pixels are enlarged. Every view is asked the counting question of
\appref{supp:method:prompts} for the program's class. E: the image does not feature the number.
\end{minipage}
\end{table}

\end{document}